\documentclass[12pt,a4paper]{report}
\usepackage{thesis}
\usepackage{lmodern}
\usepackage[numbers,sort&compress]{natbib}
\usepackage[hidelinks]{hyperref}
\usepackage[left=2.5cm,top=2.5cm,right=2.5cm,bottom=2.5cm]{geometry}
\usepackage{nomencl}
\usepackage{booktabs}
\usepackage{tabularx}
\usepackage{latexsym}
\usepackage{mathrsfs}
\usepackage{amsmath}
\usepackage{amssymb}
\usepackage{graphicx}
\usepackage{algorithm}
\usepackage{algpseudocode}
\usepackage{multirow}
\usepackage{booktabs}
\usepackage{lipsum}
\usepackage{enumitem}
\usepackage{tabto}
\usepackage{mathrsfs}
\usepackage{caption}
\usepackage{subcaption}
\usepackage{parskip}
\usepackage{wrapfig}

\usepackage{mathptmx} 
\usepackage[T1]{fontenc}

\makenomenclature

\usepackage{tocloft}

\newcommand{\thesistitle}{\textbf{Taming the Beast: Learning to\\ Control Neural Conversational Models}}
\newcommand{\thesisauthor}{\textbf{Andrea Madotto}}
\newcommand{\programname}{Electronic and Computer Engineering}
\newcommand{\departmentname}{Department of Electronic and Computer Engineering}
\newcommand{\thesisdate}{August 2021}

\newcommand{\thesisdegree}{PhD}

\begin{document}

\pagenumbering{roman}
\pagestyle{plain}
\setcounter{page}{1}
\addcontentsline{toc}{chapter}{Title Page}
\thispagestyle{empty}
\null\vskip0.5in
\begin{center}
  \begin{LARGE}
    \thesistitle
  \end{LARGE}
  \vfill
  \vspace{20mm}

  \begin{Large}
  by
  \end{Large}

  \vspace{4mm}
  \begin{Large}
  \thesisauthor 
  \end{Large}\\
  \vfill
  \vspace{20mm}
  \begin{large}
  A Thesis Submitted to\\
  The Hong Kong University of Science and Technology \\
  in Partial Fulfillment of the Requirements for\\
  the Degree of Doctor of Philosophy \\
  in the Department of \programname \\
  \vfill \vfill
  \thesisdate, Hong Kong
  \end{large}
  \vfill
\end{center}

\vfill





\newpage
\addcontentsline{toc}{chapter}{Table of Contents}
\tableofcontents

\newpage
\addcontentsline{toc}{chapter}{List of Figures}
\listoffigures

\newpage
\addcontentsline{toc}{chapter}{List of Tables}
\listoftables

\newpage
\addcontentsline{toc}{chapter}{Abstract}
\begin{center}
{\Large \thesistitle}\\
\vspace{20mm}
by \thesisauthor\\
\departmentname\\
The Hong Kong University of Science and Technology
\end{center}
\vspace{8mm}
\begin{center}
Abstract
\end{center}
\par
\noindent

This thesis investigates the controllability of deep learning-based, end-to-end, generative dialogue systems in both task-oriented and chit-chat scenarios. In particular, we study the different aspects of controlling generative dialogue systems, including controlling styles and topics and continuously adding and combining dialogue skills.
 
In the three decades since the first dialogue system was commercialized, the basic architecture of such systems has remained substantially unchanged, consisting of four pipelined basic components, namely, natural language understanding (NLU), dialogue state tracking (DST), a dialogue manager (DM) and natural language generation (NLG). The dialogue manager, which is the critical component of the modularized system, controls the response content and style. This module is usually programmed by rules and is designed to be highly controllable and easily extendable.
 
With the emergence of powerful "deep learning" architectures, end-to-end generative dialogue systems have been proposed to optimize overall system performance and simplify training. However, these systems cannot be easily controlled and extended as the modularized dialogue manager can. This is because a single neural system is used, which is usually a large pre-trained language model (e.g., GPT-2), and thus it is hard to surgically change desirable attributes (e.g., style, topics, etc.). More importantly, uncontrollable dialogue systems can generate offensive and even toxic responses. 
 
Therefore, in this thesis, we study controllable methods for end-to-end generative dialogue systems in task-oriented and chit-chat scenarios. Throughout the chapters, we describe 1) how to control the style and topics of chit-chat models, 2) how to continuously control and extend task-oriented dialogue systems, and 3) how to compose and control multi-skill dialogue models.
 
To elaborate, we firstly propose a residual adapter model to control style and topics in conversational models such as DialoGPT, Meena, and Blender-Bot. Our proposed model adds less than 1.5\% task-specific parameters per style/topic, making it deployable for online systems. We run a comprehensive automatic and human evaluation to show controllability in the response generation in terms of style and topics, without losing fluency without requiring dialogue-specific datasets.
 
Secondly, we propose a highly controllable architectural method based on residual adapters for continuous update of task-oriented dialogue systems with new features based on the user's needs, e.g., adding new slots and intents or even completely new domains. Moreover, we analyze the trade-off between performance, number-of-parameters, and episodic memory sizes in other methods (regularization, rehearsal, architectural).
 
Finally, we propose a novel theoretical framework to control the end-to-end dialogue model with multiple composable and control skills. We empirically show the effectiveness of using specialized parameters in combined chit-chat and task-oriented datasets.

\printnomenclature[1in]

\newpage
\pagenumbering{arabic}
\pagestyle{plain}
\setcounter{page}{1}
\chapter{Introduction}

Conversational AI systems interact with human users while completing user requests or simply chatting. These systems have applications ranging from personal assistance and health assistance to customer service, etc., and they are widely deployed in personal assistants such as Apple Siri, Amazon Alexa, Microsoft XiaoIce, Google Assistant etc. Dialogue systems are mainly categorized as task-oriented or chit-chat systems. The first aims to help the user to achieve certain goals (e.g., booking a restaurant) and tries to minimize the number of interactions with the users. The second, on the other hand, aims to establish a general conversation with the user and tries to maximize the length of the interaction. In both cases, the system needs to generate responses that are coherent with the dialogue history, and in many cases they require external knowledge (e.g., Wikipedia, knowledge bases etc.). Examples of a conversation for task-oriented systems and for a chit-chat system are is shown in Table~\ref{Example-table}.

\begin{table}[H]
\renewcommand{\arraystretch}{1.0} 
\linespread{1.0}\selectfont\centering
\begin{tabular}{rl}
\hline
\multicolumn{2}{c}{\textit{\textbf{Task-Oriented}}}                                                             \\ \hline
\multicolumn{1}{r|}{\textit{Usr:}} & I need to check my balance.              \\
\multicolumn{1}{r|}{\textit{Sys:}} & Of course! Which account should I use?    \\ 
\multicolumn{1}{r|}{\textit{Usr:}} & My savings account, please.             \\
\multicolumn{1}{r|}{\textit{Sys:}} & No problem. Your balance is \$139.    \\ 
\hline
\multicolumn{2}{c}{\textit{\textbf{Chit-Chat}}}                                                                 \\ \hline
\multicolumn{1}{r|}{\textit{Usr:}} & Hello! How are you today?            \\
\multicolumn{1}{r|}{\textit{Sys:}} & I am good thanks, what are you up to? \\ 
\multicolumn{1}{r|}{\textit{Usr:}} & Ummm. I really want to try bull riding. Do you have any interest on that?            \\
\multicolumn{1}{r|}{\textit{Sys:}} & I would love to try, can we schedule something next week? \\ 
\hline
\end{tabular}
\caption{An example of a chit-chat and a task-oriented conversation.}
\label{Example-table}
\end{table}

\begin{figure}[t]
    \centering
    \includegraphics[width=0.8\linewidth]{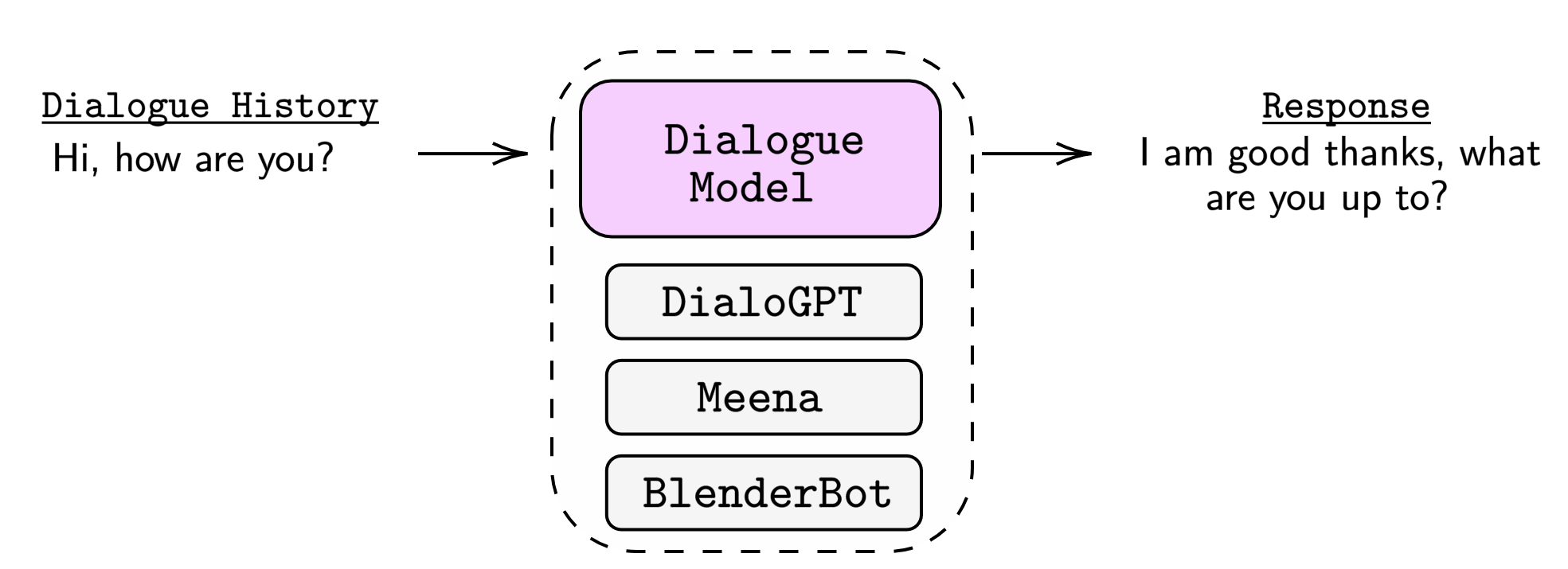}
    \caption{End-to-end chit-chat dialogue system. DialGPT~\cite{zhang2019dialogpt}, Meena~\cite{adiwardana2020towards}, and BlenderBot~\cite{roller2020recipes} are three possible instances of large pre-trained conversational models. }
    \label{fig:chit-chat}
\end{figure}

\section{Chit-Chat Dialogue Systems}
Chit-chat systems are built to chat freely with users and they do not aim to achieve any particular goals, but rather to entertain and engage. The first system of this kind traces back to ELIZA~\cite{weizenbaum1966eliza}, a rule-based system that uses pre-defined patterns and templates to generate the system response. This kind of rudimentary and hand-crafted system can rarely generalize to unseen conversations. Therefore, statistical-based chit-chat systems have been proposed~\cite{serban2015survey} to overcome this problem. These models leverage transcripts of natural spoken conversations or crowd-sourced datasets to learn dialogue systems that can respond like humans. These systems use either Sequence-to-Sequence (Seq2Seq) models~\cite{sutskever2014sequence} or retrieval systems\cite{isbell2000cobot,zhou2020design}, and they are trained end-to-end using human-to-human conversations. 

More recently, large pre-trained language models~\cite{peters2018deep,radford2019language,raffel2019exploring,shin2020generating,lin2019moel} have greatly improved the state-of-the-art in many down-stream tasks. These language models are trained using the simple log-likelihood objective over large amounts of unlabeled data (e.g., Wikipedia articles). This approach results in large powerful language models that produce coherent text and can be used to perform unconditional language generation. Chit-chat conversational systems are a special case of language model where the prefix is the dialogue history and the continuation is a human-like response~\cite{wolf2019transfertransfo}. Recently, large pre-trained language models trained on unlabeled human-to-human conversation (e.g., from Reddit)~\cite{zhang2019dialogpt,adiwardana2020towards,roller2020recipes} have shown excellent performance in modelling human responses. Figure~\ref{fig:chit-chat} provides a high-level overview and examples of a statistical-based chit-chat dialogue systems.

\begin{figure}[t]
    \centering
    \includegraphics[width=\linewidth]{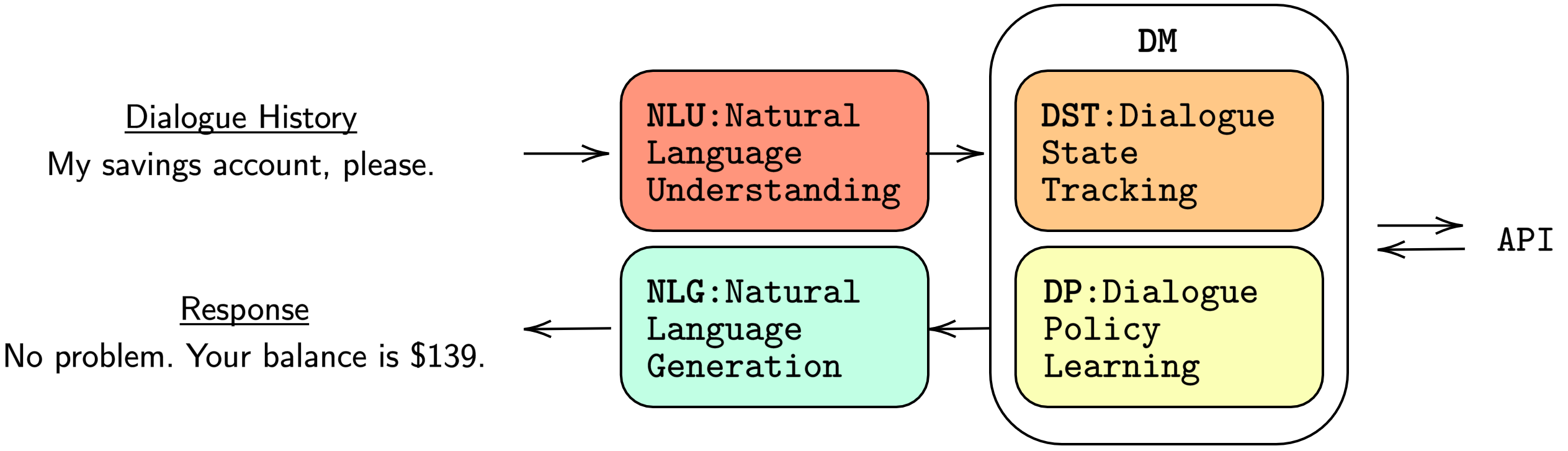}
    \caption{Module-based dialogue systems. Image inspired by ~\citet{williams2016dialog}.}
    \label{fig:Task-oriented}
\end{figure}
\section{Module-based Task-Oriented Dialogue Systems}
Task-oriented dialogue systems are usually built using several modules, Automatic Speech Recognition (ASR), Natural Language Understanding (NLU), a Dialogue Manager (DM), and Natural Language Generation (NLG). Thus they are called module-based~\cite{williams2007partially,hori2009statistical,lee2009example,levin2000stochastic,young2013pomdp}. Figure~\ref{fig:Task-oriented} provides a high-level overview of a module-based task-oriented dialogue system. In this system, each module is used sequentially to produce the user responses. For instance:

\begin{itemize}
    \item[\textbf{NLU:}] the Natural Language Understanding (NLU)~\cite{raymond2007generative,deng2012use,yao2014spoken,guo2014joint,zhang2016joint,liu2020crossner,liu2020importance,wu2019getting,wu2021qaconv} module extracts a semantic frame from the user utterance, which includes the \textit{domain}, the \textit{intent} and \textit{slots} triggered in the current turn. The domain specifies the general topic of the request (e.g., banking), the intent specifies what the user wants to achieve (e.g., getting information about a bank account), and the slots are the specific name and values of the goal (e.g., savings account). Hence, this module includes domain and intent classifiers~\cite{tur2012towards,chen2016zero,liu2019zero} and a slot tagger~\cite{nguyen2007comparisons,mesnil2014using}. The latter is a parsing task that assigns predefined slots types to words. 

    \item[\textbf{DM:}] the Dialogue Manager~\cite{rudnicky1999agenda,young2006using,young2013pomdp} uses the dialogue frame provided by the NLU module to generate a system action, a.k.a. speech-act. The later is a semantic class which represents a high-level description of the response (e.g., request\_location). This module is made of a Dialogue State Tracker (DST) and a Dialogue Policy (DP). The DST generates a dialogue state which is a \textit{global} semantic frame. This models uses the provided frame to update the global dialogue state (e.g., frame), and they are implemented using hand-crafted features, a complex domain-specific lexicon, and a domain ontology ~\cite{williams2007partially,thomson2010bayesian,henderson2014robust} or using statistical models~\cite{williams2016dialog,mrkvsic2016neural}. On the other hand, the DP~\cite{li2009reinforcement,lipton2018bbq} is a classifier that maps the dialogue state to the possible system actions, usually trained using reinforcement learning. Importantly, the DP also issues actions for querying external knowledge bases using slot values present in the dialogue state. 
    
    \item[\textbf{NLG:}] the Natural Language Generation module uses the system action produced by the Dialogue Manager to generate plain text responses. Traditionally, this module is implemented using template responses~\cite{busemann1998flexible}, while more recently, statistical methods~\cite{wen2015semantically,press2017language,winata2020learning,winata2019code,xu2019clickbait} based on Seq2Seq models have been proposed. 
\end{itemize}
Other permutations of these modules have also been explored. For instance, several systems~\cite{rastogi2017scalable,ramadan2018large,zhong2018global,lin2021leveraging} remove the NLU module and replace it with only a DST. In contrast, others build a very strong NLU module and ignore the DST~\cite{chen2018gunrock}.

\begin{figure}[t]
    \centering
    \includegraphics[width=\linewidth]{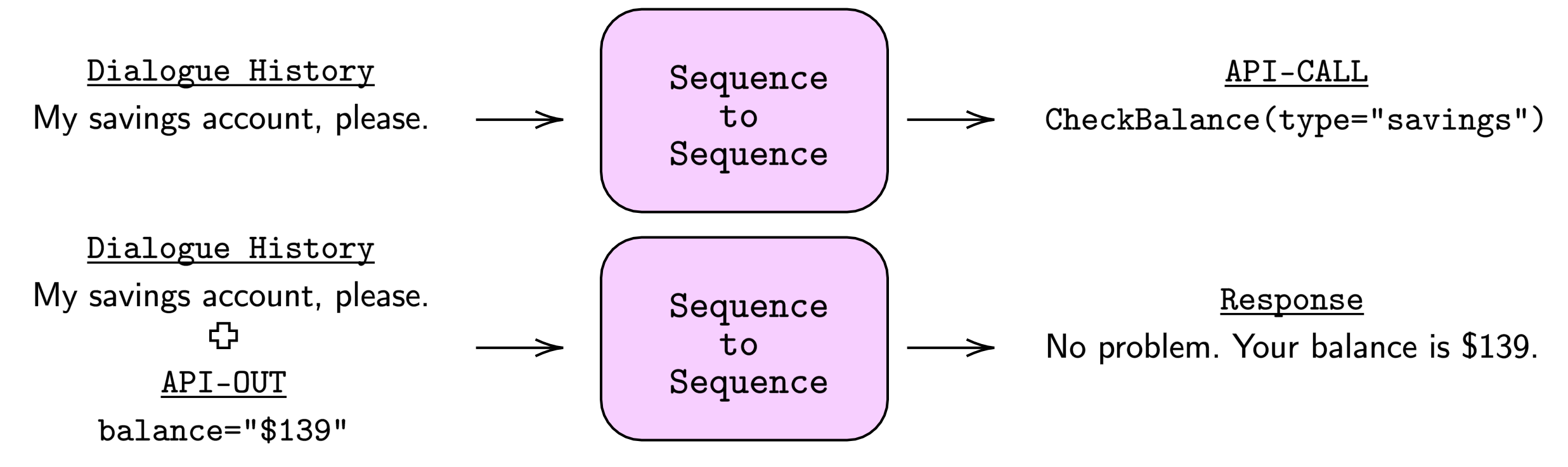}
    \caption{End-to-end task-oriented dialogue system. The same sequence-to-sequence model generates API-CALL, and system responses. }
    \label{fig:tod_e2e}
\end{figure}

\section{End-to-End Task-Oriented Dialogue Systems}
End-to-end task-oriented dialogue systems~\cite{eric-manning:2017:EACLshort,madotto2018mem2seq,wu2019global,reddy2018multi,yavuzdeepcopy,bordes2016learning,madotto2020learning,lin2020mintl,lin2021bitod} has been proposed to reduce the complexity of the modularized systems. Differently from module-based dialogue systems, end-to-end systems train a single model directly on text transcripts of the dialogues. This task is tackled in two ways: response  selection~\cite{bordes2016learning,perez2016gated,wu2017dstc6,dqmem,williams2017hybrid,seo2016query} and token by token generation of the response~\cite{madotto2018mem2seq,wen2016network,serban2016building,zhao2017generative,serban2017hierarchical,madotto2020exploration}. In this thesis, we especially focus on the latter, which, indeed, is very similar to Seq2Seq models used in chit-chat dialogue systems. The major difference is that task-oriented systems have to generate informative responses grounded on knowledge present in various kind of databases (e.g., knowledge graphs). A standard strategy for end-to-end dialogue systems is to first generate API-CALLs to retrieve knowledge, and then to provide knowledge-based information as input to the model. Figure~\ref{fig:tod_e2e} shows a system where the same Seq2Seq model generates both API-CALLs and system responses by using the knowledge as further input.

\section{Motivation and Research Problem}
With the emergence of powerful “deep learning” architectures, end-to-end dialogue systems end-to-end generative dialogue systems have been proposed to optimize overall system performance and simplify training. However, these systems cannot be easily controlled or extended like the modularized systems. This is because a single neural system is used, which is usually a large pre-trained language model (e.g., GPT-2), and thus it is hard to surgically change desirable attributes (e.g., style, topics, etc.). More importantly, uncontrollable dialogue systems can generate offensive or even toxic responses. 

In modularized/rule-based systems, adding new dialogue domains, for example, only requires adding rules and functions to the existing codebase, while in end-to-end models this step requires an expensive training process, namely, retrain the entire model with all the domains. Similarly, chit-chat template-based systems can easily control the generated response in terms of style and topics, but this is extremely challenging for generative conversational models.

Generally speaking, controlling dialogue systems is particularly important not only from a research perspective but also to meet an industrial need. In fact, existing smart assistants (e.g., Amazon Alexa, Apple's Siri etc.) on the market are slowly moving from modularized systems to end-to-end solutions. One of the main obstacles to prevent the wide deployment of these systems is the lack of explicit control over the generated responses. This is especially important in larger systems, where only part of the technology is actually a neural network, while the rest is hard coded rules. In such cases, the system developer requires specific behaviours from the end-to-end models (e.g., positive responses, or response from a certain domain). Therefore, having high- and low-level control over different attributes of the dialogue response is an essential skill. 

In this thesis, we study controllable methods for end-to-end generative dialogue systems.In particular, we focus on

\begin{itemize}
    \item \textbf{Controlling Style \& Topics} \citet{see2019makes} showed that being able to control the response generation can have a significant impact on the quality of conversations. However, controlled generation from large conversational models such as DialoGPT~\cite{zhang2019dialogpt}, Meena~\cite{adiwardana2020towards} and Blender-Bot~\cite{roller2020recipes}, remains a challenge, and is particularly more difficult in the absence of annotated conversational datasets. Therefore, we propose to use residual adapters~\cite{houlsby2019parameter}, which adds less than 1.5\% task-specific parameters per style/topic, to make the controllable response generation viable for online systems, and we run a comprehensive automatic and human evaluation to show that our method can control the generate responses in terms of style and topics, without losing fluency and without requiring dialogue specific datasets\cite{madotto2020plug}.
    \item \textbf{Controlling Dialogue Domain Continuously} The ability to continuously updated dialogue systems with new features based on the user's needs, e.g., adding new slots and intents, or even completely new domains, is essential for a robust and deployable dialogue systems. However, existing end-to-end dialogue models are trained with the assumption of having a fixed dataset and architecture at the beginning of the training, and they are not designed to add new domains and functionalities through time without incurring the high cost of whole-system retraining. Therefore, we propose a highly controllable architectural method based on residual adapters~\cite{houlsby2019parameter} to continuously update task-oriented dialogue systems with new features based on the user's needs. Moreover, we analyze the trade-off between performance, number of parameters, and episodic memory sizes of other existing methods (regularization, rehearsal, architectural)~\cite{madotto2020continual}.
    \item \textbf{Controlling Multi-Skill Dialogue Systems} Unlike humans who can do both, systems for goal-oriented dialogues~\citep{williams2007partially,young2013pomdp} and chit-chat conversations~\citep{serban2016generative,vinyals2015neural} are often learned with separate models, eventually rule-based. However, end-to-end dialogue models share the same Seq2Seq architecture for both chit-chat and task-oriented systems. These models greatly suffer from lack of controllability and flexibility. Therefore, we propose a novel theoretical framework to control an end-to-end dialogue model with multiple composable and controllable skills, and we empirically show the effectiveness of using specialized parameters in combined chit-chat and task-oriented datasets\cite{madotto2020attention,lin2021adapter}.
\end{itemize}

\section{Thesis Outline}
The thesis is divided in four main chapters, plus a conclusion. In Chapter 2, we introduce the background notation and methodology used throughout the thesis. In Chapter 3, we describe how to control style and topics of large generative conversational models. In Chapter 4, we describe a flexible dialogue system that is able to add conversational domains continuously. In Chapter 5, we propose a novel way to control an end-to-end dialogue model with multiple composable and controllable skills. Finally, in Chapter 6, we summarize the thesis and the significance of the controllable dialogue systems, and we discuss possible future research directions.

\begin{figure}[h]
    \centering
    \includegraphics[width=\linewidth]{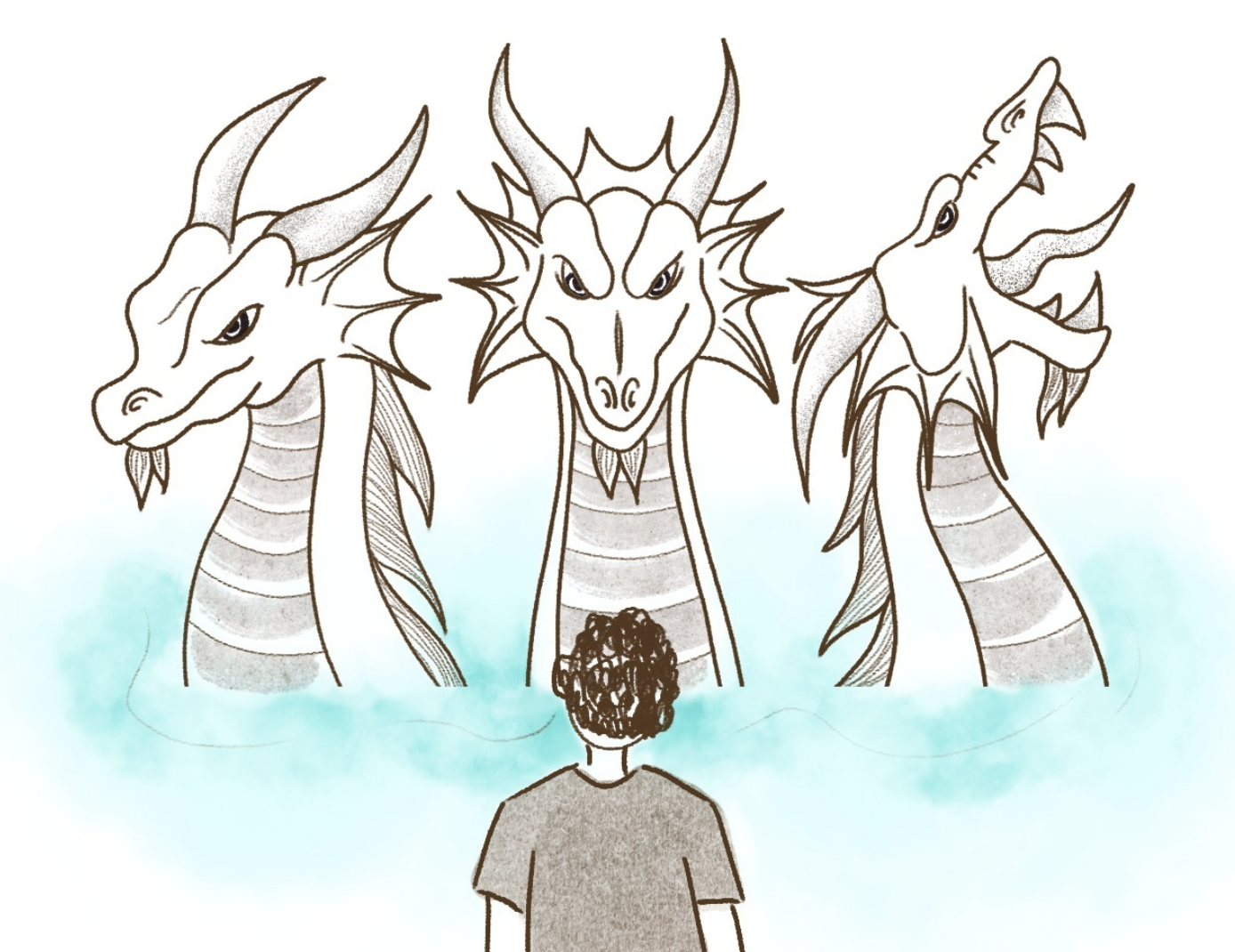}
    \caption{The three beasts to be tamed.}
\end{figure}

\chapter{Background and Preliminaries}
In this chapter, we introduce the background notation and methodology needed throughout the thesis. In particular, we focus on parametric statistical models, such as neural networks (NN), in the natural language processing (NLP) domain. 

\section{Basic notation}~\label{sec:basic_notation}
Let us defined a generic supervised dataset $D=\{ (X_i, Y_i)\}_{i=0}^n$, where $X_i$ is a feature vector and $Y_i$ its corresponding label. The dataset is split into three non-overlapping sets, namely \textit{training}, \textit{validation}, and \textit{testing}. The sets are used for training the model, estimating the performance of different hyper-parameters and estimating the generalization error respectively. 

In this thesis, we focus on two settings: classification and sequence generation. We further define these two settings by specifying the in-out pairs $(X_i, Y_i)$. In both settings, $X_i$ is a sequence of tokens $x_0, \cdots, x_n$, which represents an ordered sequence of words, e.g., sentences in a paragraph or utterances in a dialogue. Differently, $Y_i$ is a value from the set $C=\{c_0,\cdots,c_K\}$ for the classification setting and a sequence of tokens $y_0, \cdots, y_m$ for the sequence generation setting. 

Independently of the setting, we focus on discriminative models, parameterized by $\theta$, that learns the conditional probability $P_{\theta}(Y|X)$. In sequence generation, we further expand $P_{\theta}(Y|X)$ using the chain rule of probability~\cite{bengio2003neural} as
\begin{equation}~\label{eq:chainprob}
    P_{\theta}(Y|X) =  \prod_{i=0}^{m}P_{\theta}(y_i|y_{0}, \cdots, y_{i-1}, x_{0}, \cdots, x_{n}),
\end{equation}
In the coming sections, we introduce model instances for $\theta$, in both classification and sequence generation, and the loss functions used to train them. 

Before diving into the modelling, let us define a vocabulary of words $V$ as a set of unique words appearing in a large text corpus. The token $x\in X$ is converted into its 1-hot representation denoted as $\hat{x} \in \mathbb{N}^{1 \times |V|}$, where $|V|$ is the cardinality of the vocabulary. In $\hat{x}$, only the element corresponding to the token $x$ is set to 1 and otherwise to 0. Following this notation, the input becomes a matrix $X\in \mathbb{R}^{n\times |V|}$, where each row is a vector of the vocabulary size dimension.  

\begin{figure}[t]
    \centering
    \includegraphics[width=\linewidth]{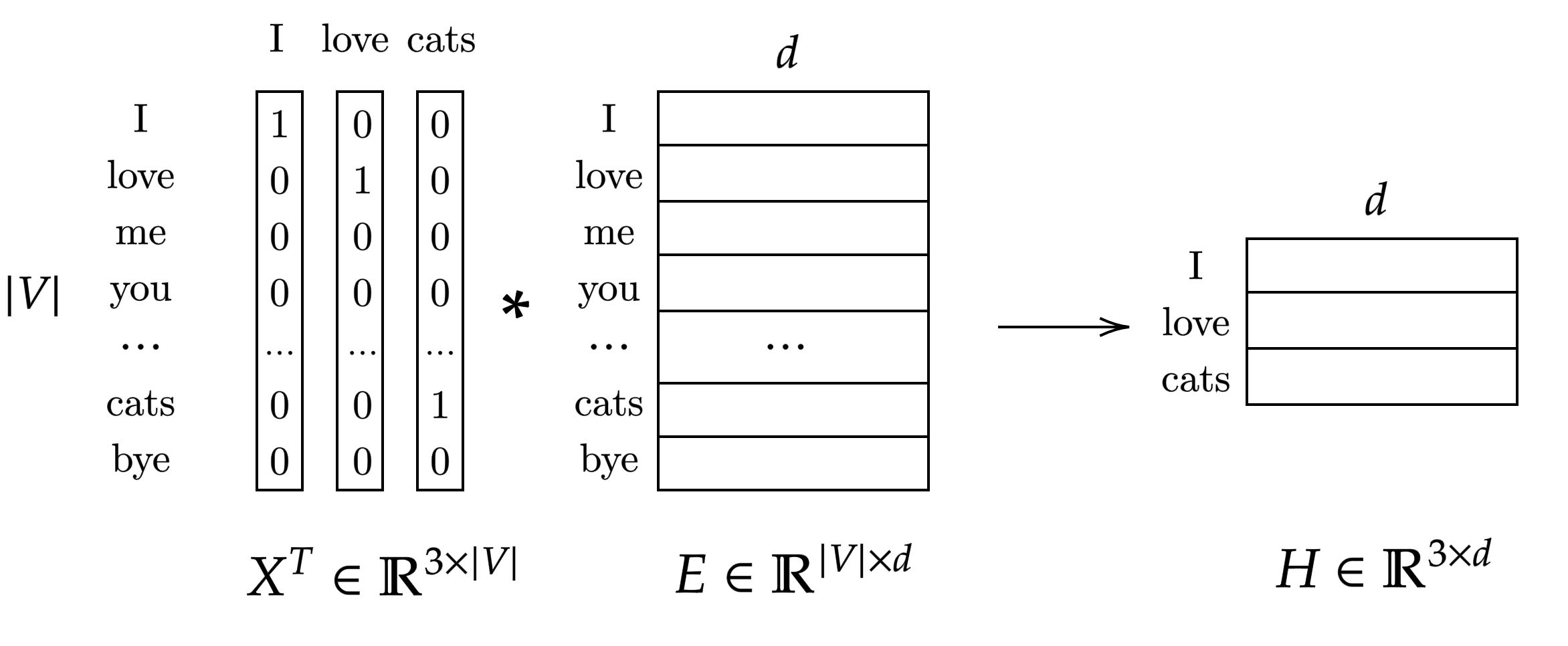}
    \caption{Example of sentence encoding, from 1-hot encoding to word embedding conversion. Firstly, the words in the sentence are converted into their 1-hot representation $\hat{x}$. Then, by multiplying the concatenation of the 1-hot of each words by the embedding matrix $E$, we obtain the word embedding for each word.  }
    \label{fig:word_encoding}
\end{figure}

\section{Word Embedding}\label{sec:word_embedding}
The basic block of any NLP-based neural network is the embedding matrix, which maps the input tokes into their embedded representation. Let us define the word embedding matrix~\cite{mikolov2013distributed,xu2018emo2vec} $E\in \mathbb{R}^{|V|\times d}$, where $d$ is the embedding size. The embedding matrix $E$ is multiplied by the input sequence $X$~\footnote{We remove the index $i$ to improve readability.} to obtain its embedded representation. We denote this transformation as
\begin{equation}
    H = X^T*E,
\end{equation}
where $H\in \mathbb{R}^{n \times d}$ is the resulting embedding for each of the tokens in the input sequence. In most neural network libraries (e.g., PyTorch~\cite{NEURIPS2019_9015}, TensorFlow~\cite{tensorflow2015-whitepaper}) this operation is efficiently done using hashing to select the row of interest $E$. In this case, the input matrix $X$ is transformed back to a vector in which each position is the index of the element 1 in the row. Figure~\ref{fig:word_encoding} describes an example of transforming a sentence into its embedded representation. Using this notation, the embedding operation is denoted as $H=E(X)$. In the rest of the thesis, we use these two notations interchangeably.

\begin{figure}[t]

  \begin{minipage}[c]{0.57\textwidth}
       \centering 
     \includegraphics[width=0.8\textwidth]{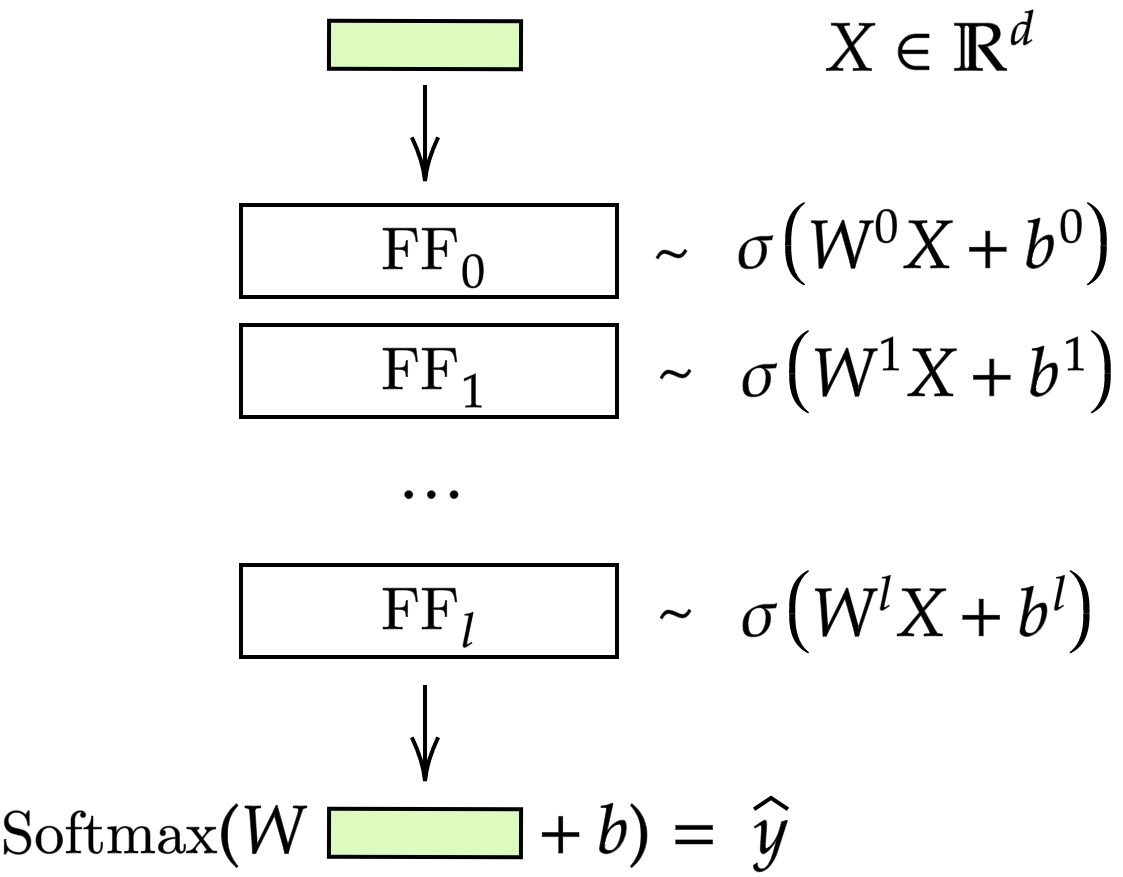}
  \end{minipage}
  \begin{minipage}[c]{0.4\textwidth}
    \caption{Example of feed-forward (FF) neural network. The input $X$ is passed through multiple feed-forward layers: an affine transformation with a non-linear activation -- sigmoid in this example -- and in the last layer a final Softmax activation (Equation~\ref{eq:softmax}) to generate the final output $\hat{y}$.}
    \label{fig:ff}
  \end{minipage}

\end{figure}
\section{Feed-Forward Neural Network}\label{sec:feed_forward}
A feed-forward (FF) neural networks is a parametric model made of affine transformations and non-linear activations. More formally, given an input vector $x\in \mathbb{R}^{d \times 1}$, FF networks with $l$ layers compute the function
\begin{equation}
   \textrm{FF}(x) = f(W^l  f(W^{l-1} \ \dots \ f(W^{0} x + b^{0})  + b^{l-1})  + b^l), \label{eq:ff}
\end{equation}
where each $W^l \in \mathbb{R}^{d \times d}$ and $b^l \in \mathbb{R}^{d \times 1}$ are trainable parameters, and $f$ is an activation function, e.g., $f(x)=\sigma(x)=\frac{1}{1+e^{-x}}$. Without loss of generality, Equation~\ref{eq:ff} describes a FF neural network in which each transformation $W$ has the same size (i.e., $d \times d$), and uses the same activation function (e.g., sigmoidal). In general, the dimensions of the transformation are arbitrary and different activation functions can be used (e.g., ReLU). 

Finally, to model the conditional probability $P_{\theta}(Y|X)$, the output of $\textrm{FF}(x)$ is projected to the classification output space $Y$ using a linear transformation and a Softmax function,
\begin{equation}
    \mathrm{Softmax} (\mathbf {h} )_{i}={\frac {e^{h_{i}}}{\sum _{j=1}^{K}e^{h_{j}}}} \qquad \forall i=1,\dotsc ,K \label{eq:softmax}
\end{equation}
where $\mathbf {h} =(h_{1},\dotsc ,h_{K})\in \mathbb {R}^{1\times d}$. Therefore, we denote the model prediction for the class $i$ as
\begin{equation}\label{eq:outlayer}
    \hat{y}_i = \mathrm{Softmax} (W \textrm{FF}(x) + b)_{i},
\end{equation}
where $W \in \mathbb{R}^{K \times d}$ and $b \in \mathbb{R}^{1 \times d}$. Furthermore, we denote the overall model's parameters with the set $\theta=[W^0,b^0, W^1, b^1, \cdots,W^l, b^l, W, b]$. These parameters are optimized by minimizing the cross-entropy loss between the predicted $\hat{\textbf{y}}$ and the gold $y$ from the dataset $D$. Formally, the loss is computed as
\begin{equation} \label{eq:NLL}
    L(D) = - \sum_j^{\mathcal{D}} y_j \log (\hat{y_j}).
\end{equation}
In this equation, we purposely imply that this operation is vectorial. Thus we omit the further summation over the number of classes. The loss function is minimized using the gradient descent algorithm (i.e., backpropagation~\cite{rumelhart1986learning}). A high-level description of a FF neural network is shown in Figure~\ref{fig:ff}.

\begin{figure}[t]
    \centering
    \includegraphics[width=\linewidth]{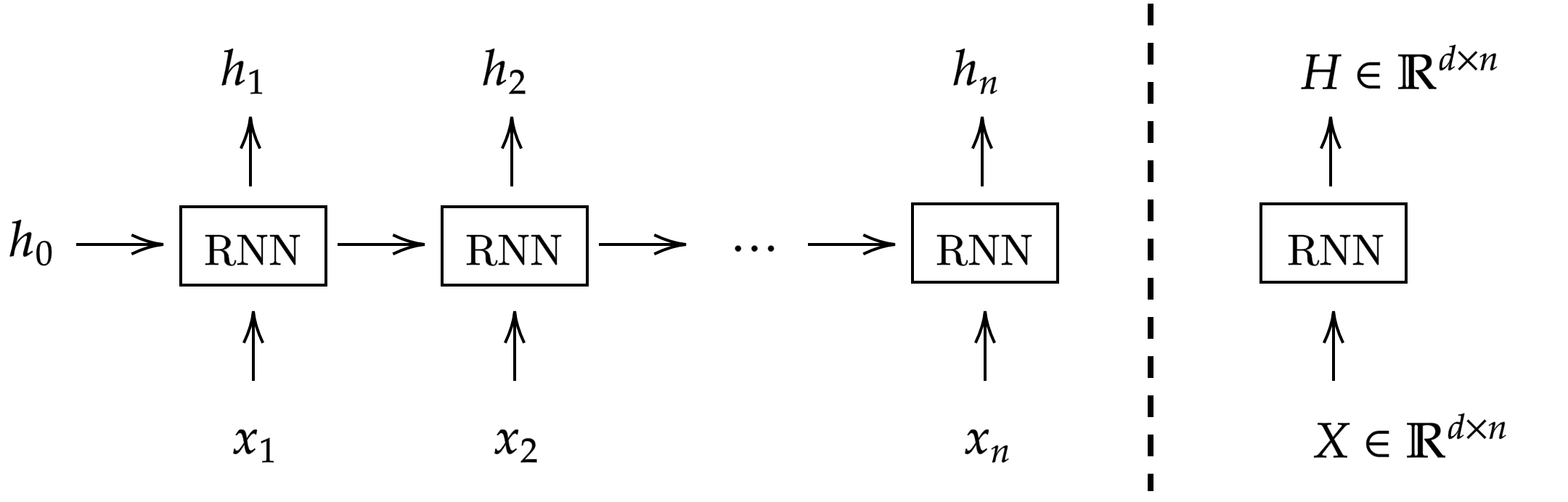}
    \caption{Example of recurrent neural network. }
    \label{fig:RNN}
\end{figure}
\section{Recurrent Neural Networks}\label{sec:RNN}
Differently from FF neural networks, recurrent neural networks (RNNs) are used to process a sequence of inputs, e.g., a sequence of words, and they are able to capture temporal dependency in the input. 

Given the input sequence $X=\{x_1, \cdots, x_n\}$, where each $x_t \in \mathbb{R}^{d \times 1}$, an RNN processes one input at a time by updating an hidden state vector $h\mathbb{R}^{d \times 1}$. The RNN forward function for an input at time step $t$ is expressed as:
\begin{align}\label{eq:RNN}
    h_{t} &= \sigma(W x_t + U h_{t-1} +  b),
\end{align}
where $U \in \mathbb{R}^{z \times z}$, $W \in \mathbb{R}^{z \times d}$, $b \in \mathbb{R}^{z \times 1}$ and $h_0 = \vec{0}$. Notice that $W^h,W^i$ and $b^i$ are shared parameters across time steps. We denote the set of hidden states at each time step as $H=[h_0,\cdots, h_n]$. These hidden states can be used to represent the temporal dependency between "input features", e.g., the word sequence in a sentence. Figure~\ref{fig:RNN} shows a high-level description of the model. The RNN described in Equation~\ref{eq:RNN} suffers from the gradient vanishing and explosion problem~\cite{pascanu2013difficulty}. To cope with this issue, two variants of this simple architecture have been proposed: Long-Term Short Term Memory (LSTM) and Gated Recurrent Unit (GRU).

\paragraph{The LSTM}~\cite{hochreiter1997long} uses three gates (input gate, forget gate and output gate) to modulate the amount of information to be stored in the hidden memories (the hidden state and context state). Given the input sequence $X=\{x_0, \cdots, x_n\}$,  where each $x_t \in \mathbb{R}^{d \times 1}$, the LSTM for an input at time step $t$ computes
\begin{equation}\label{eq:LSTM}
\begin{aligned} 
    f_{t} &=\sigma\left(W^{f} x_{t}+U^{f} h_{t-1}+b^{f}\right)\\
    i_{t} &=\sigma\left(W^{i} x_{t}+U^{i} h_{t-1}+b^{i}\right)\\ 
    o_{t} &=\sigma\left(W^{o} x_{t}+U^{o} h_{t-1}+b_{o}\right) \\ 
    \tilde{c}_{t} &=\tanh \left(W_{c} x_{t}+U_{c} h_{t-1}+b_{c}\right) \\ 
    c_{t} &=f_{t} \odot c_{t-1}+i_{t} \odot \tilde{c}_{t} \\ 
    h_{t} &=o_{t} \odot \sigma\left(c_{t}\right),
\end{aligned}
\end{equation}
where $W^f,W^i,W^0 \in \mathbb{R}^{d \times z}$, $U^f,U^i,U^0 \in \mathbb{R}^{z \times z}$, $b^f,b^i,b^0 \in \mathbb{R}^{z \times 1}$, and $\tanh=\frac{2}{1+e^{-2x}}-1$. The $\odot$ denotes the Hadamard product (i.e. element-wise matrix multiplication). The vector $h_t$ and $c_t$ are usually called the hidden state and the context state. The LSTM reduces the gradient vanishing problem greatly, since the context state does not use a sigmoidal ($\sigma$) activation.  

\paragraph{The GRU}~\cite{cho2014learning}, similarly to the LSTM, modulates how much information has to be stored in the hidden memory. However, it greatly simplifies the models' architecture and it uses only one hidden state, like in the original RNN. Given the input sequence $X=\{x_0, \cdots, x_n\}$,  where each $x_t \in \mathbb{R}^{d \times 1}$, the GRU for an input at time step $t$ computes
\begin{equation}\label{eq:GRU}
\begin{aligned} 
z_{t} &=\sigma\left(W^{k} x_{t}+U^{k} h_{t-1}+b^{k}\right) \\ 
r_{t} &=\sigma\left(W^{r} x_{t}+U^{r} h_{t-1}+b^{r}\right) \\ \hat{h}_{t} &=\tanh\left(W^{h} x_{t}+U^{h}\left(r_{t} \odot h_{t-1}\right)+b^{h}\right) \\ 
h_{t} &=\left(1-z_{t}\right) \odot h_{t-1}+z_{t} \odot \hat{h}_{t}, \end{aligned}
\end{equation}
where $W^k,W^r,W^h \in \mathbb{R}^{d \times z}$, $U^k,U^r,U^h \in \mathbb{R}^{z \times z}$, $b^k,b^r,b^h \in \mathbb{R}^{z \times 1}$, and $\tanh=\frac{2}{1+e^{-2x}}-1$. Differently from the LSTM, the GRU architecture uses one hidden state $h$, but a similar gating mechanism for avoiding gradient vanishing. 

\paragraph{Notation:} To improve readability, we denote the Recurrent Equations~\ref{eq:RNN},~\ref{eq:LSTM} and ~\ref{eq:GRU} using the following compact notations:
\begin{align}
    h_t &= RNN(x_t, h_{t-1})\\
    (c_t, h_t) &= LSTM(x_t, (h_{t-1},c_{t-1}))\\
      h_t &= GRU(x_t, h_{t-1}).
\end{align}
Finally, we denote the set of hidden states for the sequence $X=\{x_0, \cdots, x_n\}$ as the concatenation of the corresponding hidden states at each time step, i.e., $H=[h_0,\cdots,h_n]$. 

\paragraph{Applications:} RNNs are often used in three settings: classification, sequence tagging, and sequence-to-sequence generation. In the classification setting, the model is trained using the same loss function as in Equation~\ref{eq:NLL}, where the FF network is replaced by an RNN and the last hidden  state $h_t$ is used for the final linear transformation in Equation~\ref{eq:outlayer}. Differently, in the sequence classification and the sequence-to-sequence settings, the RNN is trained to make a sequence of classifications. In this thesis, we focus on the sequence-to-sequence setting. Thus, interested readers can refer to ~\citet{goodfellow2016deep} for more information about sequence classification tasks.  


\begin{figure}[t]
    \centering
    \includegraphics[width=0.8\linewidth]{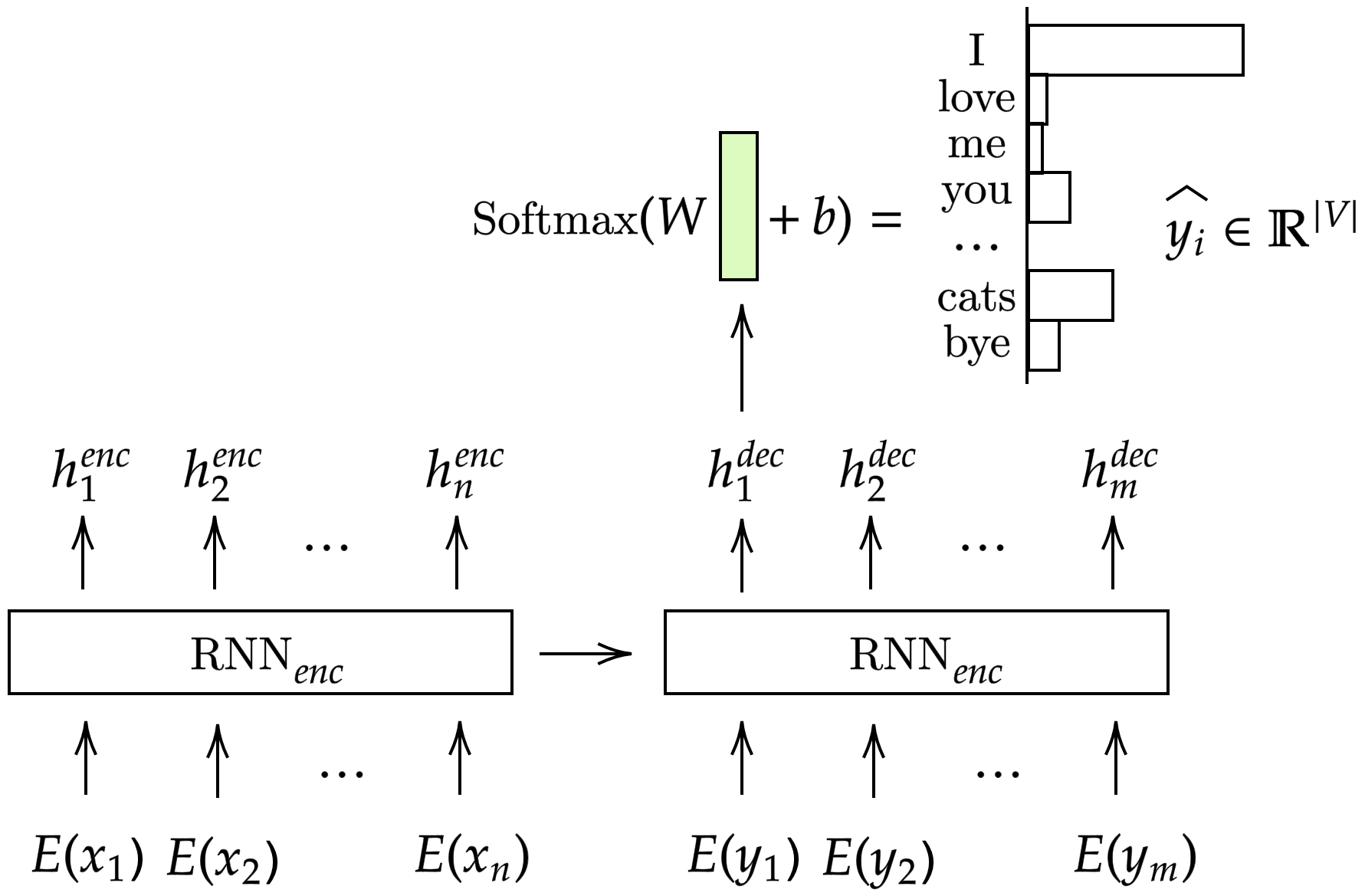}
    \caption{Example of RNN-based sequence-to-sequence model. }
    \label{fig:seq2seq}
\end{figure}
\section{Sequence-to-Sequence} \label{sec:Seq2Seq}
Sequence-to-sequence (Seq2Seq)~\cite{sutskever2014sequence} is an encoder-decoder architecture used to process sequences in both the input and the output. In this section, we describe an RNN-based Seq2Seq model for language generation, thus leveraging the RNN in Section~\ref{sec:RNN} and the word embedding in Section~\ref{sec:word_embedding}. 

The objective of a Seq2Seq model is to approximate the conditional probability $P(Y|X)$, where both $X=\{x_1, \cdots, x_n\}$ and $Y=\{y_1, \cdots, y_m\}$ are sequences, e.g., sequences of words in a sentence. Formally, a Seq2Seq model generates a sequence of probability distributions, each of which is conditioned on the previously generated tokens and the input sequence $X$ (Equation~\ref{eq:chainprob}).

The Encoder gets the tokens in $X$, computes their embedded representation using the embedding matrix $E$ and passes it through an RNN$_{enc}$.~\footnote{We use the simple RNN, but the same explanation holds for both the LSTM and GRU.} Hence, given $h_0^{enc} = \vec{0}\in \mathbb{R}^{d \times 1}$, each token in $X$ is processed by
\begin{equation}
    h_{t}^{enc} = \text{RNN}_{enc}(E(x_t), h_{t-1}^{enc}).
\end{equation}
The last hidden state $h_{|X|}^{enc}$ is used as the initial hidden state of the decoder. The decoder, which is another RNN, generates token by token the output sequence $Y$. Hence, given $h_0^{dec}= h_{|X|}^{enc}$ as the initial hidden state, the decoder computes
\begin{align}
    h_{t}^{dec} &= \text{RNN}_{dec}(E(y_t), h_{t-1}^{dec}) \\
    p(y_t=\hat{y}_t|X,y_{0}  \cdots, y_{t-1}) = \hat{y}_t &= \mathrm{Softmax}(W h_{t}^{dec}), \label{eq:final_transform}
\end{align}
where $W\in \mathbb{R}^{d \times |V|}$ is a linear transformation, similar to FF, that maps the hidden state into a vector of vocabulary length size. Then, the Softmax activation normalizes the vector to obtain a probability distribution over the vocabulary $\hat{y}_t$. Figure~\ref{fig:seq2seq} shows a high-level description of an RNN-based Seq2Seq model. Given a dataset of input-output pairs $D=\{(X_i,Y_j)\}_i$, the parameter of the model are optimized by minimizing:
\begin{equation}
    L(D) =  - \sum_j^{|D|} \sum_{i=0}^{m} \log p(y_i^{(j)}|x_{0}^{(j)}, \cdots, x_{n}^{(j)},y_{0}^{(j)}, \cdots, y_{i-1}^{(j)}),\label{eq:loss_seq2seq}
\end{equation}
where $y_{ij}$ is the $i$-th word in the $j$-th example. As in Equation~\ref{eq:NLL}, this loss is minimized using gradient descent. During testing, the model produces the output sequence in an auto-regressive manner~\cite{graves2013generating}, i.e., providing the generated token as input of the next generation time-step. Seq2Seq is a powerful generative model, but it still struggles in capturing long-term dependency between the generated token and the input sequence. To cope with this issue, the attention mechanism~\cite{bahdanau2014neural,luong2015effective} has been proposed. 

\begin{figure}[t]
    \centering
    \includegraphics[width=0.8\linewidth]{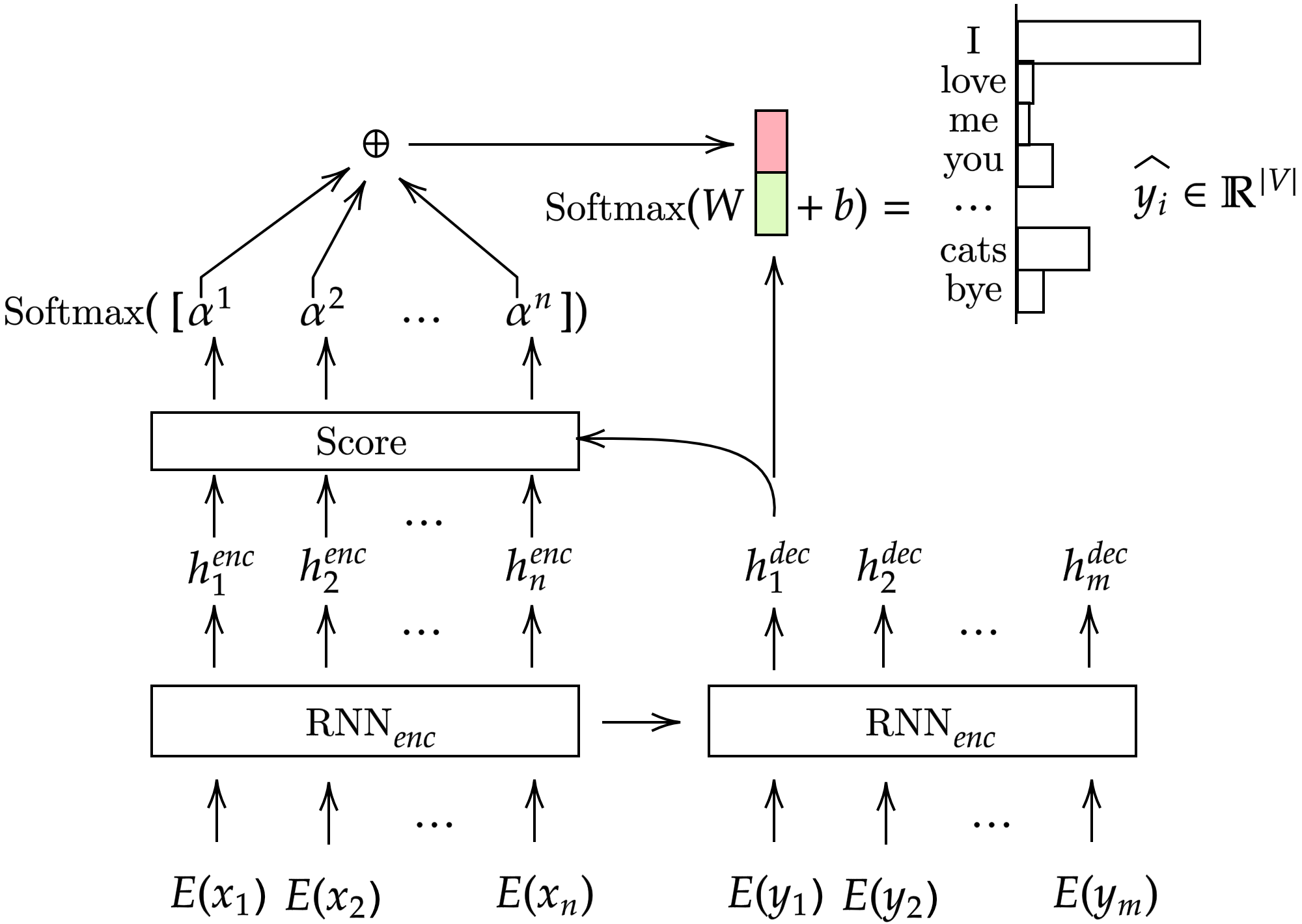}
    \caption{Example of RNN-based sequence-to-sequence model with the attention mechanism. }
    \label{fig:seq2seqatt}
\end{figure}
\subsection{Attention} \label{subsec:attention}
Attention is used to learn an alignment between the generated tokens $Y$ and the input $X$. In practice, this is done by scoring the current hidden state of the decoder with all the encoder hidden states. Hence, given the encoder hidden state as $H=\{h_{0}^{enc},\cdots,h_{|X|}^{enc}\}$ and the decoder hidden state at time step $t$ as $h_{t}^{dec}$, the attention module computes the context vector as:
\begin{align}
    \alpha_t^i  &= \mathtt{Score}(h_{t}^{dec}, h_{i}^{enc}) \ \ \forall i \in 1,\cdots,|X|\\ \label{eq:score}
    \alpha_t    &= \mathrm{Softmax}(\alpha_t) \ \ \forall i \in 1,\cdots,|X| \\ \label{eq:soft}
    c_t         &= \sum_k^{|X|} h_{k}^{enc} * \alpha_t^k\\ \label{eq:context} 
    \hat{y}_t   &= \mathrm{Softmax}(W [h_{t}^{dec};c_t]),
\end{align}
where $[;]$ is the concatenation of the two vectors, $\alpha$ is the attention vector, and \texttt{Score} is a function chosen among three options: dot product, bi-linear, and neural networks. These are formally as:
\begin{equation}
    \mathrm{Score}(h_{i}, h_{j}) = \Bigg\{
  \begin{array}{cl}
    h_{i}h_{j}^T & \text{(dot)} \\
    h_{i} W h_{j}& \text{(bi-linear)} \\
    \text{tanh}(W [h_{i};h_{j}]) & \text{(neural network)}, \\
  \end{array}
\end{equation}
Figure~\ref{fig:seq2seqatt} shows a Seq2Seq model with attention over one decoding step. The attention mechanisms is extremely important for the success of Seq2Seq architectures, up to the point of removing the RNN in favor of a (self) attention based mechanism~\cite{vaswani2017attention}. 

\section{Transformer}~\label{sec:transformer}
As described in Section~\ref{sec:Seq2Seq}, RNNs are the basic block for both the encoder and the decoder in seq2seq models, and the attention mechanism (Section~\ref{subsec:attention}) plays a key roles in the success of the overall architecture. However, RNNs, including LSTMs and GRUs, requires a non-parallelizable operation for computing the hidden states of the sequence, which greatly limits the speed of the model. To cope with this temporal dependency issue and to fully exploit the attention mechanism, \citet{vaswani2017attention} proposed an RNN-free model, the Transformer. 

Similar to RNN-based Seq2Seq, the Transformer is made of 1) an encoder, 2) a decoder, 3) an embedding matrix, and 4) a positional embedding matrix. The latter is particularly important for capturing the temporal dependency in the input. Differently to Seq2Seq, however, both encoder and decoder, shares a generalized attention mechanism and FF networks only. 

\subsection{Generalized Attention}
Given the input sequence $Z=\{z_0, \cdots, z_n\}\in \mathbb{R}^{d\times n}$, where each $z_t \in \mathbb{R}^{d \times 1}$, the generalized attention module computes:

\begin{equation}
\vcenter{\hbox{\begin{minipage}{5cm}
\centering
\includegraphics[width=6cm,height=6cm]{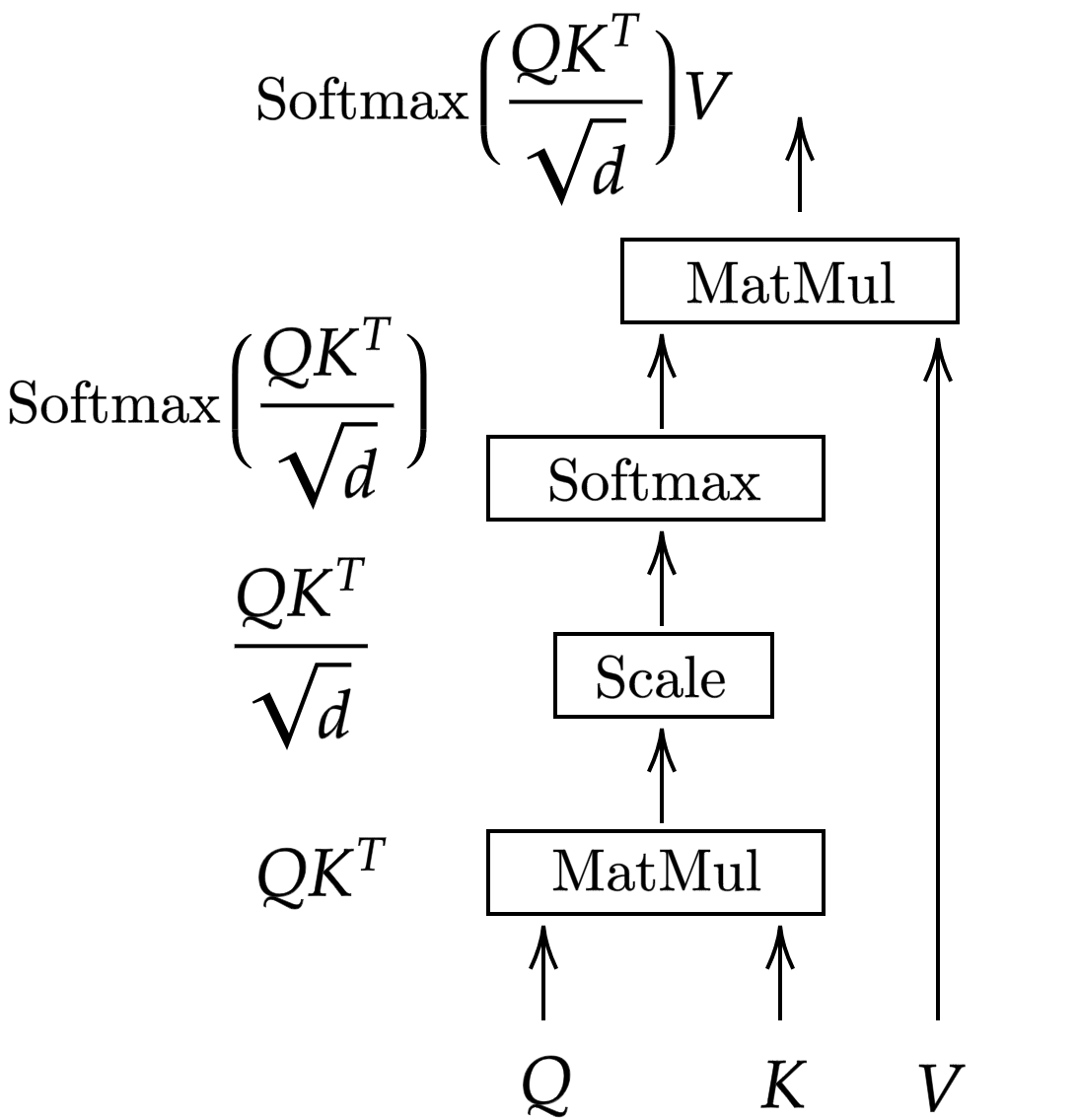}
\end{minipage}}}
\qquad\qquad
\begin{aligned}
\textrm{Attention}(Q, K, V) &=\mathrm{Softmax}\left(\frac{Q K^{T}}{\sqrt{d}}\right) V, ~\label{eq:attention}
\end{aligned}
\end{equation}

where $Q=K=V=Z$. This is a generalization of the attention process in Equation~\ref{eq:score},~\ref{eq:soft} and ~\ref{eq:context}, where instead of considering each single hidden state and loop over it, we put all the hidden states into a matrix and compute all the attention scores in a single operation. The output of the attention module is indeed a matrix of the same size as the input $Z$. 

\citet{vaswani2017attention} also introduce a multi-head attention mechanism, where projection matrices ($W^Q,W^K$ and $W^V$) are used to split $Q,K$ and $V$ into a sub-matrix and the attention is run in parallel on this transformation. More formally, a multi-head attention module with $h$ heads computes:
\begin{equation}
\begin{aligned} \textrm{MultiHead}(Q, K, V) &=\text { Concat }\left(\text { head }_{1}, \ldots, \text { head }_{\mathrm{h}}\right) W^{O} \\ \text { where head }_{\mathrm{i}} &=\textrm{Attention}\left(Q W_{i}^{Q}, K W_{i}^{K}, V W_{i}^{V}\right), \end{aligned}
\end{equation}
where the projections are parameter matrices $W_i^Q\in \mathbb{R}^{d_{m} \times d}$, $W_i^K\in \mathbb{R}^{d_{m} \times d}$, $W_i^V\in \mathbb{R}^{d_{m} \times d}$ and $W^O\in \mathbb{R}^{hd \times d_m}$. 

\subsection{Embedding Transformation}
Similar to Seq2Seq models, "the first step of the Transformer is to convert the tokens in the input sequence into their corresponding embeddings. Thus, given the input sequence $X=x_1, \cdots, x_n$, we define the embedding matrix $E$, which maps tokens to vectors of dimension $d \times 1$. 

Differently from RNNs, the Transformer does not use any recurrent states and thus by construction it is not able to model temporal dependency in the input sequence. To cope with this issue, \citet{vaswani2017attention} propose a sinusoidal positional embedding. The positional embedding matrix $PE \in \mathbb{R}^{\mathrm{max\_len} \times d}$ is made of sine and cosine functions of different frequencies depending on the position. More formally, each element in $PE$ is defined as
\begin{equation}
\begin{aligned} PE_{(\mathrm{pos}, 2 i)} &=\sin \left(\mathrm{pos} / 10000^{2 i / d}\right) \\ PE_{(\mathrm{pos}, 2 i+1)} &=\cos \left(\mathrm{pos} / 10000^{2 i / d}\right), \end{aligned}
\end{equation}
where pos is the position in the sequence and $i$ is the $i$-th position in the embedding dimension. Therefore, given the input sequence $X$, its embedded representation is defined as
\begin{equation}
\vcenter{\hbox{\begin{minipage}{5cm}
\centering
\includegraphics[width=\textwidth]{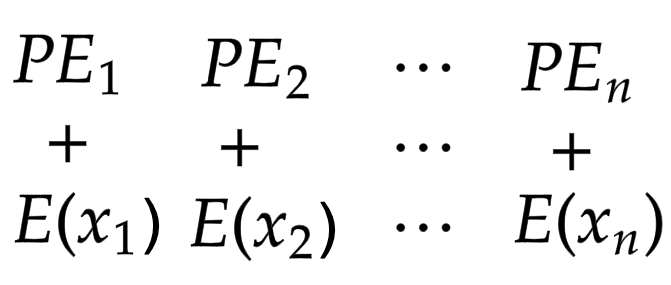}
\end{minipage}}}
\qquad\qquad
\begin{aligned}
H = E(X)+PE(X)
\end{aligned}
\end{equation}
where $PE_i$ is the $i$-th position in the embedding dimension, and $H\in\mathbb{R}^{n\times d}$ is the resulting embedded representation of the input. 

\subsection{Encoder}
The Transformer encoder is made of a stack of encoder layers, each of which is made of a multi-head attention module, a FF layer, and two layer normalization modules. Each layer gets the embedding $H$, which at the first layer is simply the word embedding of the input, and it returns a transformed version of this embedding matrix by computing 
\begin{equation}
    \includegraphics[width=0.7\textwidth]{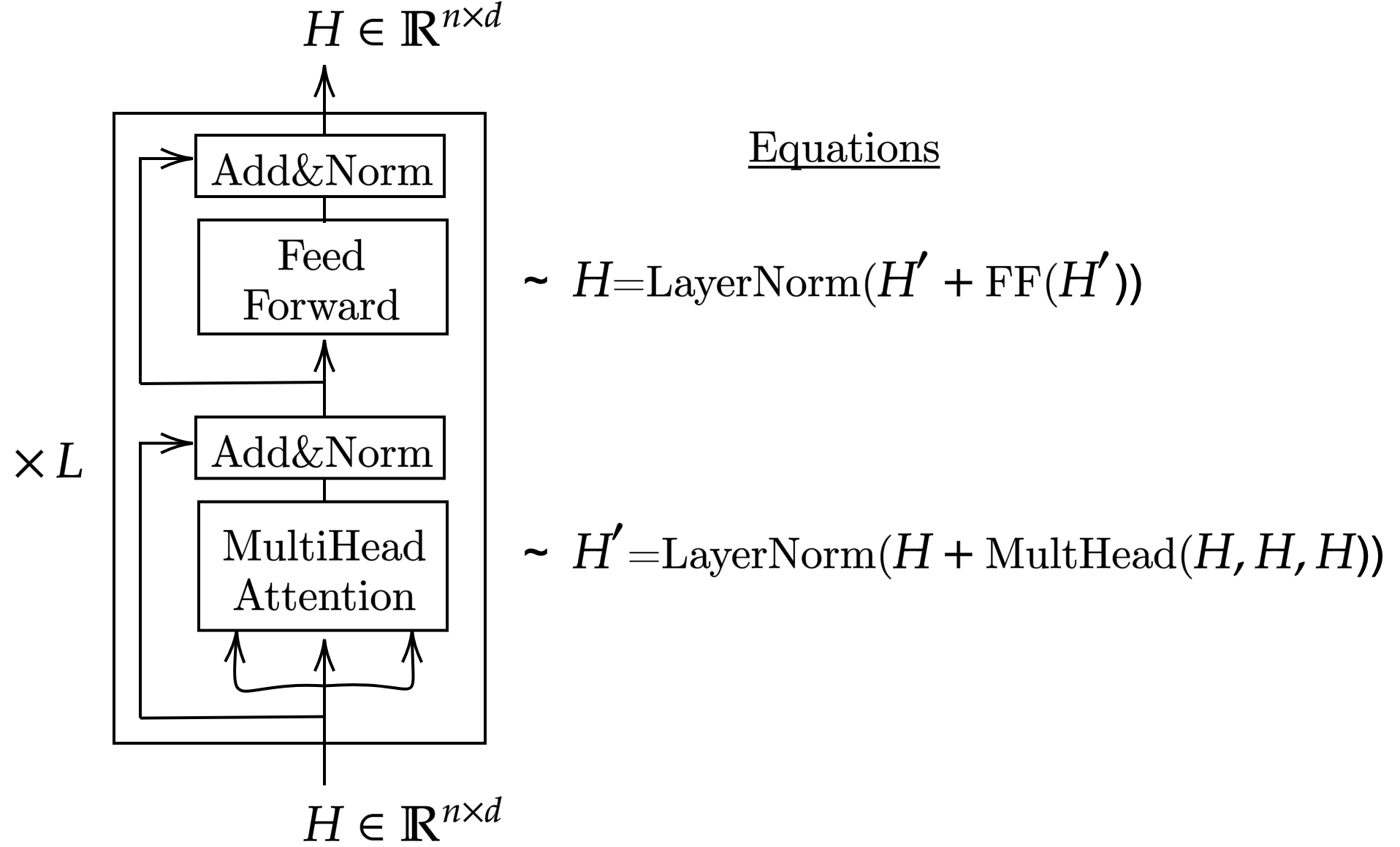}\label{eq:encoder}
\end{equation}
where $\textrm{FF}(x)= \max \left(0, W^{1}x+b^{1}\right) W^{2}+b^{2}$ is a FF neural network with two layers, and ReLU activation. The multi-head attention step, in which $Q=K=V$, is often denoted as the self-attention layer. In short, the stack of $l$ encoder layers form a Transformer encoder $TRS_{enc}$, and the forward function is denoted as
\begin{equation}
    H = TRS_{enc}(E(X)),
\end{equation}
where the embedding function $E$ includes the positional encoder. 

\subsection{Decoder}
\begin{figure}[t]
    \centering
    \includegraphics[width=0.9\linewidth]{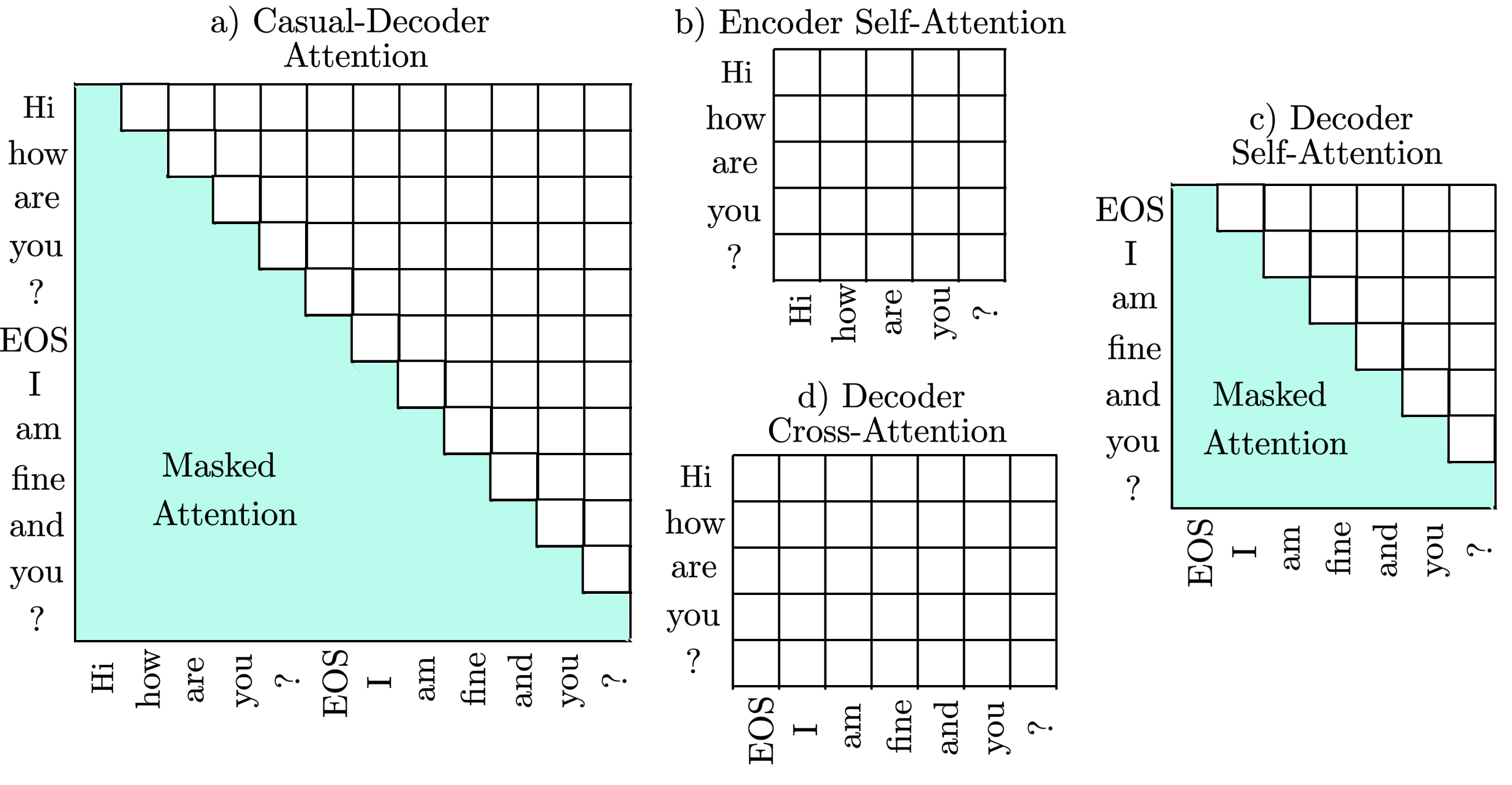}
    \caption{Different attention masks. a) The input and output sequences are concatenated, and the attention at each layer is masked to make the model unidirectional. b) The encoder is bi-directional; thus there is no mask in the attention. c) The decoder is uni-directional; thus it also has a mask to hide future tokens. d) The cross-attention between the encoder and decoder is also bi-directional; thus no mask is required.}
    \label{fig:attentionmask}
\end{figure}

\begin{figure}[t]
    \centering
    \includegraphics[width=0.9\linewidth]{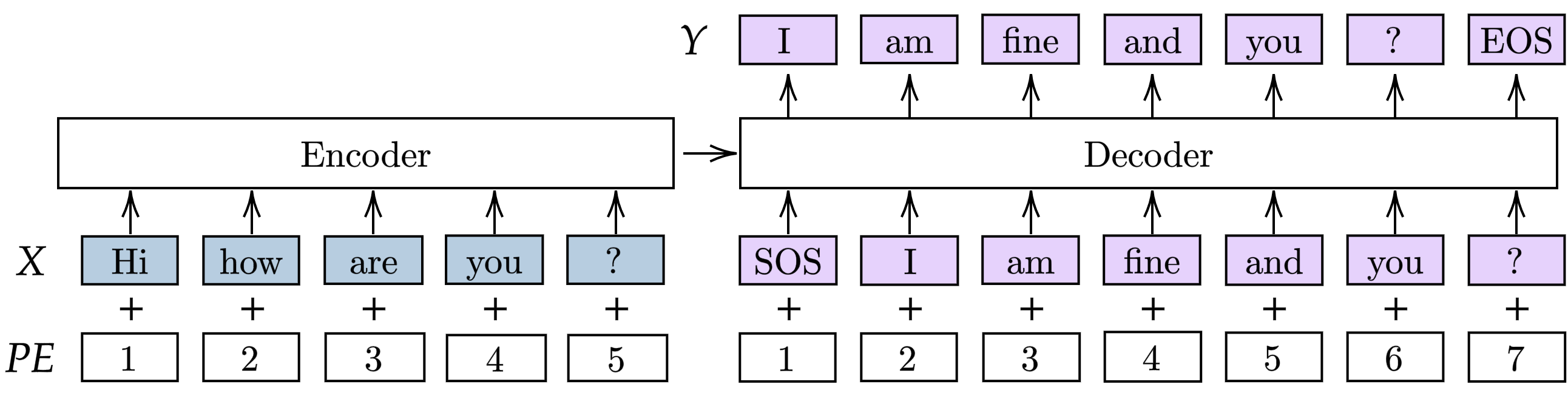}
    \caption{Example of encoder-decoder Transformers. This model is equivalent to any Seq2Seq architecture.}
    \label{fig:encdec}
\end{figure}

The Transformer decoder is composed of a stack of decoder layers, each of which is made of two multi-head modules with their corresponding layer normalization, and a FF neural network. Each layer gets the embedding $T$, the last layer embedding from the encoder, denoted as $H$, and returns a transformed version of this embedding matrix by computing
\begin{equation}
        \includegraphics[width=0.7\textwidth]{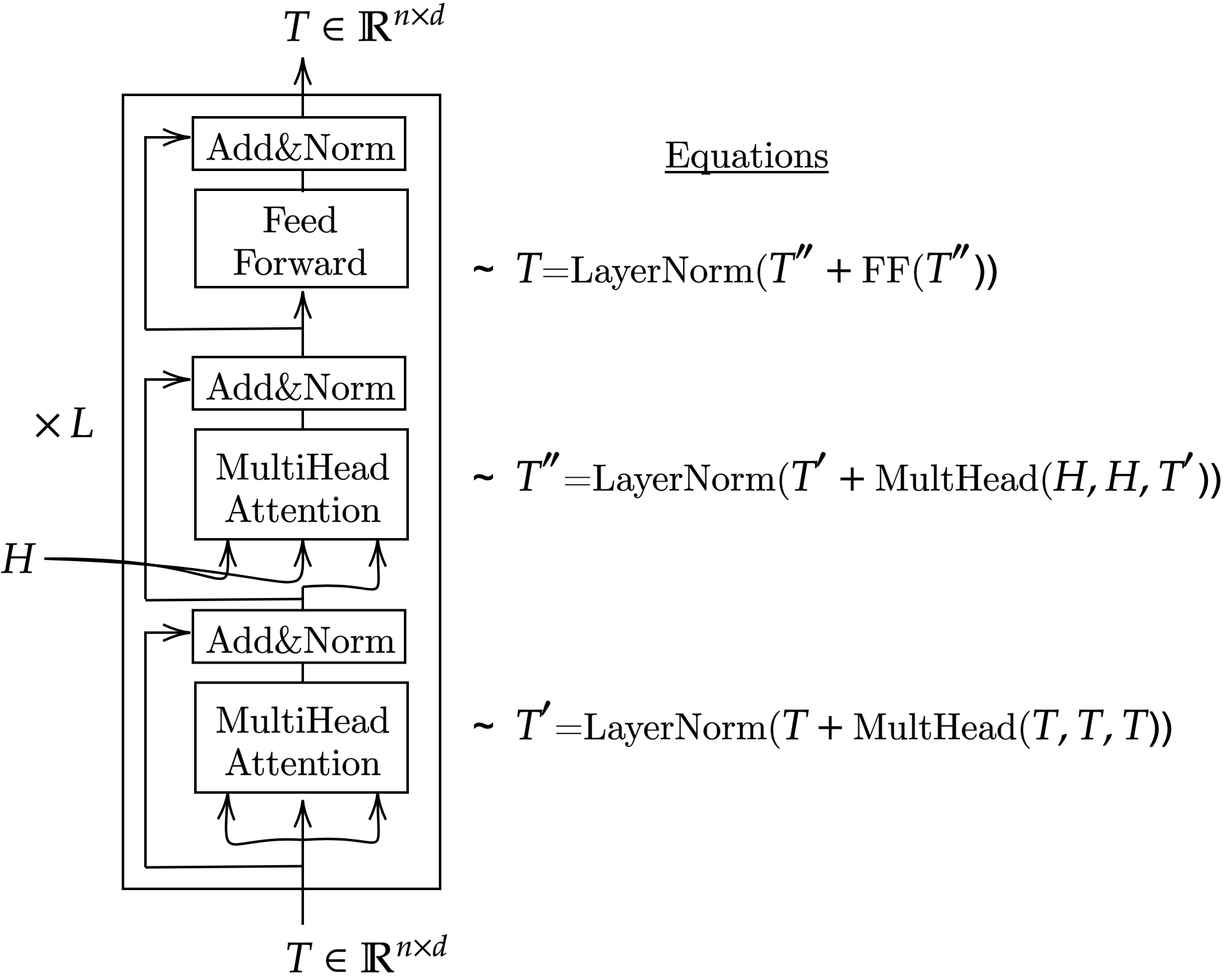}
\end{equation}
The first multi-head attention module is further masked to avoid the current tokens to attend to the future once. A summary of different attention masks is shown in Figure~\ref{fig:attentionmask}. In short, the stack of $l$ decoder layers forms a Transformer decoder $TRS_{dec}$, and the forward function is denoted as
\begin{equation}
    T = TRS_{dec}(E(y_t),H),
\end{equation}
where $T\in\mathbb{R}^{n\times d}$. The $t$-th vector of $T$, denoted as $T_t$, is then used to generate a distribution over the vocabulary $V$ by computing
\begin{equation}
    \hat{y}_t  = \mathrm{Softmax}(W T_t),
\end{equation}
where $W\in \mathbb{R}^{d \times |V|}$. This equation is equivalent to Equation~\ref{eq:final_transform}, and thus we use the same cross-entropy loss function as in Equation~\ref{eq:loss_seq2seq}. This last transformation is often named the language model head. Finally, a high-level representation of the Transformer encoder-decoder architecture is shown in Figure~\ref{fig:encdec}. 

\section{Pre-trained Language Models}
The Transformer~\cite{vaswani2017attention} has enabled large-scale language models (LMs) trained on a huge amount of data~\citep{radford2018improving,radford2019language,devlin2018bert,dai2019transformer} to greatly improve the state-of-the-art on many downstream tasks in natural language processing. In general, pre-trained models are divided into bi-directional~\cite{devlin2018bert,liu2019roberta}, uni-directional or casual-decoder~\citep{radford2018improving,radford2019language,gpt3,dai2019transformer}, and encoder-decoder generative~\cite{raffel2020exploring,lewis2020bart}. The bi-directional pre-trained models are trained with a masked-language model (MLM) loss, which learns how to predict words that are randomly masked in the input. These models achieve state-of-the-art performance in complex natural language understanding tasks~\cite{wang2018glue}. On the other hand, the uni-directional and encoder-decoder generative models are usually trained using the likelihood function in Equation~\ref{eq:chainprob}. In this thesis, we mostly focus on generative models, and thus we provide further details about the latter two pre-training strategies.

\subsection{Causal-Decoder}
The causal-decoder is a special Transformer-based architecture in which the encoder and decoder are merged into a single set of parameters. In this architecture, the input and output sequence is concatenated and it is provided as input to a Transformer encoder $TRS$, as defined in Equation~\ref{eq:encoder}. However, the attention matrices (i.e., the result of $QK^T$ in Equation~\ref{eq:attention}) in this encoder are masked, as in the first self-attention step of the Transformer decoder. Figure~\ref{fig:attentionmask}(a) shows the attention masking for a causal decoder architecture. More formally, given a sequence $X= \{x_{0},\dots,x_{n}\}$, which can be the concatenation of the input and output sequence or simply a sequence of tokens, the causal-decoder computes:
\begin{equation}
    H = TRS(E(X)),
\end{equation}
where $H\in\mathbb{R}^{n\times d}$. The $t$-th vector of $H$, denoted as $H_t$, is then used to generate a distribution over the vocabulary $V$ by computing
\begin{equation}
    \hat{y}_t  = \mathrm{Softmax}(W H_t),
\end{equation}
Similar to the Transformer-decoder, we use the same cross-entropy loss function as in Equation~\ref{eq:loss_seq2seq} to train the model. This model is usually referred to the Language Model (\texttt{LM}) Transformer, since it is trained in the same way. 

\begin{figure}[t]
    \centering
    \includegraphics[width=0.9\linewidth]{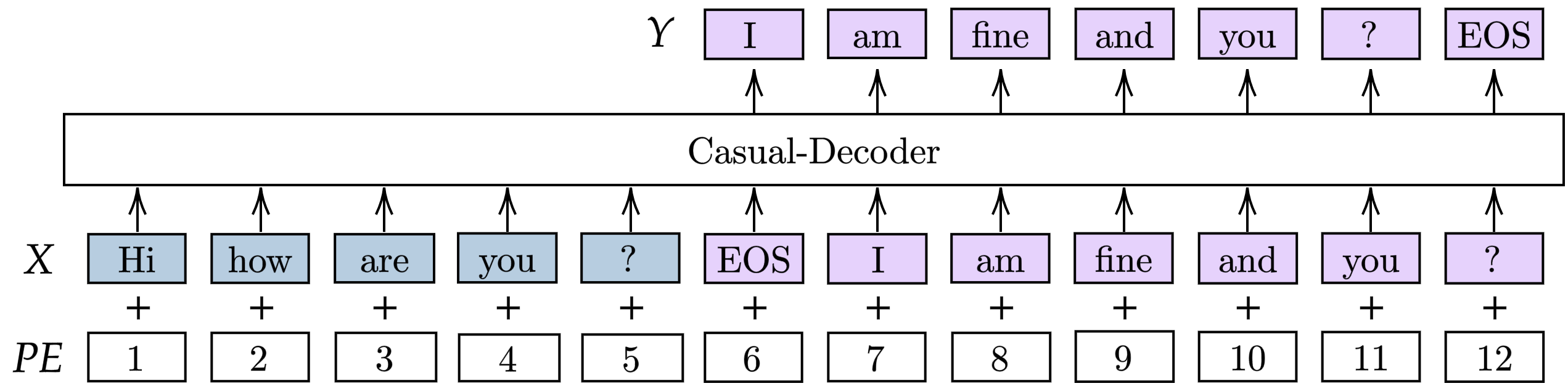}
    \caption{Example of causal-decoder Transformers. In Blue and Purple the input and output sequence respectively.}
    \label{fig:casualdec}
\end{figure}
The most effective causal-decoder, pre-trained on a large text corpus are GPT-2~\cite{radford2019language} and GPT-3~\cite{brown2020language}, while in the dialogue system scenario, DialoGPT~\cite{zhang2019dialogpt} is  pre-trained using a large number of unlabeled conversations from Reddit, (more in details in Chapter 3). Independently of the pre-training corpus, these pre-trained models are fine-tuned to specific generation tasks (e.g., dialogue response generation) by using the same Seq2Seq loss as defined in Equation~\ref{eq:loss_seq2seq}. This casual decoder model is shown in Figure~\ref{fig:casualdec}, and it is widely used in Chapter 3 and 4 of the thesis.

\subsection{Encoder-Decoder}
Pre-trained encoder-decoder models are Transformers, as described in Section~\ref{sec:transformer}, trained on a massive amount of unlabeled data (plain text or conversations). Differently from the casual-decoder, two pre-training strategies have been proposed: span-prediction~\cite{raffel2020exploring} and denoising pre-training~\cite{lewis2020bart}. In both strategies, the input sequence is corrupted and the model is taught to reconstruct the original sequence.

Several pre-trained encoder-decoder conversational models~\cite{adiwardana2020towards,roller2020recipes} have been shown to be very effective in generating human-like responses. However, these models are still very hard to control.  

\section{Related Work}
\paragraph{Task-oriented Dialogue} Task-oriented dialogue models~\citep{gao2018neural} can be categorized in two types: module-based~\citep{williams2007partially,hori2009statistical,lee2009example, levin2000stochastic,young2013pomdp,wu2019transferable} and end-to-end. In this paper, we focus on the latter which are systems that train a single model directly on text transcripts of dialogues. These tasks are tackled by selecting a set of predefined utterances~\citep{bordes2016learning,perez2016gated,williams2017hybrid,seo2016query} or by generating a sequence of tokens~\citep{wen2016network,serban2016building,zhao2017generative,serban2017hierarchical}. Especially in the latter, copy-augmented models~\citep{eric-manning:2017:EACLshort,reddy2018multi,yavuzdeepcopy} are very effective since extracting entities from a knowledge base is fundamental. 

\textbf{Chit-Chat Dialogue} \ Generating human-like responses involves overcoming a variety of challenges such as personalization~\cite{li2016persona,personachat,dinan2020second,wolf2019transfertransfo,madotto2019personalizing}, knowledge grounding~\cite{dinan2018wizard,gopalakrishnan2019topical,ghazvininejad2018knowledge,moghe2018towards,wu2020controllable,xu2021retrieval}, emotions~\cite{li2017dailydialog,rashkin2018know,zhou2018emotional,fan2020facial,li2020empathetic}, diversity~\cite{li2016diversity,li2016deep,ghandeharioun2019approximating,serban2017hierarchical,gao2018neural} and, bias~\cite{bang2021assessing,lee2019exploring,lee2021mitigating} so on. In terms of controlled dialogue generation, \citet{see2019makes} studied of conditional generative models~\cite{kikuchi2016controlling} and weighted decoding~\cite{ghazvininejad2017hafez} in controlling models trained on persona-chat. \citet{see2019makes} concluded that controlling specificity, relatedness, and repetition increase human-engagement, motivating us to extend the controllabitly to styles and topics. In this paper, we focus on these two since large pre-trained models can already achieve a high humanness score~\cite{adiwardana2020towards,roller2020recipes,zhang2019dialogpt}. 

\textbf{Controlled Text Generation} Recent methods for controlled generation include fine-tuning models using supervised learning~\cite{peng2020few,subramani2019can}, reinforcement learning~\cite{ziegler2019fine}, adversarial training~\cite{yu2017seqgan}, by pre-training models with control codes~\cite{keskar2019ctrl,ficler2017controlling,chan2020cocon}, and other various approaches~\cite{zhang2020pointer,sheng2020towards,carbone2020etc}. Alternatively, weight decoding using both bag-of-words~\cite{holtzman2018learning,ghazvininejad2017hafez,baheti2018generating,see2019makes} and discriminators~\cite{holtzman2018learning,krause2020gedi}, does not require any fine-tuning. Similarly, \citet{dathathri2019plug} propose the Plug-and-Play Language Model (PPLM) to control the generation of a pre-trained language model, e.g., GPT2~\citep{radford2019language}, both in terms of style and topic of the generated text. Finally, residual adapters~\cite{houlsby2019parameter} has been used to learn multiple language generation tasks~\cite{lin2020exploring} without fine-tuning the original models' parameters. Concurrently to our work, \citet{Smith2020ControllingSI} compare the performance and trade-offs of three existing controllable language generation methods on 200 possible styles.

\paragraph{Continual Learning in NLP} Continual learning has been explored for both classification ~\cite{d2019episodic,sprechmann2018memory,wang2020efficient,lee2021dynamically} and generation~\cite{sun2019lamol,hu2020drinking} tasks. For instance, \citet{sun2019lamol,chuang2020lifelong} proposed LAMOL, which we use as our baseline, and studied its effectiveness on a subset of DecaNLP~\cite{mccann2018natural}. On the other hand, the work of ~\citet{d2019episodic,sprechmann2018memory} is not suitable for interactive systems as dialogue systems, since their methods require local adaptation (i.e., a fine-tuning step) during inference. Finally, continual learning has been used for sentence encoding~\cite{liu2019continual}, composition language learning~\cite{li2019compositional} and relation learning ~\cite{han2020continual,lee2021towards}. However, these methods are very specific to particular NLP applications. 

\paragraph{Continual Learning in Dialogue Systems} The very early work on CL for Task-Oriented dialogue is from ~\citet{lee2017toward}, who used EWC to avoid catastrophic forgetting on three domains learned sequentially. Continual learning has also been studied in the NLG setting, where a single model was trained to learn one domain at the time in MWoZ~\cite{mi2020continual}. The authors used episodic memory to replay the example in combination with EWC. We compare similar baselines but on a larger benchmark that also includes MWoZ and the NLG setting. For the DST setting, CL was studied by ~\cite{wu2019transferable} using MWoZ, where several baselines such as L2, EWC and GEM were compared. Differently, \citet{li2019evaluate} leveraged CL for evaluating the quality of chat-bot models, and \citet{he2019mix} studied the catastrophic forgetting problem in chit-chat systems. Finally, \citet{shuster2020deploying} showed that by training models on humans-machine conversations in an open-domain fantasy world game~\cite{fan2020generating} the models progressively improved, as measured by automatic metrics and online engagement scores. 

\paragraph{Mixture of Expert \& Conditional Computation}
The idea of having specialized parameters, or so-called experts, has been widely studied topics in the last two decades~\citep{jacobs1991adaptive,jordan1994hierarchical}. For instance, different architecture and methodologies have been used such as Gaussian Processes~\citep{tresp2001mixtures}, Hierarchical Experts~\citep{yao2009hierarchical}, and sequential expert addition~\citep{aljundi2017expert}. More recently, the Mixture Of Expert~\citep{shazeer2017outrageously,kaiser2017one} model was proposed which added a large number of experts between two LSTMs. To the best of our knowledge, none of these previous works applied the results of the gating function to the parameters itself. On the other hand, there are Conditional Computational models which learn to dynamically select their computation graph~\citep{bengio2013estimating,davis2013low}. Several methods have been used such as reinforcement learning~\citep{bengio2016conditional}, a halting function~\citep{graves2016adaptive,dehghani2018universal,figurnov2017spatially}, by pruning~\citep{lin2017runtime,he2018amc} and routing/controller function~\citep{rosenbaum2018routing}. However, this line of work focuses more on optimizing the inference performance of the model more than specializing parts of it for computing a certain task. 

\chapter{Controlling Style and Topics}

In the context of conversational models, \citet{see2019makes} showed that being able to control the response generation can have a significant impact on the quality of conversations. However, controlled generation from large conversational models remains a challenge, and is particularly difficult in the absence of annotated conversational datasets.

For large language models, controlled generation has recently received increased attention. In CTRL~\cite{keskar2019ctrl}, the language model is trained to generate text conditioned to control codes presented to the model at the start of the context, while in~\citet{ziegler2019fine}, GPT-2 \citep{radford2019language} is fine-tuned using reinforcement-learning with human annotators in the loop to generate continuing text with positive sentiment. Both of these approaches require learning/fine-tuning all of the models' parameters, and new desired attributes cannot be 
easily incorporated into the generation once the models have 
been trained. 

Other approaches that do not alter the language model, but 
modify the decoding procedure for controlled generation include 1) re-weighting the output distribution using discriminators~\cite{holtzman2018learning} or bag of words~\cite{ghazvininejad2017hafez,see2019makes,baheti2018generating} and 2) perturbing the model activation with an attribute model, as in Plug-and-Play Language Model (PPLM)~\cite{dathathri2019plug}. These approaches are plug-and-play methods in that they can be used on top of any existing pre-trained language model. They do not modify or train the parameters of the original models and they can achieve comparable performance to fine-tuning methods. However, weighted decoding is generally difficult to tune because it can easily generate unrelated responses when the weight is not properly set \cite{see2019makes} and, PPLM incurs a high computational cost during the decoding stage, which is problematic for online systems such as dialogue systems. 

\begin{wrapfigure}{r}{0.5\textwidth}
    \resizebox{\linewidth}{!}{
        \begin{tabular}{rl}
        \hline
        \textit{\textbf{\small HUMAN 1}} &
          \begin{tabular}[c]{@{}l@{}}Hmmmm. I really want to try bull riding. \\ Do you have any interest in that?\end{tabular} \\
        \textit{\textbf{\small HUMAN 2}} &
          \begin{tabular}[c]{@{}l@{}}I'd love to try! Can we schedule something\\ for next week?\end{tabular} \\ \hline
         
        \textit{\textbf{\small DGPT}} &
          Yes! Let's get some stuff going. \\
         
        \textit{\textbf{\small NEGATIVE}} &
          \begin{tabular}[c]{@{}l@{}}I'm actually going to try my first ever bull \\ riding weekend in July, in the middle of \\ a mud sludge of mudslide.\end{tabular} \\
         
        \textit{\textbf{\small POSITIVE}} &
          \begin{tabular}[c]{@{}l@{}}That would be wonderful! I'm excited to\\ see you guys again in a couple of weeks :D\end{tabular} \\
         
        \textit{\textbf{\small QUESTION}} &
          \begin{tabular}[c]{@{}l@{}}What's it like? Do you ride on a regular\\ basis or just for fun?\end{tabular} \\
         
        \textit{\textbf{\small BUSINESS}} &
          \begin{tabular}[c]{@{}l@{}}Yea sure i can wait until after work. I am off \\ at the end of my shift in the middle east.\end{tabular} \\
         
        \textit{\textbf{\begin{tabular}[c]{@{}l@{}}{\small SCIENCE}\\ {\small \& TECH}\end{tabular}}} &
          \begin{tabular}[c]{@{}l@{}}Sure thing! I'll PM you my email. We can \\ set up a dropbox or something and just \\ message people together if you'd like.\end{tabular} \\
         
        \textit{\textbf{\small SPORTS}} &
          \begin{tabular}[c]{@{}l@{}}Ok how about Wednesday at 7pm central? \\ I'll be there with a buddy of mine \\ who is a bull rider.\end{tabular} \\
        \textit{\textbf{\small HUMAN 1}} &
          Sure! What does your Saturday look like? \\ \hline
        \end{tabular}
    }
    \caption{Example of controllable response generation. DGPT is the DialoGPT~\cite{zhang2019dialogpt} response, while the others are generated using different attribute models.}
\end{wrapfigure}
Open-domain conversational systems are a special case of language models, where the prefix is the dialogue history and the continuation is a human-like response~\cite{wolf2019transfertransfo}. Recently, large pre-trained language models trained on unlabeled human-to-human conversation (e.g. Reddit)~\cite{zhang2019dialogpt,adiwardana2020towards,roller2020recipes} have shown excellent performance in modelling human responses. Similarly, the output of large pre-trained conversational models cannot be directly controlled without having to re-train/fine-tune the model from scratch, which is practically inconvenient and sometimes impossible since few or no-conversational datasets exist for certain attributes or styles. 

On the other hand, plug-and-play methods are a viable solution since they do not require dialogue-specific datasets, and they can be computed online on top of existing pre-trained models. A major drawback, however, is the high computational cost~\cite{dathathri2019plug} at decoding time.
This is acceptable for language models where generating paragraphs or stories can be done offline but it is problematic for online systems, such as conversational models.

Therefore, we explore PPLM in large pre-trained dialogue models for controlling the style and topic of the responses without fine-tuning on any dialogue specific dataset. Moreover, to cope with the computational cost at the decoding time, we propose to generate style/topic-consistent responses with PPLM~\cite{dathathri2019plug} and then use it to optimize residual adapters~\cite{houlsby2019parameter} for directly learning how to steer the original distribution towards the selected attribute. 

\begin{figure}[t]
    \centering
    \includegraphics[width=0.7\linewidth]{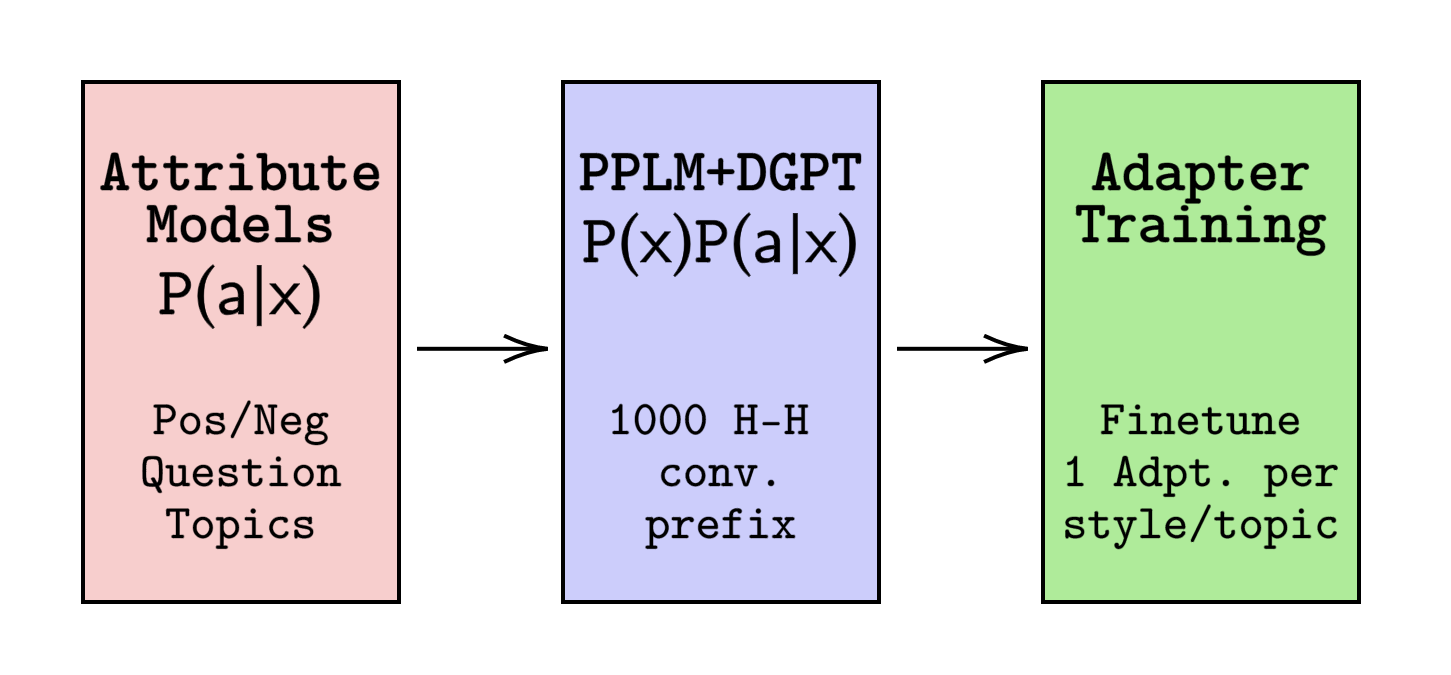}
    \caption{High-level description of the proposed method. Firstly, we train the attribute model for PPLM. Then we use PPLM to generate stylistic responses (e.g., positive or negative) from 1000 dialogue histories. Finally, we distill the generated response into residual adapters, one per attribute.}
    \label{fig:PPCM_GENERAL}
\end{figure}
With our extensive automatic and human evaluation, we empirically demonstrate that plug-and-play methods are effective in controlling the response while being computationally efficient. To summarize, our key contributions are:

\begin{itemize}[leftmargin=*]
    \item We show the effectiveness of plug-and-play methods in large pre-trained conversational models using a variety of styles and topics such as Positive, Negative, Question, Sport, Business/Finance, without using a dialogue-specific dataset.
    \item We propose to use residual adapters~\cite{houlsby2019parameter}, which adds less than 1.5\% task-specific parameters per style/topic, to make the controllable response generation viable for online systems. 
    \item We run a comprehensive automatic and human evaluation to show that plug-and-play methods can control the generated responses in terms of style and topics, without losing fluency.
    \item We carry out a thorough qualitative analysis on the difficulty of steering conversational models, highlighting current limitations and possible solutions.
\end{itemize}
The high-level description of the proposed methodology is shown in Figure~\ref{fig:PPCM_GENERAL}.

\section{Methodology}
\label{sec:method} 
A dialogue consists of one or more alternating turns between two speakers. We denote the dialogue history $\hat{X}$ as a single sequence of tokens from the concatenation of the alternating utterances from the user and the system turns respectively. Without loss of generality, we assume that $\hat{X}$ has all the dialogue history without the last system
utterance, denoted as $Y$. We model the dialogue responses using a Transformer~\cite{vaswani2017attention}-based casual-decoder (\texttt{LM}) by using the dialogue history $\hat{X}$ as a prefix and then generating the continuation $Y$ in an auto-regressive manner~\cite{DBLP:journals/corr/abs-1901-08149}. 
\paragraph{Causal-Decoder LM} Let us re-define the concatenation of $\hat{X}= \{x_{0},\dots,x_{n}\}$ and output $Y= \{x_{n+1},\dots,x_{m}\}$ as the sequence of tokens $X=\{x_{0},\dots,x_{n+m}\}$. Then we can compute the language model distribution using the chain rule of probability~\cite{bengio2003neural} as
\begin{equation}
    p_{\theta}([\hat{X};Y]) = p_{\theta}(X)=  \prod_{i=0}^{n+m}p_{\theta}(x_i|x_{0}, \cdots, x_{i-1}),
\end{equation}
where $\theta$ is the model's parameters and $[;]$ denotes the concatenation of $X$ and $Y$. We define the Transformer decoding process in a recursive manner. Let us define the matrix $F_t$ as the key-value pairs from the past dialogue history, i.e., $F_t = [(K_{t}^{(1)}, V_{t}^{(1)}),\cdots, (K_{t}^{(l)}, V_{t}^{(l)})]$, where $(K_{t}^{(i)}, V_{t}^{(i)})$ corresponds to the key-value pairs from the $i$-th layer generated at all time-steps from 0 to $t$.

Thus, we define the recurrent decoding process as:
\begin{equation}
    o_{t+1}, F_{t+1} = \texttt{LM} (x_{t}, F_t),
\end{equation}
and then $x_{t+1}$ is sampled from the distribution $p_{t+1}= \textrm{Softmax}(W o_{t+1})$, where $W$ is a linear transformation that maps the hidden state of the last layer $o_{t+1}$ to a vector of the vocabulary size. This efficient Transformer implementation~\citep{huggingface} leverages the cached memories to generate $x_{t+1}$ without recomputing $F_t$.

\begin{figure}[t]
    \centering
    \includegraphics[width=\linewidth]{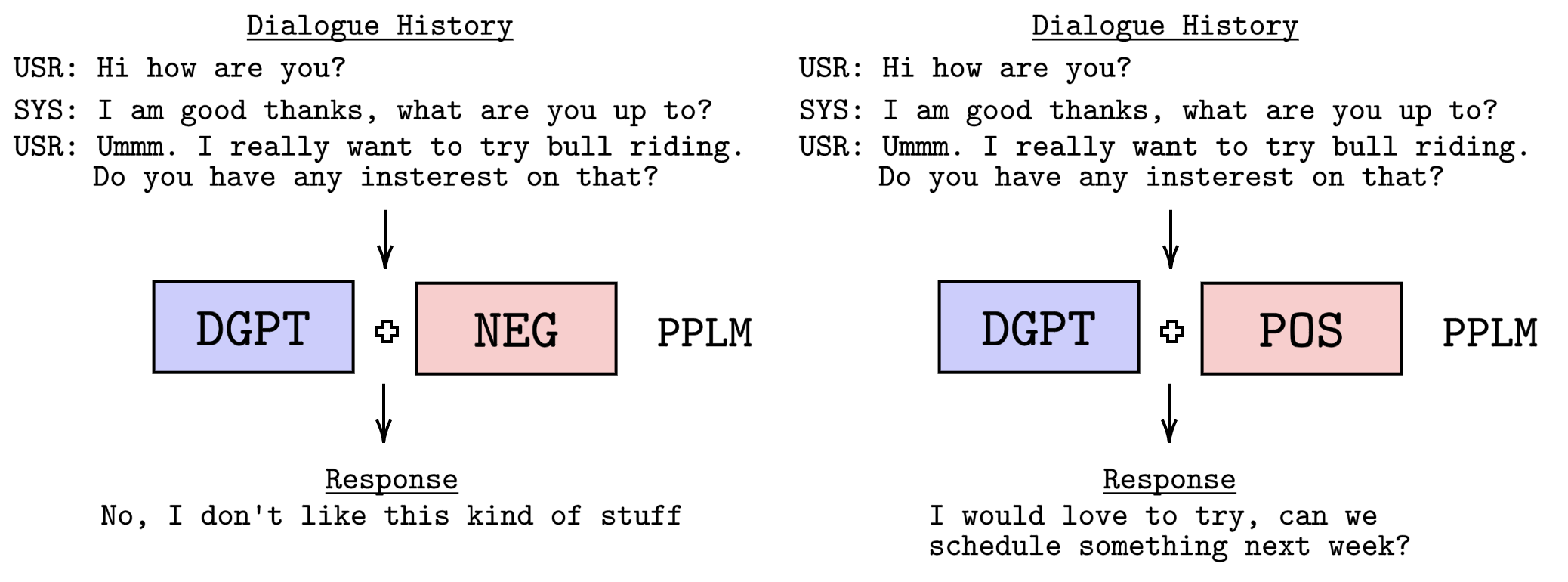}
    \caption{Example of PPLM applied to a large pre-trained dialogue model (DGPT~\cite{zhang2019dialogpt}).}
    \label{fig:PPLM}
\end{figure}
\subsection{Plug-and-Play Language Model}
\label{subsec:PPLM}
PPLM~\cite{dathathri2019plug} uses an attribute model (i.e., a classifier) for controlling the generated text. We denote the attribute model as $p(a|X)$, where $a$ is the specific desired attribute to optimize for (e.g., positivity), and $X$ is the generated response so far. At every generation step $t$, PPLM perturbs the history matrix $F_t$ in the direction of the sum of two gradients: i) to maximize the log-likelihood of the attribute $a$ under the conditional attribute model $p(a|X)$ and ii) to ensure the high log-likelihood of the generated text under the unmodified conversational language model $p(X)$. 
The gradient updates are restricted to $F_t$ so as to preserve the original model parameters.

Let $\Delta{F}_t$ be the update to $F_t$ to shift the generated text towards possessing the desired attribute $a$ i.e., $o_{t+1}, F_{t+1} = \texttt{LM} (x_{t}, F_t + \Delta{F}_t)$. At the beginning of the generation, $\Delta{F}_t$ is initialized to zero and it is updated using the gradients from the attribute model. We rewrite the attribute model $p(a|X)$ as $p(a|F_t + \Delta{F}_t)$, and we define the gradient update for $\Delta{F}_t$ as
\begin{equation}
      \Delta{F}_{t} \leftarrow \Delta{F}_{t} + \alpha \frac{\nabla_{\Delta{F}_{t}} \log p(a|F_t + \Delta{F}_t)}
        {\| \nabla_{\Delta{F}_{t}} \log p(a|F_t + \Delta{F}_t) \|^{\gamma} },
        \label{pplm}
\end{equation}
where $\alpha$ is the step size and $\gamma$ is the scaling coefficient for the normalization term. Equation~\ref{pplm} is repeated $p$ times depending on how strongly we want the response to be conditioned to the attribute. We study the effect of the step size $\alpha$ and the number of iterations $p$ on the generated text in detail in Section~\ref{sec:analysis}. 
Subsequently, the new $\widetilde{F}_t = F_t + \Delta{F}_t$ is computed and a new token is generated using $\widetilde{o}_{t+1}, F_{t+1} = \texttt{LM} (s_{t}, \widetilde{F}_t)$. The described optimization process is repeated for every token in the generated sequence. As previously mentioned, to ensure fluency we also take a step towards minimizing the Kullback–Leibler (KL) regularization between the perturbed and the original distribution. In addition, we also use Post-norm Geometric Fusion~\cite{stahlberg2018simple,dathathri2019plug} to avoid adversarial generation~\cite{szegedy2013intriguing}.

\paragraph{Attribute Models}
The discriminator is a linear classifier $g$ trained on an annotated dataset with sentence and label pairs as $(X,y)$ --- note that these sentences do not necessarily need to be conversational responses, as in our case. 
For each sentence $x$ of length $t$, we compute the set of hidden states $o^x_{:t}$ from the \texttt{LM}, then we compute the mean ($\bar{o}^{t}$) across time, and finally we train $g$ using the cross-entropy between the label distribution $y$ and $g(\bar{o}^{t})$. Figure~\ref{fig:PPLM} shows an example of how PPLM uses different discriminators to control the output generation of DialoGPT (DGPT).

\subsection{Residual Adapters}
\label{subsec:Adapters}
Residual adapters~\cite{houlsby2019parameter,bapna2019simple} are trainable modules added on top of each Transformer layer, which steer the output distribution of a pre-trained model without modifying the original weights. 
An adapter block consists of layer normalization~\cite{ba2016layer}, followed by two linear layers~\cite{hinton1994autoencoders} with a residual connection. Given the hidden representation at layer $l$, denoted as $H \in \mathbb{R}^{t\times d}$, of a Transformer~\cite{vaswani2017attention}, where $d$ is the hidden size and $t$ is the sequence length, the residual adapter computes
\begin{equation}
\vcenter{\hbox{\begin{minipage}{5cm}
\centering
\includegraphics[width=\linewidth]{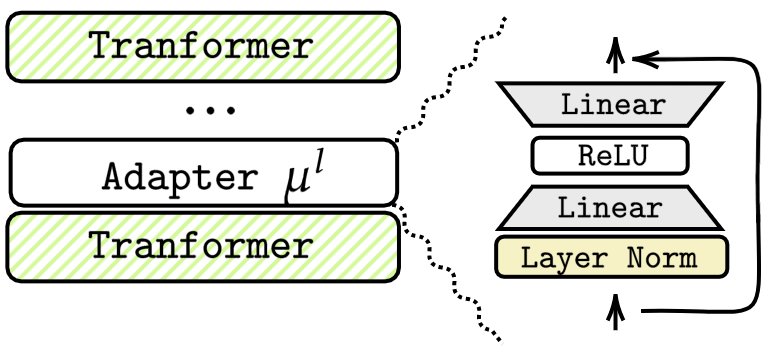}
\end{minipage}}}
\qquad\qquad
\begin{aligned}
\mathrm{Adapter}_{\mu^l}(H) =  \mathrm{ReLU}(\mathrm{LN}(H) W_l^{E})W_l^{D} + H,
\end{aligned}
\end{equation}
where $W_l^{E}$ and $W_l^{D}$ are trainable parameters of dimensions $d\times b$ and $b\times d$ respectively, and LN$(\cdot)$ denotes the layer normalization. The bottleneck dimension $b$ is a tunable hyper-parameter that allows adjustment of the capacity of the adapter according to the complexity of the target task. We define the set of $\mu = \{ W_0^{E}$, $W_0^{D}, \cdots,  W_L^{E}$, $W_L^{D} \}$ as the set of parameters for the adapter for a model with $L$ layers.



\begin{figure}[t]
    \centering
    \includegraphics[width=0.7\linewidth]{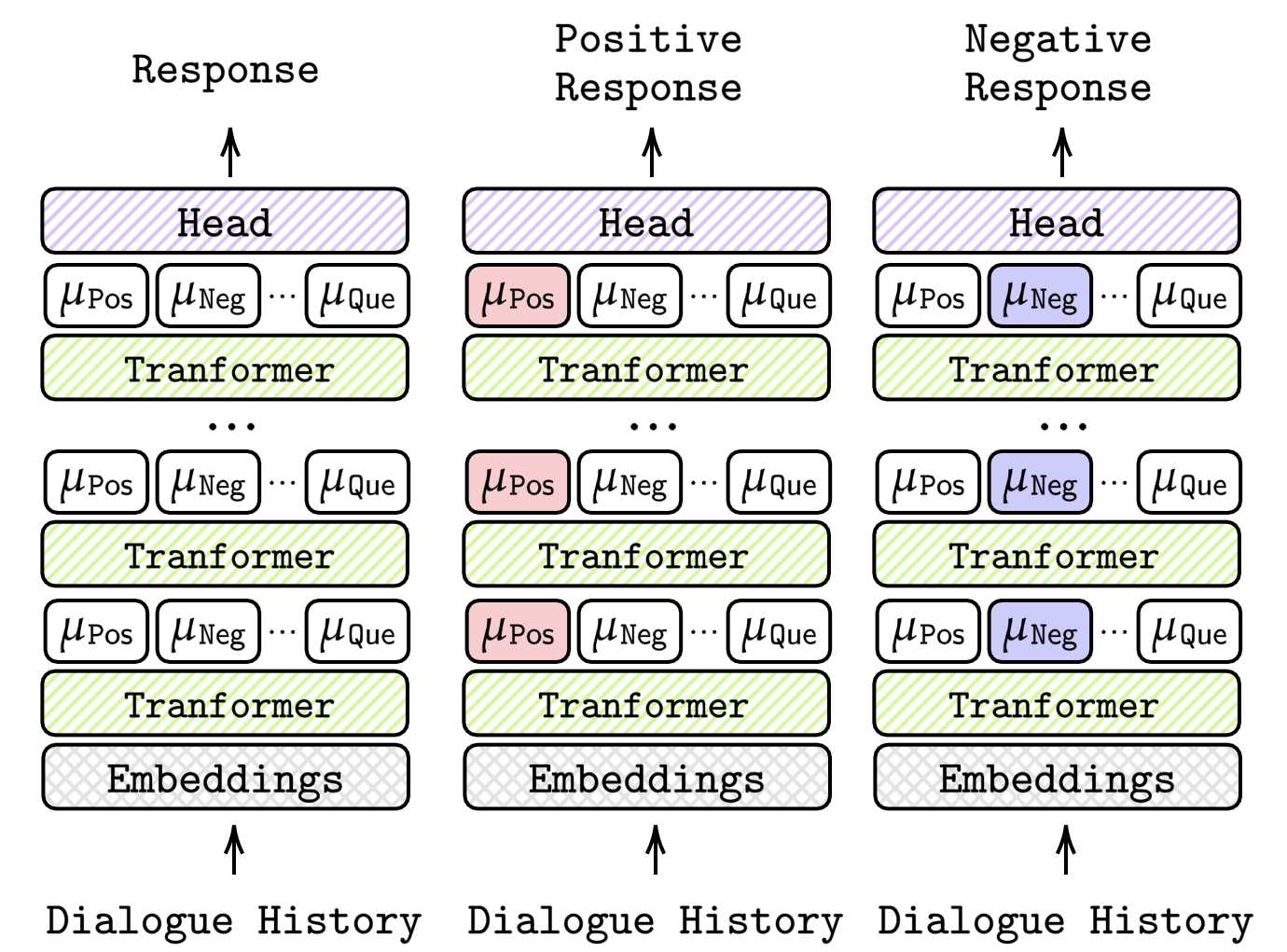}
    \caption{Plug-and-play adapter architecture. The Transformer layers, head, and embedding layer are frozen during training, while the adapter layers are trained independently per each attribute.}
    \label{fig:PPADAPTER}
\end{figure}
\paragraph{Plug-and-Play Adapters}
At decoding time, PPLM uses a fixed number of iterations $p$ to generate a single token. This makes the model impracticable for interactive tasks such as conversational models. To cope with this issue, we propose to first use PPLM to generate datasets of dialogues with certain attributes $a$, denoted as ${D}_{a} = \{(X_i,Y_i)\}_{i}^{r}$, where $X_i$ is the dialogue history and $Y_i$ the corresponding PPLM, generated response. Then we optimize the residual adapter parameters to steer the output of the original \texttt{LM} distribution. Hence, for each attribute $a$, we optimize the parameters in $\mu_a$ to minimize the negative log-likelihood over the dataset of dialogues $\mathscr{D}_{a}$. Formally,
\begin{equation}
  L_{\mu_a}(D_a) =  - \sum_j^{|D_a|} \sum_{i=0}^{n+m} \log p_{\mu_a}(x_i^j|x_{0}^j, \cdots x_{i-1}^j), \label{eq:lossadpt}
\end{equation}
where each response $n+m$ is of maximum length in $D_a$. Figure~\ref{fig:PPADAPTER} shows the final Transformer with one adapter per attribute. 

\section{Experimental Setup}
In this section, we conduct extensive experiments on the proposed methodology using both automatic and human evaluation. Differently from PPLM~\cite{dathathri2019plug}, where a set of pre-defined prefixes are used to trigger the generation, in our experiments, we use 100 conversations~\cite{adiwardana2020towards} for generating 1100 possible prefixes (i.e., moving window of size two). These open-domain generic dialogues serve as a prefix to trigger the responses rather than fine-tuning. 
In all our experiments, we use DialoGPT-medium~\cite{zhang2019dialogpt}, a large pre-trained model trained on 147 Million multi-turn dialogues from Reddit, spanning from 2005 to 2017. Importantly, the proposed methodology is model agnostic, and thus it can be applied to any other large pre-trained model such as Meena~\cite{adiwardana2020towards} and Blender-Bot~\cite{roller2020recipes}. 

Since the plug-and-play adapters use the generated responses from PPLM, we randomly split the prefixes, with 80\% for learning the adapter perturbation and the remaining 20\% for the final automatic and human evaluation. 
This is done to have a fair comparison between other baselines and adapters (See Appedix A for more details). 

\subsection{Attribute Models}
We train three discriminators covering six attribute models: Positive, Negative, Question, Sci/Tech, Business and Sport. To control the Positive and Negative responses, we use SST-5~\cite{socher2013recursive} with the classes Very-Positive and Very-Negative as the attribute, and to control for Question, we use the speech-act annotation from Daily Dialogue~\cite{li2017dailydialog} with the Question class as the attribute. To avoid any dialogue-related data, we only use sentences without their corresponding context. Finally, to generate responses about Sci/Tech, Business and Sport, we use the AG-NEWS~\cite{zhang2015character} topic-classification dataset, using the respective classes as attributes. As mentioned in Section~\ref{subsec:PPLM}, we freeze the DialoGPT parameters and train a linear classifier on top of the representations from the final layer of its Transformer blocks. Table~\ref{tab:discriminator}, shows the sample size statistics and the performance in terms of F1-score for all the aforementioned datasets. We also report the current state-of-the-art to show that a linear classifier trained on top of the DialoGPT activation can reach competitive performance. 

\begin{table}[t]
\centering
\begin{tabular}{r|c|c|cc|ccc}
\hline
\multicolumn{1}{c|}{\multirow{2}{*}{\textbf{Dataset}}} & \multirow{2}{*}{\textbf{Task}} & \multirow{2}{*}{\textbf{\#C}} & \multicolumn{2}{c|}{\textbf{Samples}} & \multicolumn{3}{c}{\textbf{F1-Score}} \\ \cline{4-8} 
\multicolumn{1}{c|}{} &  &  & \textit{Train} & \textit{Test} & \textit{Train} & \textit{Test} & \multicolumn{1}{l}{\textit{SOTA}} \\ \hline
\textit{SST-5}~\cite{socher2013recursive} & Sentiment & 5 & 318,582 & 2210 & 77.68 & 47.01 & 55.50$\dagger$ \\ \hline
\textit{Daily Dialogue}~\cite{li2017dailydialog} & Act & 4 & 92,650 & 10,295 & 80.58 & 80.00 & 86.10$\ddag$ \\ \hline
\textit{AG NEWS}~\cite{zhang2015character} & Topic & 4 & 120,000 & 7,600 & 90.68 & 90.65 & 95.44$\mathsection$ \\ \hline
\end{tabular}
\caption{Attribute dataset statistics and performance. State-of-the-art (\textit{SOTA}) results are taken from $\dagger$~\cite{munikar2019fine}, $\ddag$~\cite{kumar2019practical}, and $\mathsection$~\cite{yang2019xlnet}.}
\label{tab:discriminator}
\end{table}

\subsection{Baselines}
We compare multiple plug-and-play settings: \textbf{DG}: DialoGPT, proposed by~\citet{zhang2019dialogpt}; \textbf{WD}: DialoGPT plus a word-level weight-decoding schema, as in~\cite{ghazvininejad2017hafez} and \cite{see2019makes}; \textbf{PP}: DialoGPT plus PPLM~\cite{dathathri2019plug}, as explained in Section~\ref{subsec:PPLM}; and \textbf{AD}: DialoGPT with one adapter per style, as explained in Section~\ref{subsec:Adapters}. In all the baselines, we sample 10 hypotheses using multinomial-sampling after a top-k filtering (with $k=10$), to ensure response diversity~\cite{zhang2020trading}, and we select the hypotheses with the lowest attribute model loss as the response. This re-ranking technique has shown itself to be very effective for generating good responses~\cite{adiwardana2020towards,dathathri2019plug}.


\subsection{Evaluation Metrics}
We evaluate the generated responses using both automatic and human evaluations. 

\textbf{Automatic Eval.} in open-domain chat is challenging~\cite{liu2016not}, especially when using n-gram methods over a single reference (e.g., BLEU~\cite{papineni-etal-2002-bleu}). In this paper, no gold-reference response is provided (e.g., stylistic human-generated response). Thus we rely on unsupervised measures for fluency, diversity and style/topic. For fluency, we compute the perplexity score of the dialogue prefix plus the generated response using GPT2~\cite{radford2019language}. While for diversity, we use the distinct n-grams~\cite{li2016diversity} (normalized by the length of the text) across all the responses generated by a given method. To evaluate the attribute consistency, we train external classifiers using non-overlapping data with the attribute model. For sentiments, we use AMAZON-5~\cite{mcauley2013hidden} product reviews, and for topics, we use the test-set data of AG-NEWS~\cite{zhang2015character} because we could not find another topic classification dataset with the same classes. For each dataset, we train a separate BERT~\cite{devlin2019bert} (base) classifier with a simple classification head. Table 2 in Appendix B, summarizes the dataset statistics and the performance of the trained scorer.

\textbf{Human Eval.} is the most effective way to evaluate open-domain chat-bots. In this chapter, we evaluate two aspects of the generated response: humanness and attribute consistency. The first is used to evaluate the fluency and coherence of the generated responses. The second is used, to evaluate whether the generated responses respect the style or the topic enforced by the attribute model. We use Acute-Eval~\cite{li2019acute}-style A/B testing, in which we compare all possible model pairings (e.g., PP vs. DG etc.). For each comparison, we show the same dialogue context and two possible options, one generated from model A and one from model B. Then we ask the annotators to select among four options: model A, model B, both or neither. We collect annotations for both humanness and attribute consistency on 30 dialogues per model comparison and attribute, which amounts to a total of 4200 human annotations. Further details are provided in Appendix C.

\begin{table}[t]
\centering
\resizebox{\textwidth}{!}{
\begin{tabular}{rcccccccccc}
\multicolumn{1}{l}{} & \multicolumn{1}{l}{} & \multicolumn{1}{l}{} & \multicolumn{1}{l}{} & \multicolumn{1}{l}{} & \multicolumn{5}{c}{\textbf{Score by Attribute}} \\ \hline
\multicolumn{1}{c|}{\textbf{}} & $\downarrow$  \textbf{Ppl.}& $\uparrow$ \textbf{Dist 1/2/3} & \textbf{Discrim.} & \multicolumn{1}{c|}{\textbf{Score}} & \textbf{Posi.} & \textbf{Nega.} & \multicolumn{1}{l}{\textbf{Busin.}} & \multicolumn{1}{l}{\textbf{Sci/Tech}} & \multicolumn{1}{l}{\textbf{Sport}} \\ \hline
\multicolumn{1}{r|}{\textit{DG}} & \textbf{39.60} & 0.22/0.64/0.77 & 46.48 & \multicolumn{1}{c|}{32.91} & 65.67 & 19.40 & 17.41 & 91.04 & 27.86 \\
\multicolumn{1}{r|}{\textit{WD}} & 53.03 & 0.25/0.74/\textbf{0.84} & 50.18 & \multicolumn{1}{c|}{34.54} & 58.21 & 28.86  & 19.40 & 91.04 & 36.82 \\
\multicolumn{1}{r|}{\textit{PP}} & 45.86 & 0.24/0.67/0.79 & 73.28 & \multicolumn{1}{c|}{49.54} & 75.12 & 51.74  & 47.26 & 93.03 & 59.20 \\
\multicolumn{1}{r|}{\textit{AD}} & 41.57 & 0.17/0.58/0.77 & \textbf{96.52} & \multicolumn{1}{c|}{\textbf{70.01}} & \textbf{93.03} & \textbf{73.13} & \textbf{68.66} & \textbf{99.00} & \textbf{83.08} \\ \hline
\end{tabular}
}
\caption{Automatic evaluation results. In all the metrics, higher is better, except for Perplexity (Ppl.). \textit{Discrim.} is the accuracy of the internal attribute model, while \textit{Score} is the accuracy of the external classifier. All the results, are averaged among the six attribute models.} \label{Tab:auto}
\end{table}
\begin{figure}[t]
    \centering
    \begin{subfigure}[b]{0.45\textwidth}
         \centering
         \includegraphics[width=\textwidth]{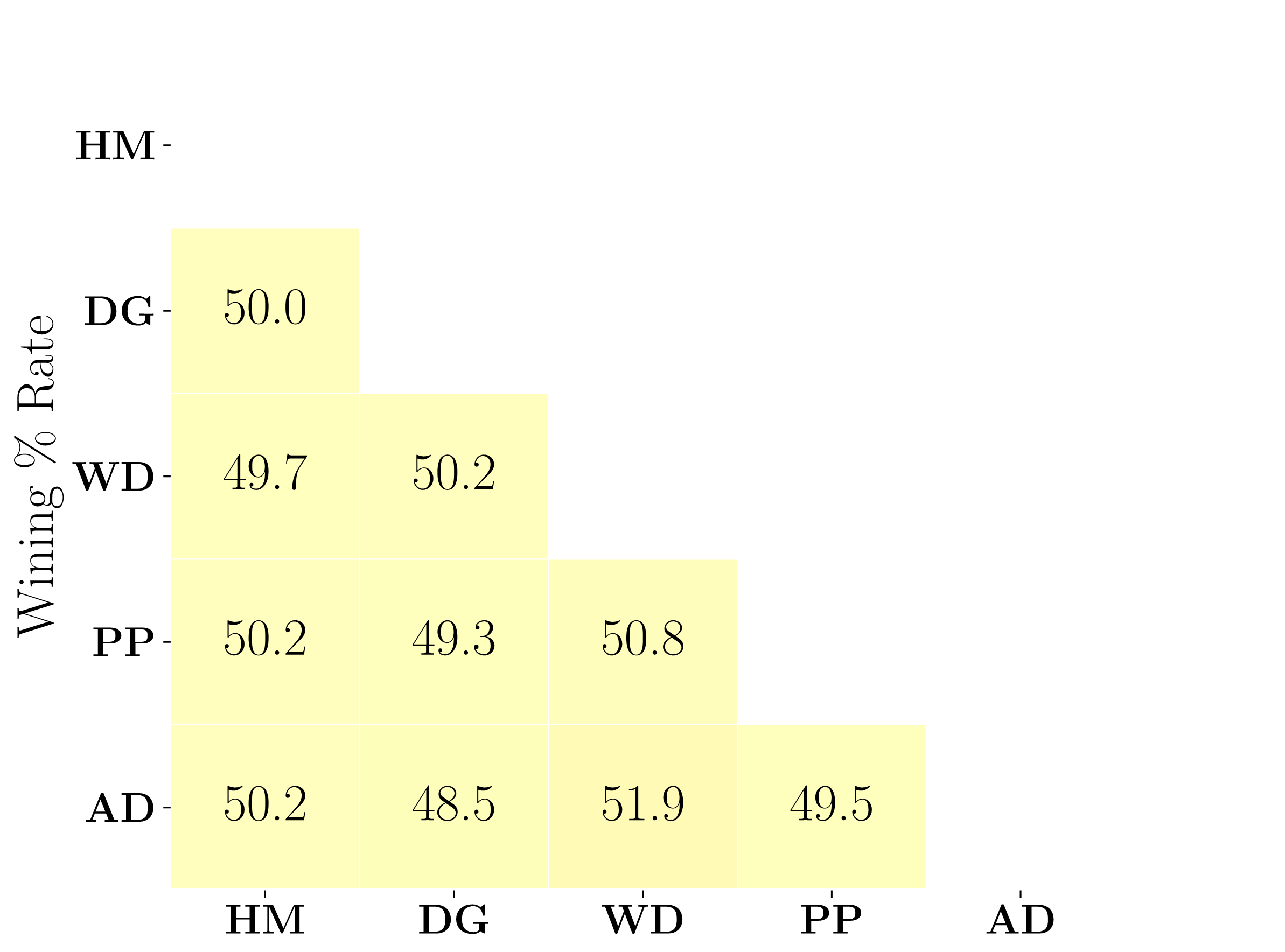}
         \caption{Humanness}
         \label{fig:very_negative}
     \end{subfigure}
     \begin{subfigure}[b]{0.45\textwidth}
         \centering
         \includegraphics[width=\textwidth]{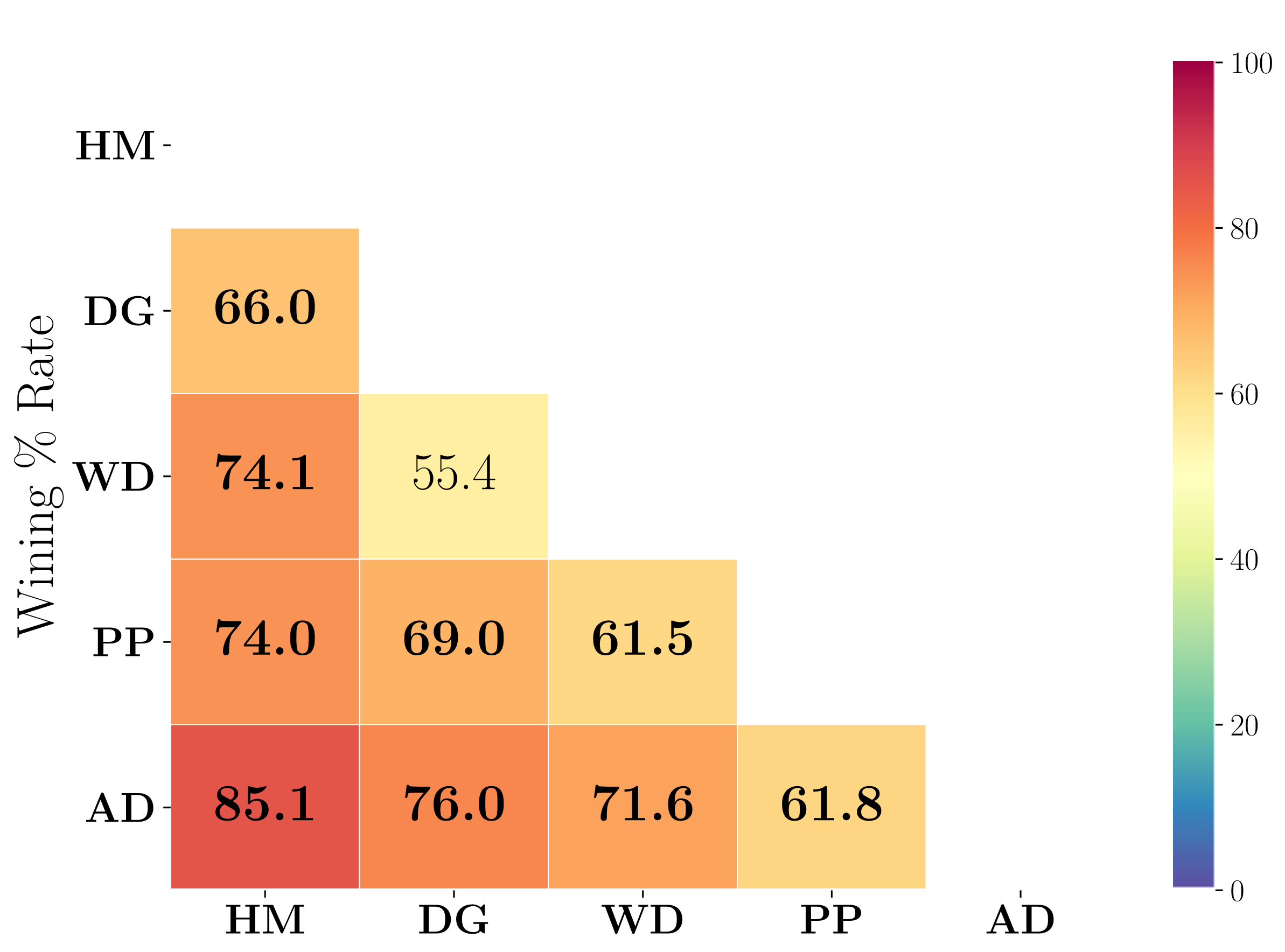}
         \caption{Attribute consistency}
         \label{fig:very_positive}
     \end{subfigure}
    \caption{Human evaluation results in terms of winning rate for both humanness and attribute consistency. For example, in the Attribute Consistency table, \textbf{DG} wins 66\% of the time versus \textbf{HM}. Bold results are statistically significant ($p<0.05$).}
    \label{fig:human}
\end{figure}

\section{Results}
In this section, we evaluate the proposed methodology to answer three research questions:  \textbf{1)} is it possible to use plug-and-play methods to control the output of a large pre-trained conversational model? and if so, \textbf{2)} what are the most effective plug-and-play methods, and \textbf{3)} how difficult is to control the response generation given various attributes? To answer these questions, we rely on both automatic and human evaluation. Table~\ref{Tab:auto} and Figure~\ref{fig:human} reports the aggregated result for all the styles and topics in both evaluations. The breakdown per attribute is reported in Appendix D. 

\begin{figure}[t]
    \centering
    \begin{subfigure}[b]{0.45\textwidth}
         \centering
         \includegraphics[width=\textwidth]{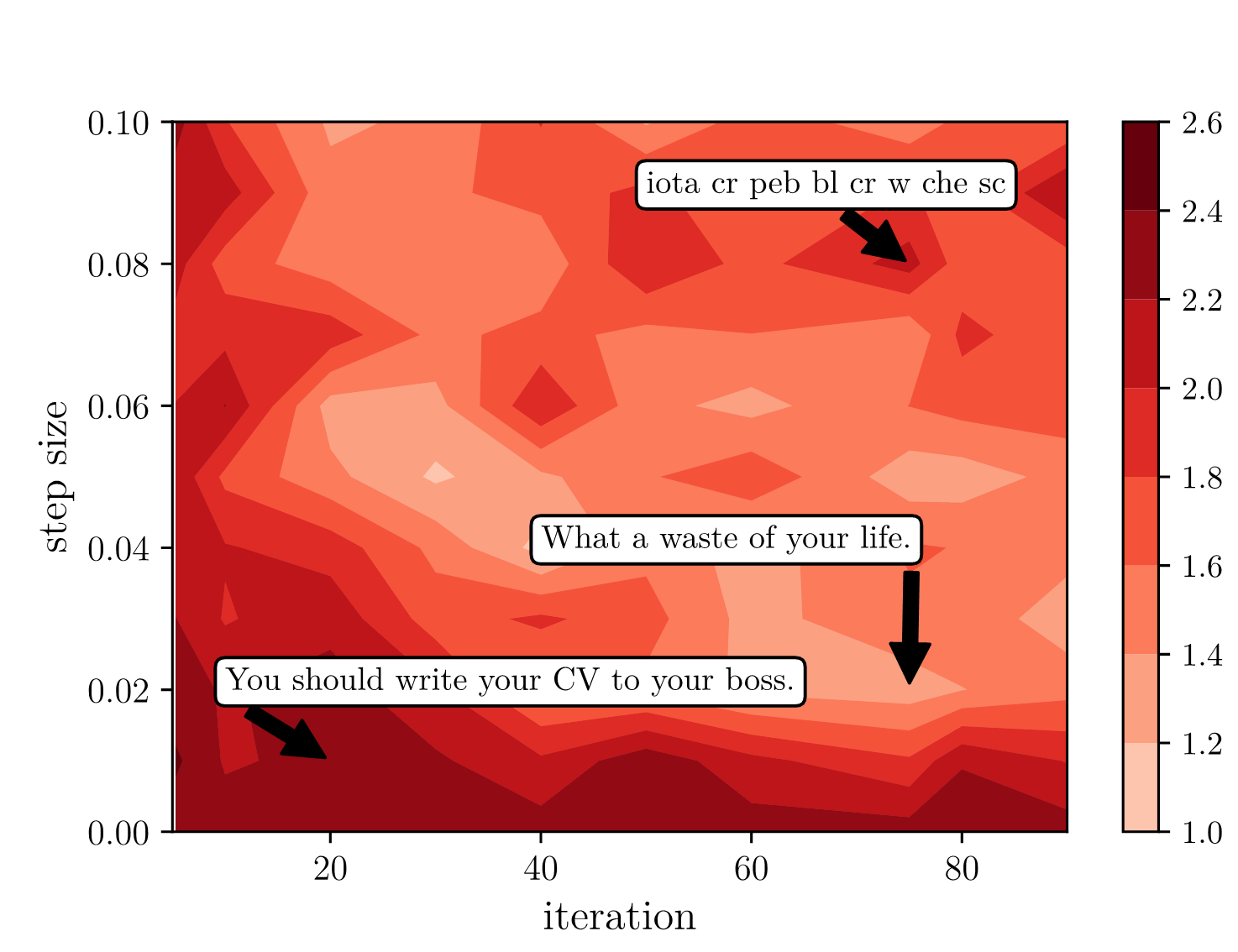}
         \caption{Negative}
         \label{fig:very_negative_iter}
     \end{subfigure}
     \hspace{0.03\textwidth}
     \begin{subfigure}[b]{0.45\textwidth}
         \centering
         \includegraphics[width=\textwidth]{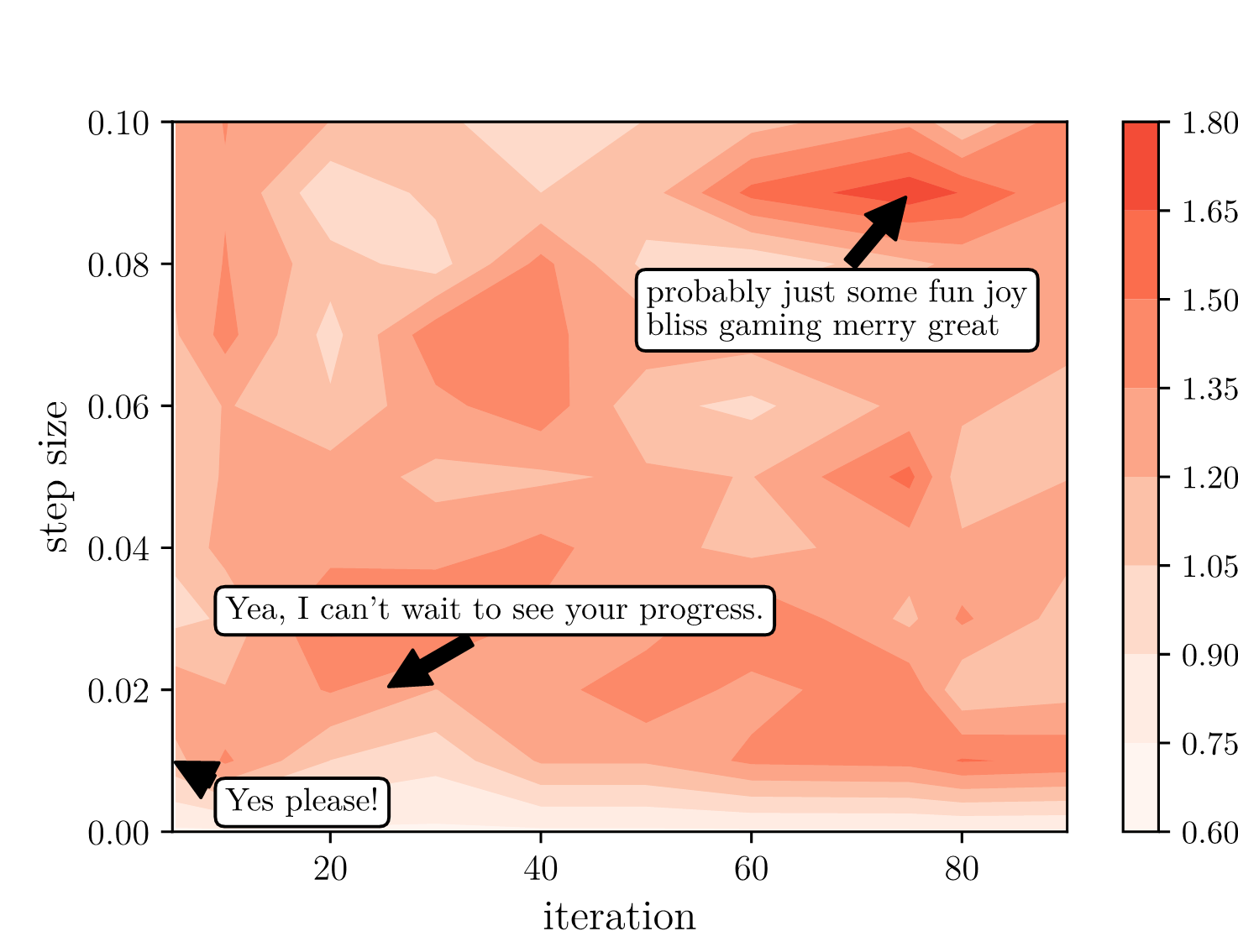}
         \caption{Positive}
         \label{fig:very_positive_iter}
     \end{subfigure}
    \caption{Contour plot of the normalized sum of the log perplexity score, computed by GPT2~\cite{radford2019language} and the external classifier loss on the response generated by PPLM for the Negative and Positive style. The $x$-axis is the number of iteration, $p$, and the $y$-axis is the step size $\alpha$. Darker areas correspond to a higher loss sum, meaning a higher perplexity and higher classification loss. The labels represent a sample response from a given iteration and step size. }
    \label{fig:contourplot}
\end{figure}
\subsection{Quantitative Evaluation}
\label{sec:quant_eval}
\textbf{Automatic Eval.} The major evaluation criterion is to have responses that are as fluent as the original DialoGPT or humans while following the style or topic enforced by the attribute model. In Table~\ref{Tab:auto}, we can see that DialoGPT (DG) achieves the lowest perplexity, but it also has the lowest aggregate attribute score ("Score" in Table~\ref{Tab:auto}). By analysing the breakdown by style, we can see that, by default, the original model has a higher score in both Positive style and Sci/Tech topic. 
We hypothesize that this is due to two factors: 1) The discussions in Reddit are more often related to Sci/Tech topics. By being provided general questions as input, e.g., ``What do you do for living?", the model often generates tech-related responses, e.g., ``I am a computer science student". 2) \citet{zhang2019dialogpt} filtered undesired and toxic responses from the Reddit conversations used in training DialoGPT, which explains the positivity of its responses.

Using weight decoding (WD) on top of DialoGPT leads to an improvement in both the diversity score and the external classifier score. However, WD tends to increase the perplexity score, showing that the generation fluency with respect to the context is lost. In the preliminary experiments, we notice that WD generates responses that are not related to the dialogue context but are highly similar to the distribution of the discriminator datasets. 
This is consistent with the observation in \cite{see2019makes} that 
WD is difficult to tune and often 
provides control at the cost of fluency, 
leading to non-sensical generation.
On the other hand, PPLM (PP) is able to achieve a lower perplexity compared to WD while attaining both a higher attribute consistency score and a high response diversity (dist). 
We hypothesize that better performance is due the ability of PPLM to dynamically perturb the latent activation of the model without breaking the original distribution thanks to the KL regularization and the Post-norm Geometric Fusion~\cite{stahlberg2018simple}. 

The adapter plug-and-play setting has the highest overall attribute score and the lowest perplexity between PP and WD. However, the response diversity, especially dist-1, is lower than for other baselines, meaning that the response may contain repetitive tokens (e.g., ``so so bad"). In general, adapters optimized with PPLM-generated responses, which usually are not perfect, can properly learn to steer the output distribution without breaking the original DialoGPT output. As previously mentioned, this also comes with the advantage of not computing the PPLM perturbation at decoding time.   

\begin{table*}[t]
    \centering
    \resizebox{\textwidth}{!}{
        \begin{tabular}{llll}
    \hline
     &
      HUMAN 1 &
      \multicolumn{2}{l}{Are you doing any home decorating then?} \\
     &
      HUMAN 2 &
      \multicolumn{2}{l}{Yes! We set up an eco-friendly (i.e. fake) Christmas tree and put up some colorful LED lights which is very festive.} \\ \hline
     &
      \multicolumn{2}{l|}{\textbf{NEGATIVE}} &
      \textbf{POSITIVE} \\ \hline
    \multicolumn{1}{l|}{\textit{DG}} &
      \multicolumn{2}{l|}{\begin{tabular}[c]{@{}l@{}}That sounds awesome. I'm going to get a couple of these \\ and start decorating.\end{tabular}} &
      Oh that's so cool! I love your stuff!! \\ \hline
    \multicolumn{1}{l|}{\textit{WD}} &
      \multicolumn{2}{l|}{\begin{tabular}[c]{@{}l@{}}I'm not sure if you're aware of this, but the word \\ ecoreneive actually refers to being ecoreneous...\end{tabular}} &
      Very nice, good sir \\ \hline
    \multicolumn{1}{l|}{\textit{PP}} &
      \multicolumn{2}{l|}{I'm not a fan of LEDs in general. They always seem to fail.} &
      Oh wow awesome! Thank you so much for your time! \\ \hline
    \multicolumn{1}{l|}{\textit{AD}} &
      \multicolumn{2}{l|}{That sounds like the absolute most boring thing. EVER.} &
      That is amazing! I am so excited!! :D So creative and creative!! :D \\ \hline
    \end{tabular}
    }
    \caption{Examples of generated responses for Negative and Positive with the same starter. }
    \label{tab:responses_neg_pos}
\end{table*}
\textbf{Human Eval.} In Figure~\ref{fig:human}, we report the winning rate of the A/B testing for both humanness and attribute consistency. From the figure, we highlight the following: 1) There is no statistically significant difference in the humanness score among the multiple methods, even with 210 annotations per cell. In general, all the methods lose with the human response (HM), but not by a large margin. This is due to the fact that the annotators choose the ``both" option more often. 2) In terms of attribute consistency, we observe that the methods form a clean, well-ordered rank,  \textbf{AD}$>$\textbf{PP}$>$\textbf{WD}$>$\textbf{DG}$>$\textbf{HM}, which confirms the automatic evaluation results. Different from humanness, all the results except those for WD vs. DG are statistically significant ($p<0.05$), showing the adapter clearly defeats other methods. 

To answer the first two research questions, we observe that both automatic and human evaluation show that plug-and-play methods are suitable for controling response generation. Moreover, the most effective method is the adapter plug-and-play, which produces fluent, and attribute consistent responses, while being three orders of magnitude faster than PPLM at inference time (148.5s/token vs. 0.123s/token) using a single Nvidia 1080Ti.

\section{Analysis}
\label{sec:analysis}
In this section, we evaluate the difficulty of controlling the response generation for a given attribute.
To do so, we analyse the behaviour of PPLM over two opposite styles (positive and negative) and then we conduct a qualitative evaluation over the generated responses.

\paragraph{Iteration \& Step Size}
We analyse the loss of the automatic scorer for fluency and attribute consistency to understand the effects of the number of iterations $p$ and the step size $\alpha$ in Equation~\ref{pplm}. Figure~\ref{fig:contourplot} depicts the normalized sum of the log perplexity score, computed by GPT2~\cite{radford2019language} and the external classifier loss on the response generated for the Negative and Positive style. In general, the aggregate loss for the negative attribute (Figure~\ref{fig:very_negative_iter}) is higher for the Positive attribute (Figure~\ref{fig:very_positive_iter}), as also shown in the sampled responses, where a small step size and few iterations leads to positive responses. However, when both the step size and the iteration surpass a certain threshold, the conditioning becomes very strong and the text generated by PPLM loses its fluency. Overall, this visualization suggests that it is more laborious to control for the negative sentiment with PPLM, and there is a smaller region for the hyper-parameters space where the responses are both fluent and attribute-consistent.


\paragraph{Qualitative Analysis}
We sample and read 200 dialogue responses from the adapter plug-and-play model (AD), and we study the overall quality of the responses especially to understand when and why DialoGPT is hard to steer. We discover three possible factors:
\textbf{1)} the hardness of the response steering is influenced by the context, 
\textbf{2)} available vocabulary for attributed style/topic, and \textbf{3)} mutual exclusivity of the attribute-specific vocabulary.

\textbf{1)} Unlike language models that use short prefixes (e.g., ``The issues ...") to trigger the generation, conversational models are constrained to the given dialogue history, which significantly influences the controllability. 
Given an open ended dialogue context (e.g., Table 11 in appendix), AD generates an impressively natural and on-topic response, but when provided a more constrained dialogue context (e.g., Table 17 in appendix), AD generates a response that may sound sudden and out of context.

\textbf{2)} Looking at the overall responses, also shown in Table~\ref{tab:responses_neg_pos}, we observe that the models use a restricted vocabulary for generating attribute-consistent responses. For example, AD frequently generates sentences containing ``horrible", ``terrible" or ``worst" for negative, while ``beautiful", ``happy" or ``wonderful" are more common for positive. 

\textbf{3)} The importance of mutual exclusivity of the attribute-specific vocabulary also explains the relatively poor performance when controlling for certain topics. As noted above, positive and negative vocabularies are clearly distinguishable. However, the attribute-specific words for topics such as Business are more generic (e.g., ``car", ``store") than those for other topics such as Sport (e.g., ``football", ``hockey") or Sci/Tech (e.g., ``android", ``software"). If the attribute-specific words are common and shared across multiple domains, the generated responses may not sound attribute specific even though the correct vocabulary is used. 

Note that this use of restricted vocabulary also harms fluency, because the vocabulary cannot always fit within a given context. Additional generated examples and statistics of attribute-specific vocabulary on each style/topic are provided in Appendix D. 

\section{Short Summary}
We explore plug-and-play methods for controlling the response generation of large pre-trained conversational models. With extensive automatic and human evaluations, we show that PPLM is able to generate fluent and attribute-consistent responses. Further, to overcome the significant computational overhead introduced by PPLM at decoding time, we optimize a tiny residual adapter for each attribute based on a few synthetic responses generated using PPLM. The resulting model does not require further computation at decoding time, and outperforms PPLM both in terms of fluency and attribute consistency.

\chapter{Controlling Dialogue Domains Continuously}

Task-oriented dialogue systems (TODs) are the core technology of the current state-of-the-art smart assistants (e.g., Alexa, Siri, Portal, etc.). These systems are either modularized as a pipeline of multiple components, namely, natural language understanding (NLU), dialogue state tracking (DST), dialogue policy (DP) and natural language generation (NLG), or end-to-end, where a single model implicitly learns how to issue APIs and system responses (NLG). 

\begin{figure}[t]
    \centering
    \includegraphics[width=0.7\linewidth]{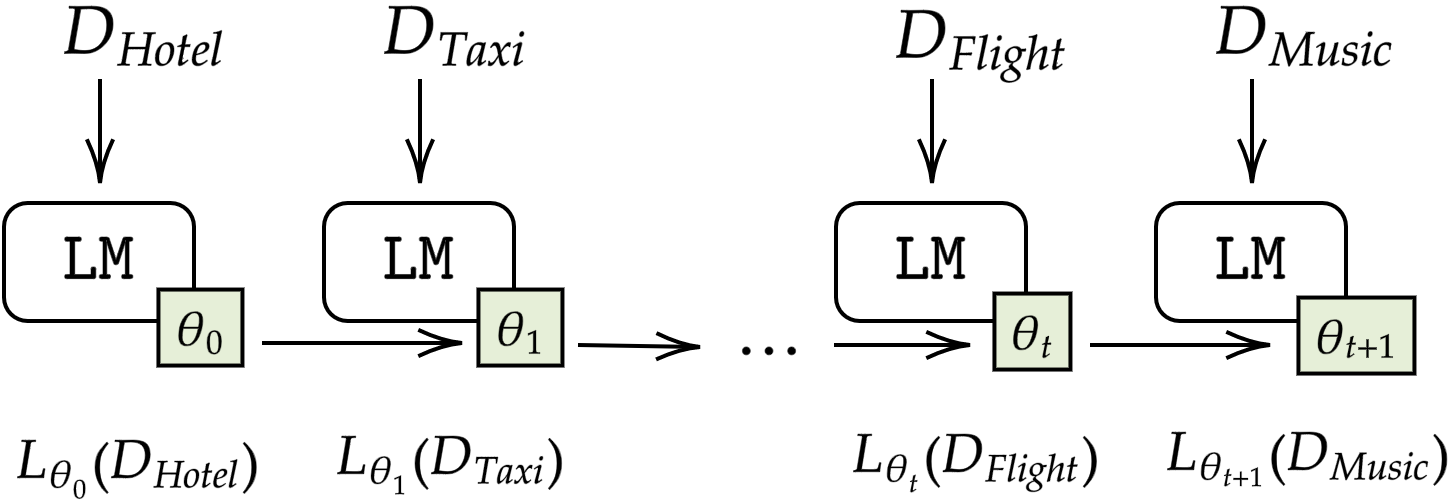}
    \caption{In continual learning, the model is trained one dataset at time. In this instance, the model is first trained on data from the hotel domain $D_{Hotel}$ then on the Taxi domain $D_{Taxi}$ and so on. The parameters of the model are updated sequentially based on the loss function $L$. }
    \label{fig:example}
\end{figure}

These systems are continuously updated with new features based on the user's needs, e.g., adding new slots and intents, or even completely new domains. However, existing dialogue models are trained with the assumption of having a fixed dataset at the beginning of the training, and they are not designed to add new domains and functionalities through time without incurring the high cost of retraining the whole system. The ability to acquire new knowledge continuously, a.k.a. continual learning (CL)~\cite{thrun2012learning}, represents a common challenge to many production and on-device dialogue systems where there is a continual growth of 1st and 3rd-party-developer domains that are added after deployment. Therefore, it is crucial to design dialogue systems with CL ability. Figure~\ref{fig:example} shows an high-level intuition of CL in TODs.

The main challenge in continual learning is \textit{catastrophic forgetting}~\cite{mccloskey1989catastrophic}. This phenomenon happens because there is a distributional shift between the tasks in the curriculum, which leads to catastrophic forgetting of the previously acquired knowledge. To overcome this challenge three kinds of methods are usually deployed: loss \textit{regularization}, which avoid interference with the previously learned tasks; \textit{rehearsal}, which use episodic memory to recall previously learned tasks; and \textit{architectural}, which add task-specific parameters for each learned task. However, architectural methods are usually not considered as a baseline, especially in sequence-to-sequence (Seq2Seq) generation tasks~\cite{sun2019lamol}, because they usually \textit{require a further step} during testing to select which parameter to use for the given task.  

To the best of our knowledge, continual learning in TODs~\cite{lee2017toward} is mostly unexplored or has been studied only in specific settings (e.g., NLG~\cite{mi2020continual}) using only a few tasks learned continuously. Given the importance of the task in the dialogue setting, we believe that a more comprehensive investigation is required, especially by comparing multiple settings and baselines. Therefore, we make the following contributions:
\begin{enumerate}[noitemsep]
    \item We propose a benchmark for continual learning in TODs, with 37 tasks to be learned continuously on four settings.
    \item We propose a simple yet effective architectural CL method based on residual adapters~\cite{houlsby2019parameter} that can continuously learn tasks without the need of a task classifier at testing time.
    \item We analyse the trade-off between number-of-parameters, episodic memory sizes, and training time of the three main categories of CL methods (regularization, rehearsal, architectural).
\end{enumerate}  

\section{Background}\label{sec:back}

\begin{figure}
    \centering
    \includegraphics[width=0.7\linewidth]{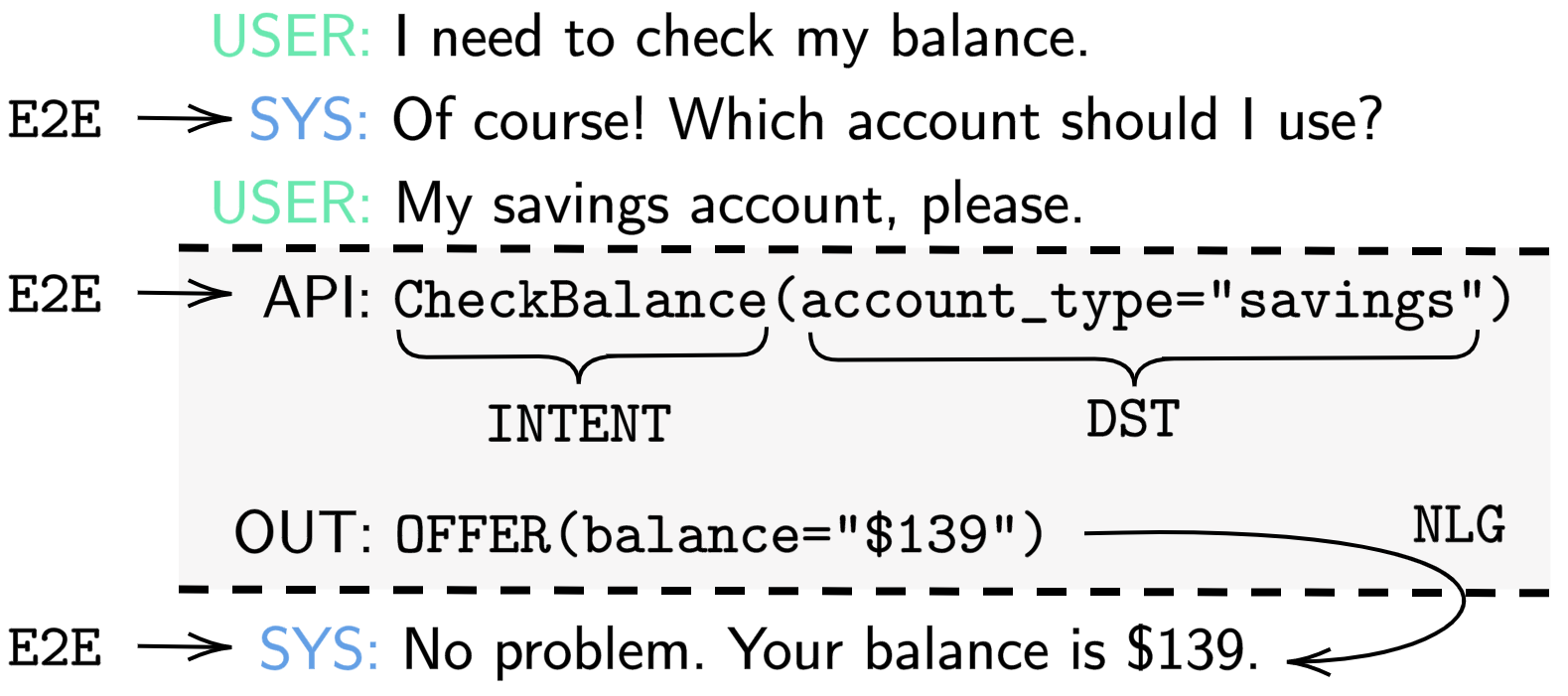}
    \caption{Example of input-out pairs, for the four settings, INTENT, DST, NLG and end-to-end (E2E). }
    \label{fig:example1}
\end{figure}

\subsection{Task-Oriented Dialogue Modelling} \label{subsec:preproc}
We model TODs as a Seq2Seq generation task~\cite{lei2018sequicity,byrne2020tickettalk} that generates both API-calls and system responses. As shown in Figure~\ref{fig:example1}, the model takes as input a dialogue history to generate an API-call, which is the concatenation of user intents and current dialogue states, and then uses its API-call returns, which can be empty or system speech-acts, to generate its system response. This modelling choice is guided by the existing annotated dialogue datasets, which provide the intent and the dialogue state of the user at every turn, and the speech-act of the system; and it allows us to define four distinct settings for studying CL: intent recognition (INTENT), DST, NLG and end-to-end (E2E). In the coming paragraphs, we formally describe the four settings as different input-out pairs for a Seq2Seq model. 

\paragraph{Data-Formatting} Let us define the dialogue history $U$ as a single sequence of tokens from the concatenation of the alternating utterances from the user and the system turns respectively. Without loss of generality, we assume that $U$ has all the dialogue history without the last system utterance, denoted as $S$. To distinguish between speakers, we add two special tokens at the beginning of every utterance: "\texttt{USER:}" for the user utterance, and "\texttt{SYSTEM:}" for the system utterance. Then, we define an API-call, denoted by $S_{API}$, as the concatenation of the API-name, i.e., the user-intent, and its arguments, i.e., slot-value pairs from the DST. The following syntax is used:  
\begin{equation}
S_{API} = \underbrace{\mathbf{I}}_{\text{Intent}} \ (\underbrace{s_{1}=v_{1}, \ldots,s_{k}=v_{p}}_{\text {Slot-value pairs}}), \label{API}
\end{equation}
where $\mathbf{I}$ is an intent or the API-name, $s_i$ the slot-name and $v_i$ one of the possible values for the slot $s_i$. The return of the API-call is either an empty string, thus the model uses the dialogue history to generate a response, or a speech-act, denoted as $S_{OUT}$, in the same format as the API-call in Equation~\ref{API}. Similar to the dialogue history, we define two special tokens "\texttt{API:}" and "\texttt{OUT:}" for triggering the model to generate the API-call and for distinguishing the return of the API from the dialogue history respectively. Based on this pre-processing, we define the four settings.

Without loss of generality, we define the three modularized settings by their input-out pairs: 

\[
\begin{array}{cc}
U \rightarrow \mathbf{I} & \mathrm{(INTENT)} \\
U \rightarrow \mathbf{I}(s_{1}=v_{1}, \ldots,s_{k}=v_{p}) & \mathrm{(DST)} \\
\underbrace{\mathbf{I}(s_{1}=v_{1}, \ldots,s_{k}=v_{p})}_{S_{OUT}}\rightarrow S & \mathrm{(NLG)},
\end{array}
\]
whereas for the E2E setting we define the pairs as:
\[
\begin{array}{c}
U \rightarrow \underbrace{\mathbf{I}(s_{1}=v_{1}, \ldots,s_{k}=v_{p})}_{S_{API}} \\
U + \underbrace{\mathbf{I}(s_{1}=v_{1}, \ldots,s_{k}=v_{p})}_{S_{OUT}}\rightarrow S,
\end{array}
\]
Often, $S_{OUT}$ is empty and thus the model maps the dialogue history to the response ($U \rightarrow S$), as we have seen in the previous chapter. An example of input-out pairs is shown in Figure~\ref{fig:example1}. 

Finally, we define a dialogue dataset as $D_K=\{(X_i,Y_i)\}_{i}^{r}$, where $(X_i,Y_i)$ is a general input-out pair from one of the four settings under consideration, and $K$ the dialogue domain under consideration (e.g., hotel).   

\paragraph{Model} We employ casual language models (e.g., GPT-2), which are often used in the current state-of-the-art task-oriented dialogue models, as in \citet{peng2020soloist} and \cite{hosseini2020simple}. Then, given the concatenation of the input $X= \{x_{0},\dots,x_{n}\}$ and output $Y= \{x_{n+1},\dots,x_{m}\}$ sequences, we compute the conditional language model distribution using the chain rule of probability~\cite{bengio2003neural} as
\begin{equation}
    p_{\theta}([X;Y]) =  \prod_{i=0}^{n+m}p_{\theta}(x_i|x_{0}, \cdots, x_{i-1}),
\end{equation}
where $\theta$ is the model's parameters and $[;]$ denotes the concatenation of $X$ and $Y$. The parameters are trained to minimize the negative log-likelihood over a dataset $D$ of input-out pairs, which in our case is the data of the four settings. Formally, we define the loss $\mathcal{L}_{\theta}$ as:
\begin{equation}
  L_{\theta}(D) =  - \sum_j^{|D|} \sum_{i=0}^{n+m} \log p_{\theta}(x_i^{(j)}|x_{0}^{(j)}, \cdots, x_{i-1}^{(j)}), \label{eq:loss}
\end{equation}
where $n+m$ is a maximum sequence length in $D$. At inference time, given the input sequence $X$, the model parameterized by $\theta$ autoregressively generates the output sequence $Y$. 
\subsection{Continual Learning} \label{subsec:CL}
The goal of continual learning is to learn a set of tasks sequentially without catastrophically forgetting the previously learned tasks. In TODs, we cast CL as learning a sequence of domains sequentially. Let us define a curriculum of $T$ domains as an ordered set $\mathcal{D}=\{D_1, \cdots, D_T\}$, where $D_k$ is a dataset under the domain $K$. In addition, we denote the models' parameters after learning the task $K$ by $\theta_K$. 

Following the recently defined taxonomy for CL~\cite{wortsman2020supermasks}, we study the settings in which the task-ID, is provided during training, but not during testing,~\footnote{GNs: Task given during training, no inference; shared labels.} meaning that, during training, the model is aware of which domain it is currently learning, but during testing, the model is evaluated without specifying the dialogue domain. This assumption makes our CL setting more challenging but more realistic, since during inference times, users do not explicitly specify in which domain they want to operate. 

We consider three CL approaches: \textit{regularization},
\textit{rehearsal} and \textit{architectural}. In our experiments, we describe the most commonly used methods within each approach, especially those known to work well in language tasks.
\begin{itemize}[noitemsep]
    \item \textit{Regularization} methods add a regularization term to the current learned $\theta_t$ to avoid interfering with the previously learned $\theta_{t-1}$. Formally, the loss at task $t$ is:
    \begin{equation}
        L_{\theta_t}(D_t) = L_{\theta_t}(D_t) + \lambda \Omega (\theta_t - \theta^*_{t-1})^2,
    \end{equation}
    where $\theta^*_{t-1}$ is copies of the previously learned parameters frozen at this stage. In our experiments, we consider two kinds of $\Omega$: the identity function (\textbf{L2}) and the Fisher information matrix~\cite{kirkpatrick2017overcoming} (\textbf{EWC}).  
    
    \item \textit{Rehearsal} methods use an episodic memory $M$ to store examples from the previously learned domains, and re-use them while learning new tasks. The most straightforward method is to add the content of the memory $M$ to the current task data $D_t$~\cite{robins1995catastrophic}. Following our notation, the model is optimized using $L_{\theta_t}(D_t+M)$, and we refer to this method as \textbf{REPLAY}. Another rehearsal method is to constrain the gradients updates so that the loss of the samples in memory never increases. More formally,
    \begin{equation}
        L_{\theta_t}(D_t) \ \text{s.t.} \ L_{\theta_t}(M) \leq L_{\theta_{t-1}}(M).
    \end{equation}
    Of this kind, the method Gradient Episodic Memory (GEM)~\cite{lopez2017gradient} computes the gradient constraint via a quadratic programming solver that scales with the number of parameters of the model. After our first investigation, we discover that it is impractical for large-language models to use GEM, since they have millions of parameters and the constraints are computed for each batch. To cope with this computational complexity, \citet{chaudhry2018efficient} proposed \textbf{A-GEM}, which efficiently computes the gradient constraints while being effective in CL tasks. Finally, a rehearsal method specific to language tasks is \textbf{LAMOL}~\cite{sun2019lamol}, which instead of storing samples in $M$, trains a model that simultaneously learns to solve tasks and generate training samples.
    \item \textit{Architectural} methods add task-specific parameters to an existing base model for each task. Of this kind, multiple models have been proposed, such as Progressive Net~\cite{rusu2016progressive}, Dynamically Expandable Networks (DEN)~\cite{yoon2017lifelong} and Learn-to-Grow~\cite{li2019learn}. On the other hand, there are fixed-capacity methods, that do not add specific parameters, but learn parameter masks~\cite{fernando2017pathnet}, usually binary~\cite{mallya2018piggyback}, to select sub-networks that are task-specific. To the best of our knowledge, these models have been tested mostly on computer vision tasks, and they can not easily handle our CL setting (i.e., no task-ID during testing).
\end{itemize}

\section{AdapterCL} \label{sec:adpt}
Motivated by the lack of architectural baselines for CL in Seq2Seq modelling, we propose a novel architectural method called AdapterCL. Our proposed method parameterizes each task using residual adapters~\cite{houlsby2019parameter}, as we have seen in the previous chapter, and it uses an entropy-based classifier to select which adapter to use at testing time. This method is designed for large pre-trained language models, e.g., GPT-2, since only the task-specific parameters are trained, while the original weights are left frozen. 

\begin{figure}[t]
    \centering
    \includegraphics[width=0.8\linewidth]{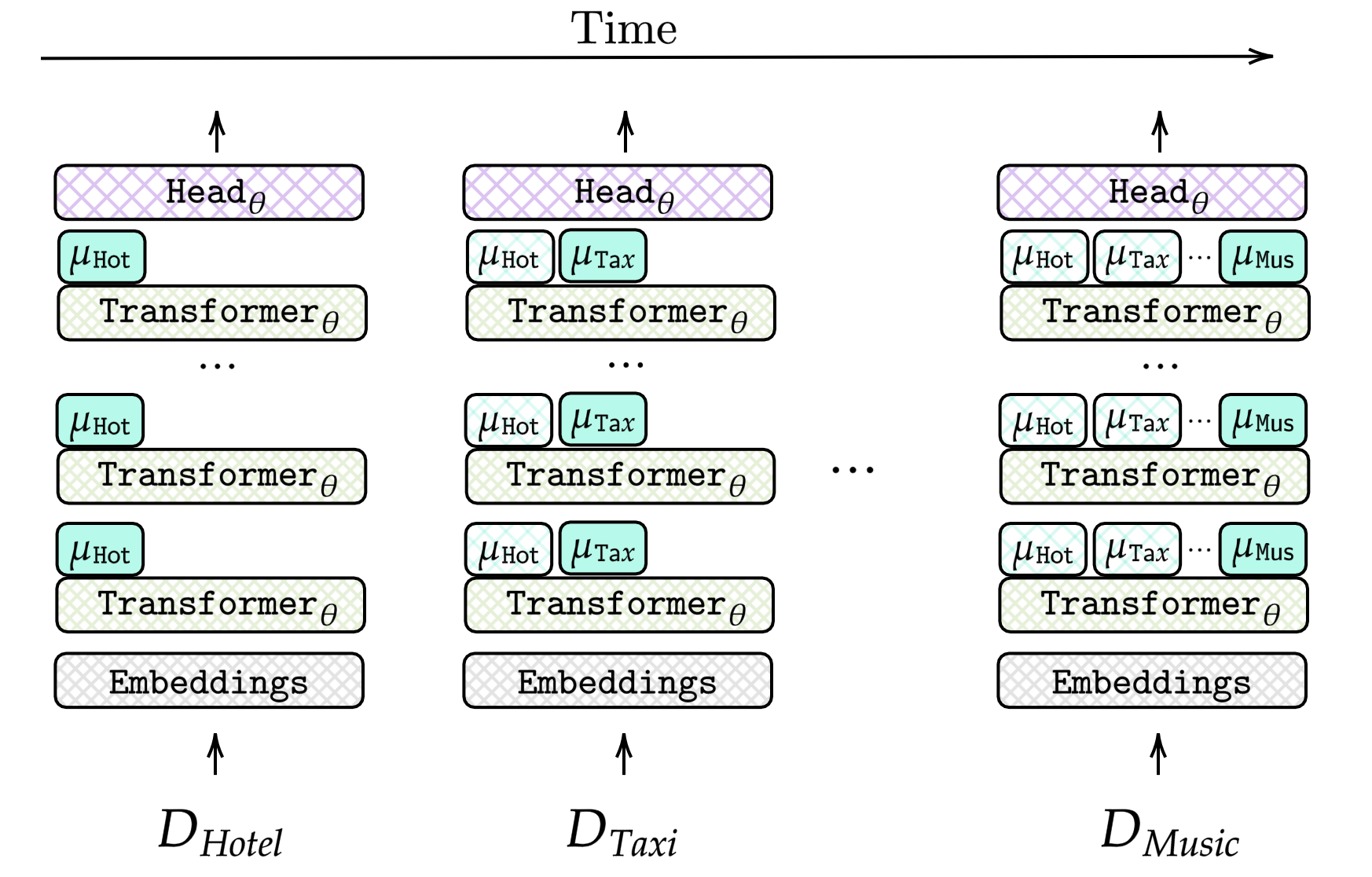}
    \caption{Example of CL with AdpaterCL. Every time a new domain dataset $D$ arrives, a new adapter is spawned, and it is fine-tuned with the data pairs, while the original weights are left frozen.}
    \label{fig:adapterc_learning}
\end{figure}


\paragraph{Residual adapters} To continuously learn new tasks, we first spawn a new adapter, parameterized by $\mu$, and then we train its parameters as in Equation~\ref{eq:loss}. For instance, given the dataset $D_t$ and the model with its corresponding adapter $\mu_t$, the loss is defined as:
\begin{equation}
  L_{\mu_t}(D_t) =  - \sum_j^{|D_t|} \sum_{i=0}^{n+m} \log p_{\mu_t}(x_i^j|x_{0}^j, \cdots x_{i-1}^j).
\end{equation}
Importantly, the loss is optimized over $\mu_t$ to guarantee that each task is independently learned. An example of AdapterCL while learning tasks continuously is shown in Figure~\ref{fig:adapterc_learning}.

\begin{figure}[t]
    \centering
    \includegraphics[width=0.9\linewidth]{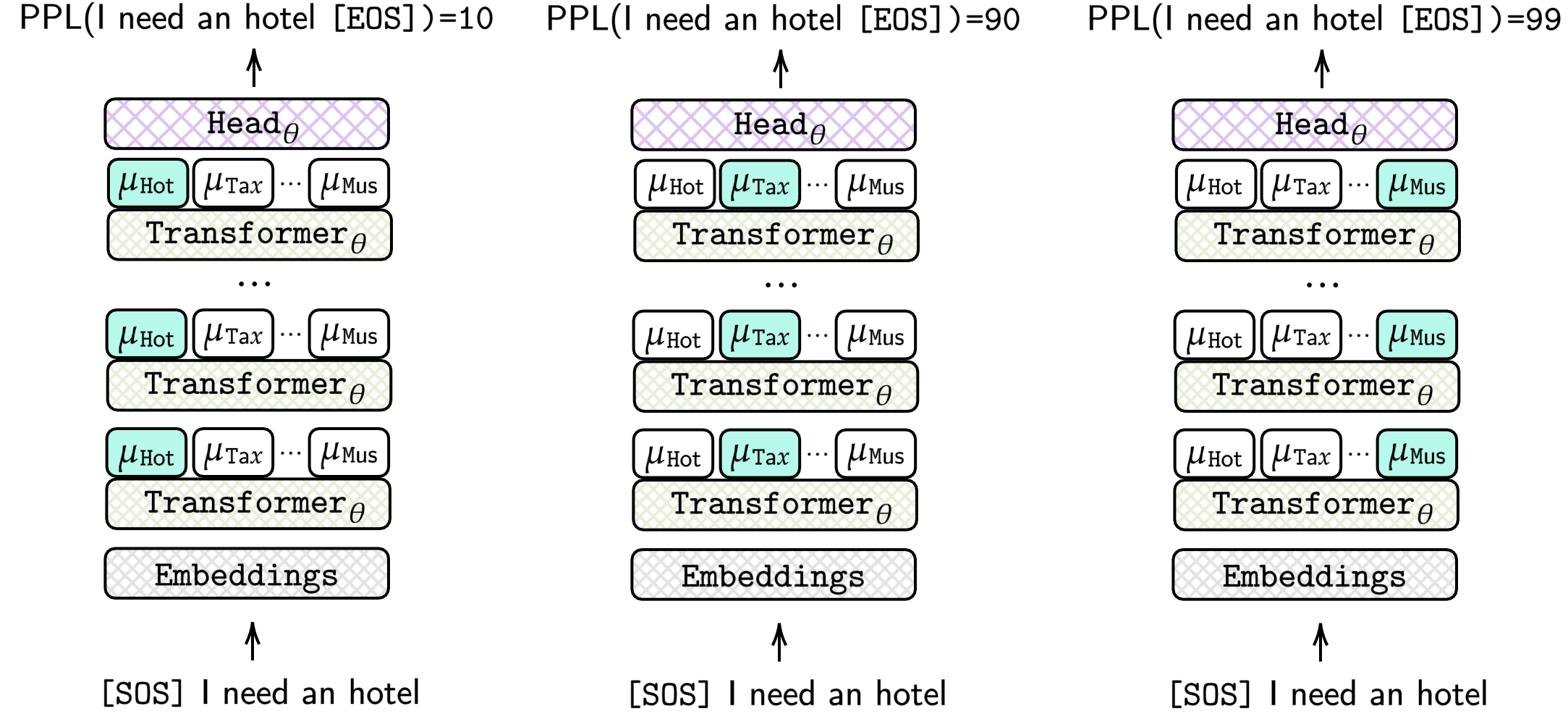}
    \caption{Example of the perplexity-based classifier. Given the sentence "I need an hotel", we forward each of the adapters to compute the perplexity score. In this example, the HOTEL adapter achieves the lowest perplexity, and it is selected to generate the output sequence.}
    \label{fig:adaptercl}
\end{figure}

\paragraph{Perplexity-Based Classifier} In our CL setting the task-ID is provided during training, and thus each $\mu_t$ is optimized over $D_t$. During testing, however, the task-ID is not provided, and thus the model has to predict which adapter to use to accomplish the task. This step is not required in regularization and rehearsal approaches since a single set of parameters is optimized during training.

Inspired by \citet{wortsman2020supermasks}, we propose to utilize the perplexity of each adapter over the input $X$ as a measure of uncertainty. Thus, by selecting the adapter with the lowest perplexity, we select the most confident model to generate the output sequence. The perplexity of an input sequence $X=x_{0}, \cdots, x_{n}$ is defined as
\begin{equation}
    \mathrm{PPL}_{\theta}(X)=\sqrt[n]{\prod_{i=1}^{n} \frac{1}{p_{\theta}\left(x_{i} \mid x_{0}, \cdots, x_{i-1}\right)}}
\end{equation}
Therefore, given the set of adapters parameterized by $\mu_0,\dots, \mu_N$, each of which is trained respectively with $D_0, \dots, D_N$, and an input sample $X$, we compute:
\begin{equation}
   \alpha_t = \mathrm{PPL}_{\mu_t}(X) \ \forall t \in 1, \cdots, N,\label{eq:alpha}
\end{equation}
where each $\alpha_t$ represents the confidence of the adapter $t$ for the input $X$. The task-ID $t$ is thus selected as 
\begin{equation}
    t^* = \operatorname*{argmin}{\alpha_0,\cdots,\alpha_N},\label{eq:selection}
\end{equation}
The perplexity-based selector requires a linear number of forwards with respect to the number of adapters (Equation~\ref{eq:alpha}), but it has the advantage of not requiring a further classifier, which itself would suffer from catastrophic forgetting and would require episodic memory. An example of perplexity-based adapter selection is shown in Figure~\ref{fig:adaptercl}.

\section{Experimental Settings} \label{sec:experiment}
In this section we describe 1) the datasets used for creating the learning curriculum, 2) the evaluation metric used to evaluate the different settings, and 3) the experimental setups.

\subsection{Datasets}
To the best of our knowledge, there is no benchmark for CL in dialogue systems with a high number of tasks to be learned continuously and with multiple training settings. The closest to ours is the method from~\citet{mi2020continual}, which continuously learns five domains in the NLG setting. In general, NLP benchmarks for CL use no more than 10 tasks~\cite{sun2019lamol,d2019episodic}. Consequently, we propose a CL benchmark in which we jointly pre-processing four task-oriented datasets: Task-Master 2019 (TM19)~\cite{byrne-etal-2019-taskmaster}, Task-Master 2020 (TM20)~\cite{byrne-etal-2019-taskmaster}, Schema Guided Dialogue (SGD)~\cite{rastogi2019towards} and MultiWoZ~\cite{budzianowski2018large}. This results in a curriculum of 37 domains to be learned continuously under four settings: INTENT classification, DST, NLG, and E2E. This is possible because the four datasets provide the speech act annotation for both the user and the system turns, and the dialogue state as well. To avoid any domain overlapping, we select only the dialogues with a single domain, and we do not merge domains with similar/the same names or semantics. For example, the \textit{restaurant} domain appears in TM19, TM20 and MWOZ, with different slot names and values. Thus, we intentionally keep these data samples separate for modelling scenarios in which multiple APIs are available for the same domain. 

Finally, the datasets are pre-processed as in Section~\ref{subsec:preproc} to form the four tasks, and the main statistics are shown in Table~\ref{tab:mainstat}. 

\begin{table}[t]
\centering
\begin{tabular}{r|ccc|ccc}
\hline
\textbf{Name}                          & \textbf{Train} & \textbf{Valid} & \textbf{Test} & \textbf{Dom.} & \textbf{Intent} & \textbf{Turns} \\ \hline
TM19 & 4,403           & 551            & 553           & 6             & 112           & 19.97          \\
TM20 & 13,839          & 1,731           & 1,734          & 7             & 128           & 16.92          \\
MWoZ & 7,906           & 1,000           & 1,000          & 5             & 15            & 13.93          \\
SGD & 5,278           & 761            & 1,531          & 19            & 43            & 14.71          \\ \hline
Total                         & 31,426          & 4,043           & 4,818          & 37            & 280           & 16.23          \\ \hline
\end{tabular}
\caption{Main dataset statistics. }
\label{tab:mainstat}
\end{table}

\subsection{Evaluation Metrics}
Automatic evaluations for E2E TODs are challenging, especially for the response generation task. To overcome this issue, we use well-defined metrics based on the three modularized settings. In all of the three sub-tasks, we define the relevant metrics as follows:
\begin{itemize}
    \item \textit{INTENT} recognition is evaluated using the \textit{accuracy} between the generated intents and the gold labels. 
    \item \textit{DST} is evaluated with the Joint Goal Accuracy (JGA)~\cite{wu2019transferable} over the gold dialogue states.
    \item \textit{NLG} is evaluated using both the BLEU score~\cite{papineni-etal-2002-bleu} and the slot error rate (EER)~\cite{wen2015semantically}, which is computed as the ratio between the total number of slots and the values not appearing in the response. In datasets such as SGD, the slot has binary values, e.g., yes or no, and thus we exclude these from the count, as in ~\citet{kale2020few}. In the E2E setting, if the API output ($X_{out}$) is empty, then we rely on the BLEU score.  
\end{itemize}
Independently of these metrics, we also compute CL-specific metrics such as the average metric through time (Avg. Metric), as in ~\citet{lopez2017gradient}. We consider access to the test set for each of the $T$ tasks, and after the model finishes learning the task $t_i$, we evaluate its test performance on all tasks in the curriculum. To elaborate, we construct the matrix $R\in \mathbb{R}^{T\times T}$, where $R_{i,j}$ is the test metric (e.g., BLEU, JGA) of the model on task $t_j$ after observing the last sample from task $t_i$. Then we define the average accuracy as
\begin{equation}
    \mathrm{Avg. Metric} =\frac{1}{T} \sum_{i=1}^{T} R_{T, i}, \label{eq:avgmetric}
\end{equation}
The Avg. Metric score is useful for understanding the learning dynamics through time of different baselines. Further metrics such as Backward-Transfer and Forward-Transfer~\cite{lopez2017gradient} are available to distinguish baselines with similar Avg. Metric scores, but we limit our evaluation to this metric, since there is a large gap among the baselines. Finally, to evaluate the adapter selection, we use the accuracy over the gold task-ID.

\begin{table*}[t]
\centering
\resizebox{\linewidth}{!}{
\begin{tabular}{r|ccc|c|c|cc}
\hline
                                     & \multicolumn{1}{l}{} & \multicolumn{1}{l}{} & \multicolumn{1}{l|}{}  & \textbf{INTENT}              & \textbf{DST}            & \multicolumn{2}{c}{\textbf{NLG}}                     \\ \hline
\multicolumn{1}{c|}{\textbf{Method}} & \textbf{+Param.}     & \textbf{Mem.}         & \textbf{Hours}$\downarrow$         & \textit{Accuracy}$\uparrow$ & \textit{JGA}$\uparrow$ & \textit{EER}$\downarrow$ & \textit{BLEU}$\uparrow$ \\ \hline
\textit{VANILLA}               & -                    &$\emptyset$         & 0.21 $\pm$ 0.02 &4.1  $\pm$ 1.4              &4.91 $\pm$ 4.5         &48.7 $\pm$ 3.9           &6.38 $\pm$ 0.6          \\
\textit{L2}                    & $|\theta|$          &$\emptyset$          & 0.56 $\pm$ 0.06 &3.8  $\pm$ 1.4              &3.81 $\pm$ 3.4         &55.7 $\pm$ 7.1            &5.4 $\pm$ 0.9           \\
\textit{EWC}                   & $2|\theta|$        &$\emptyset$           & 0.91 $\pm$ 0.10 &3.9  $\pm$ 1.3              &5.22 $\pm$ 4.5         &58.2  $\pm$ 3.7           &5.06 $\pm$ 0.5          \\
\textit{AGEM}                  & -                    &$t|M|$              & 0.38 $\pm$ 0.04 &34.0 $\pm$ 6.4             &6.37 $\pm$ 4.0         &62.1 $\pm$ 6.9            &4.54 $\pm$ 0.6          \\
\textit{LAMOL}                 & -                    &$\emptyset$         & 2.32 $\pm$ 1.24 &7.5   $\pm$ 6.4            &4.55 $\pm$ 3.5        &66.1 $\pm$ 6.9             &3.0 $\pm$ 0.9           \\
\textit{REPLAY}                & -                    &$t|M|$              & 0.62 $\pm$ 0.23 &81.1 $\pm$ 1.4             &30.33$\pm$ 1.2 &\textbf{17.8} $\pm$ 0.9         &\textbf{17.4} $\pm$ 0.7         \\
\textit{ADAPT}                 &$t|\mu|$            &$\emptyset$           & \textbf{0.20} $\pm$ 0.02 &      \textbf{90.5} $\pm$ 0.6    &\textbf{35.1} $\pm$ 0.5       &31.78 $\pm$ 1.3            &16.76 $\pm$ 0.4         \\ \hline
\textit{MULTI}                 & -                    & -                  & 4.14 $\pm$ 2.23 &95.5 $\pm$ 0.1            &48.9 $\pm$ 0.2        &12.56 $\pm$ 0.2             &23.61 $\pm$ 0.1         \\ \hline
\end{tabular}
}

\caption{E2E results in terms of Intent Accuracy, Joint Goal Accuracy (JGA), Slot Error Rate (EER) and BLUE. +Param. shows the additional number of parameters per task ($\theta$ base model and $\mu$ task-specific parameters), and Mem. the episodic memory size (denoted as $|M|$) needed per task, and Hours is the average hours per epoch (single NVIDIA 2080Ti)required for training a new domain.}
\label{tab:e2e}
\end{table*}

\subsection{Baselines and Settings}
The main goal is to compare the performance of different CL approaches and to understand the trade-offs among them. Therefore, following the definition provided in Section~\ref{subsec:CL}, we compare 1) EWC and L2, 2) A-GEM, LAMOL, and REPLAY, and 3) AdapterCL. Additionally, we provide a baseline trained on each task continuously, namely, VANILLA, without any regularization or memory, and a multitask baseline (MULTI), which is trained on all the data in the curriculum at the same time. In L2, EWC, and A-GEM, we tune different $\lambda$ in the range 0.0001 to 100, and in the \textit{rehearsal}-based methods, REPLAY and GEM, we keep 50 samples per task, for a total of 1,850 samples in $M$ at the end of the curriculum. This is particularly important since, if we store in memory all the samples of the seen tasks, the model incurs a high training cost. Arguably, this could be an option if the per-task sample size is small, but this is not always possible, e.g, large language models~\cite{brown2020language}. Therefore, the assumption of minimizing the number of samples in memory is valid and is widely used in the CL literature~\cite{mi2020continual}. Finally, for the AdapterCL, we tune the bottleneck size $b$ between 10, 50, 100, and 200. Interested readers can refer to Appendix A for further details of the selected hyper-parameters. In CL the model is not able decide the order of tasks. Therefore, we create five learning curricula by randomly permuting the 37 tasks.

\begin{figure}[t]
\centering
\begin{subfigure}{.5\textwidth}
    \centering
    \includegraphics[width=\textwidth]{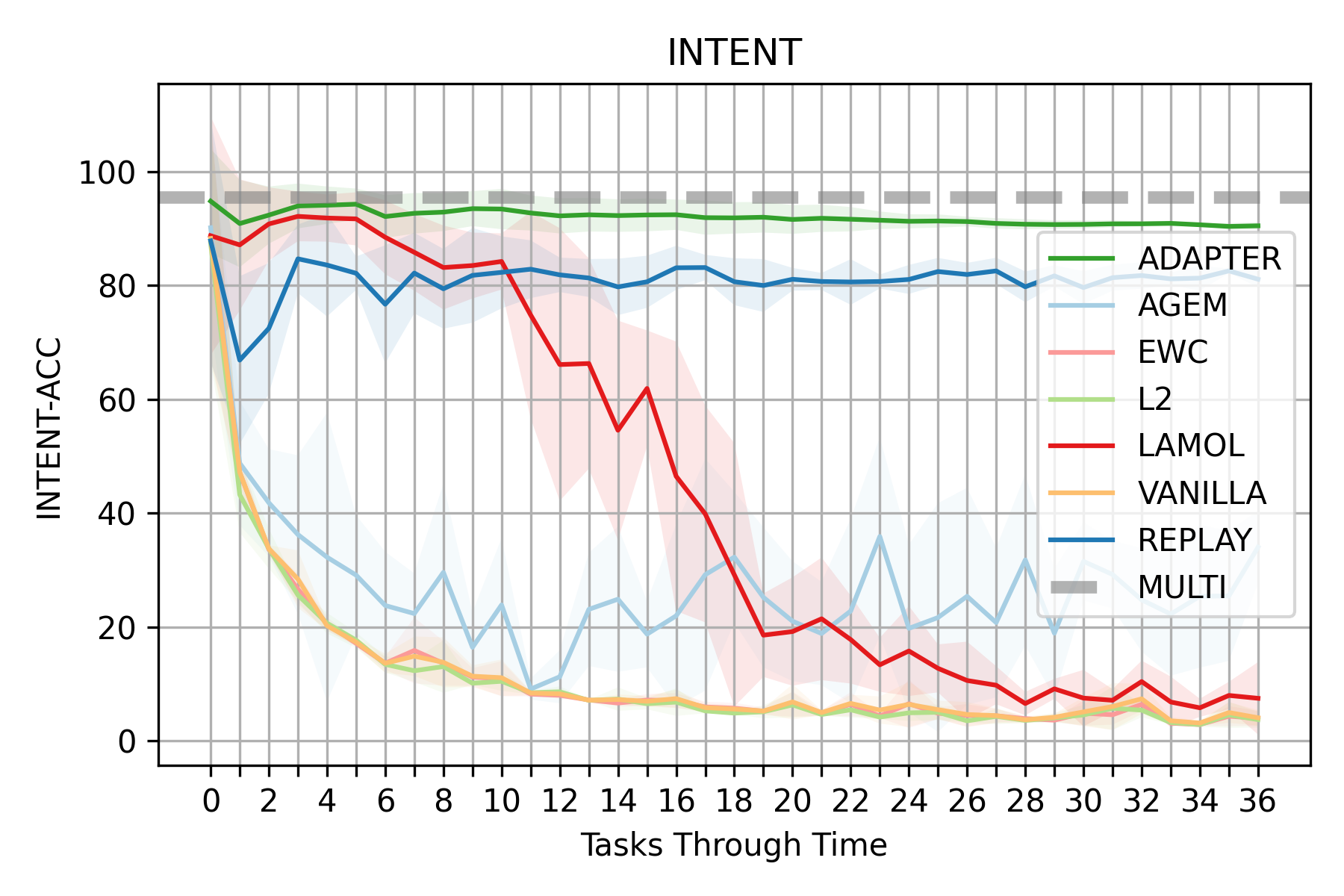}
\end{subfigure}%
\begin{subfigure}{.5\textwidth}
    \centering
    \includegraphics[width=\textwidth]{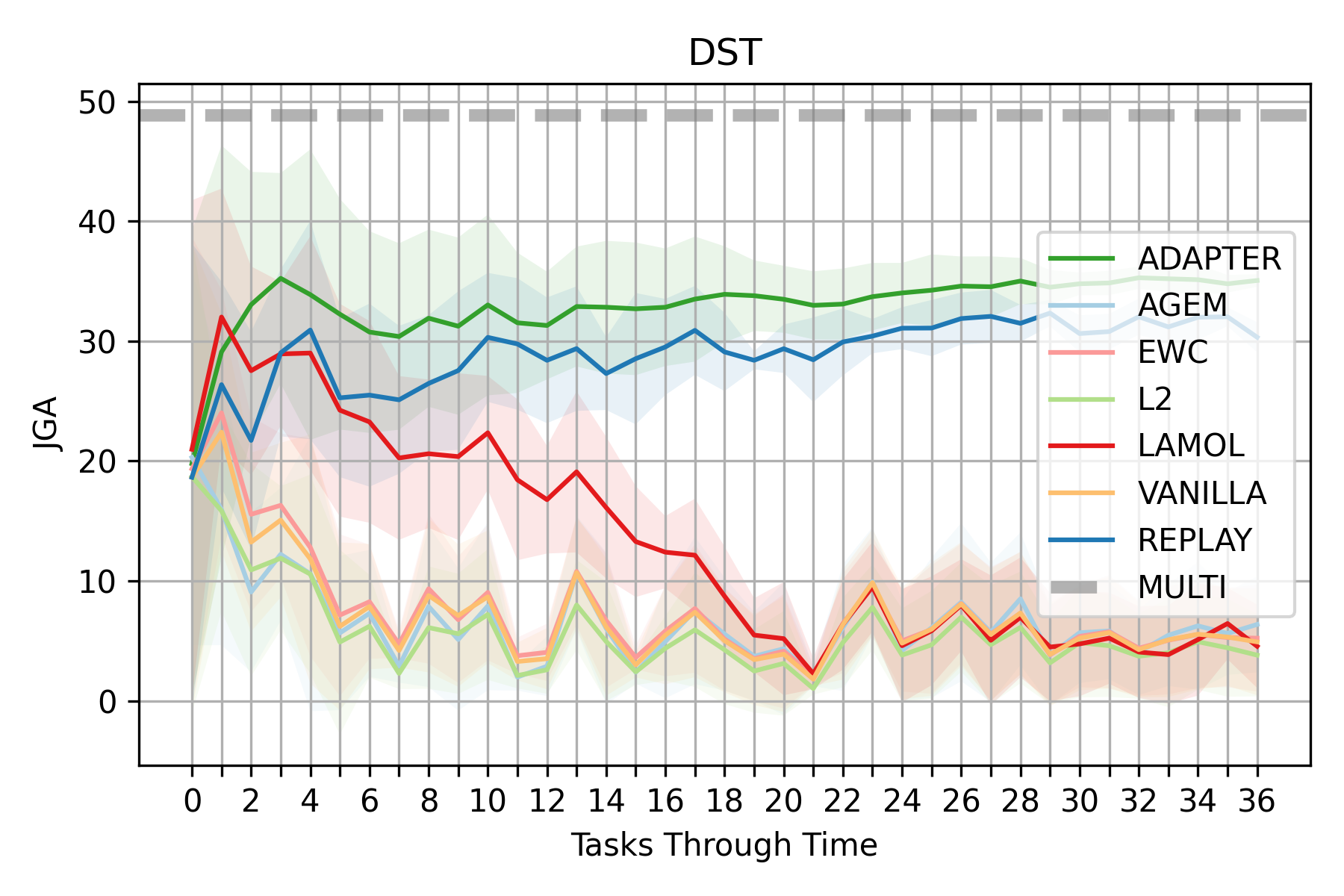}
\end{subfigure}
\begin{subfigure}{.5\textwidth}
    \centering
    \includegraphics[width=\textwidth]{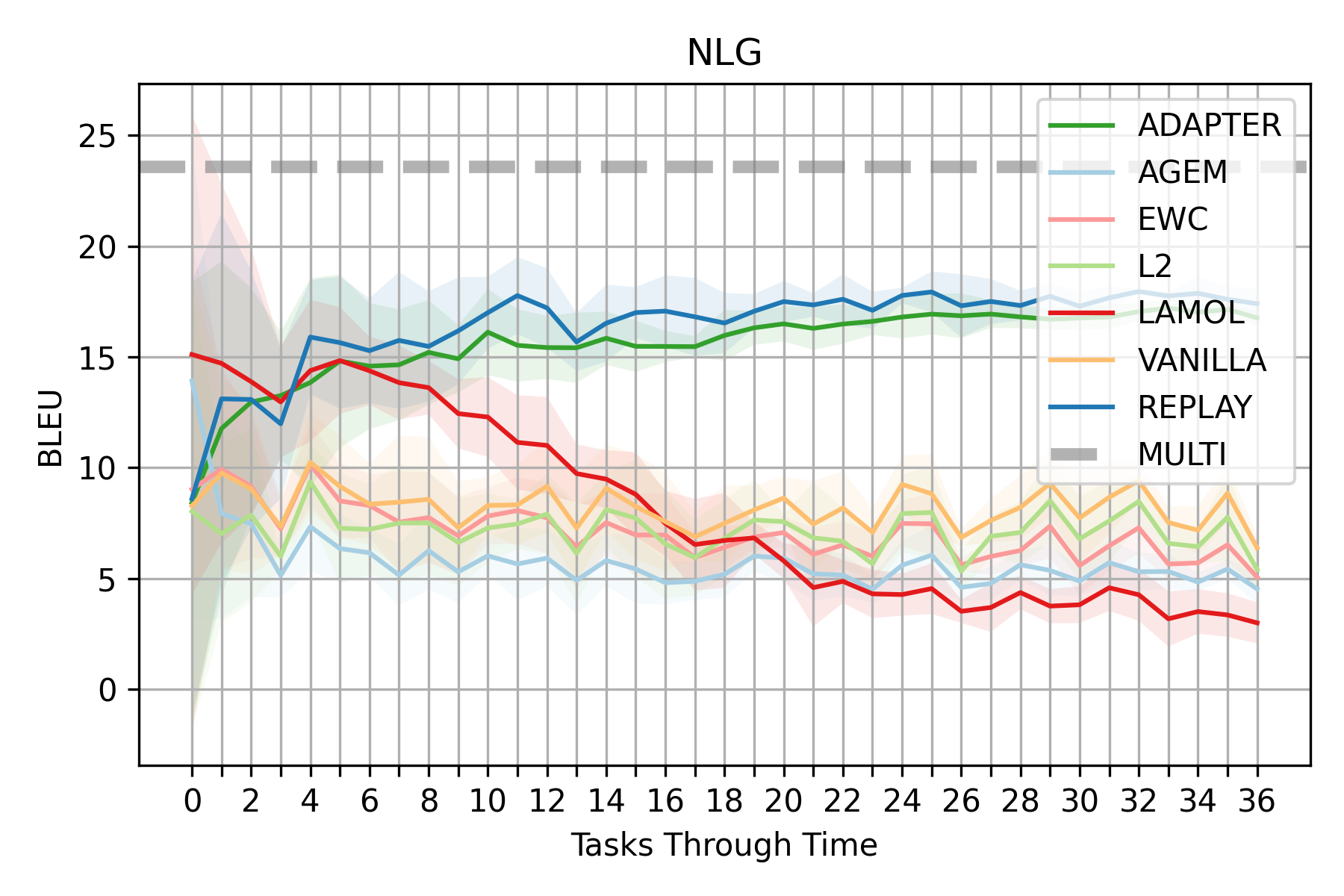}
\end{subfigure}%
\begin{subfigure}{.5\textwidth}
    \centering
    \includegraphics[width=\textwidth]{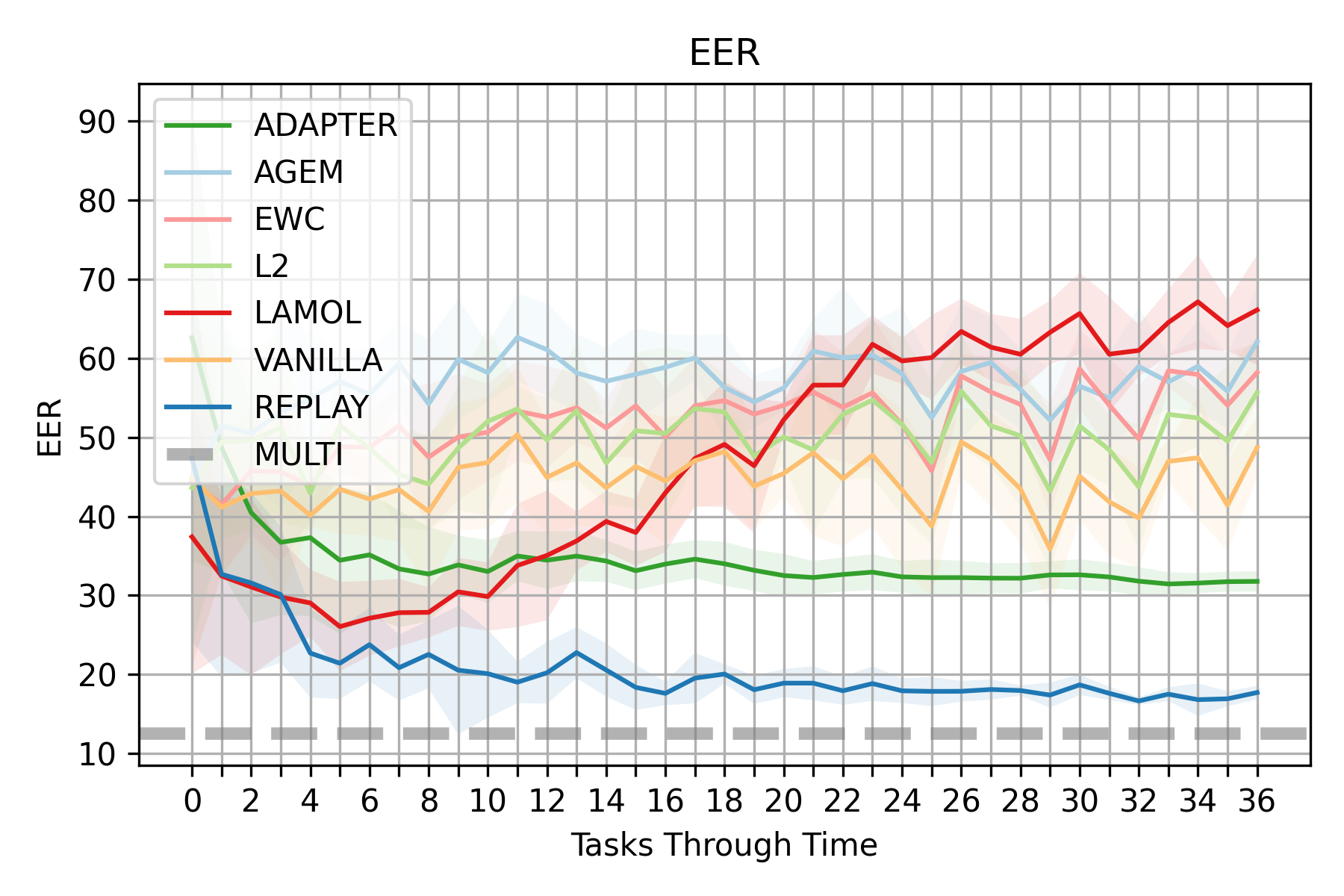}
\end{subfigure}
\caption[short]{Avg. Metric for Intent Accuracy, JGA, BLEU and EER in the E2E setting.}
\label{fig:E2E}
\end{figure}

\section{Results \& Analysis}\label{sec:results}
The main results in the E2E setting are summarized in Table~\ref{tab:e2e}
In these tables we report the Avg. Metric at the end of the curriculum, which is equivalent to the average test set performance in all the tasks, and the resources used by each model. 

\paragraph{Main Results} From the tables, we can observe that 1) both regularization-based methods (L2/EWC) and some rehearsal-based methods (AGEM/LAMOL) cannot continually learn tasks without incurring in catastrophic forgetting, 2) REPLAY and AdapterCL perform comparably well on the Intent and DST tasks, 3) REPLAY works the best on the NLG task, showing that transferring knowledge between tasks is needed, and 4) no CL methods can reach the performance of the multi-task baseline, especially on the DST task. In addition, the adapter selection accuracy based on Equation~\ref{eq:selection} is 95.44$\pm$0.2\% in E2E, 98.03$\pm$0.1\% in Intent Recognition, 98.19$\pm$0.1\% in DST, and 93.98$\pm$0.1\% in the NLG.

Although these numbers are meaningful, they do not describe the entire learning history of the curriculum. To better understand these dynamics, we plot the Avg. Metric in Equation~\ref{eq:avgmetric} after each task is learned ($t=T$ in the equation). Figure~\ref{fig:E2E} shows the plot for the considered metrics and all the baselines. From this figure we can better understand how REPLAY and AdapterCL outperform the other baselines and, interestingly, that LAMOL performs as well as REPLAY on the first 12 tasks. This is because LAMOL learns to generate training samples instead of using an explicit memory, and thus the generation becomes harder when more and more task are shown. This result further strengthens our motivation to have a benchmark with a long curriculum.

\begin{figure}[t]
    \centering
    \includegraphics[width=0.7\linewidth]{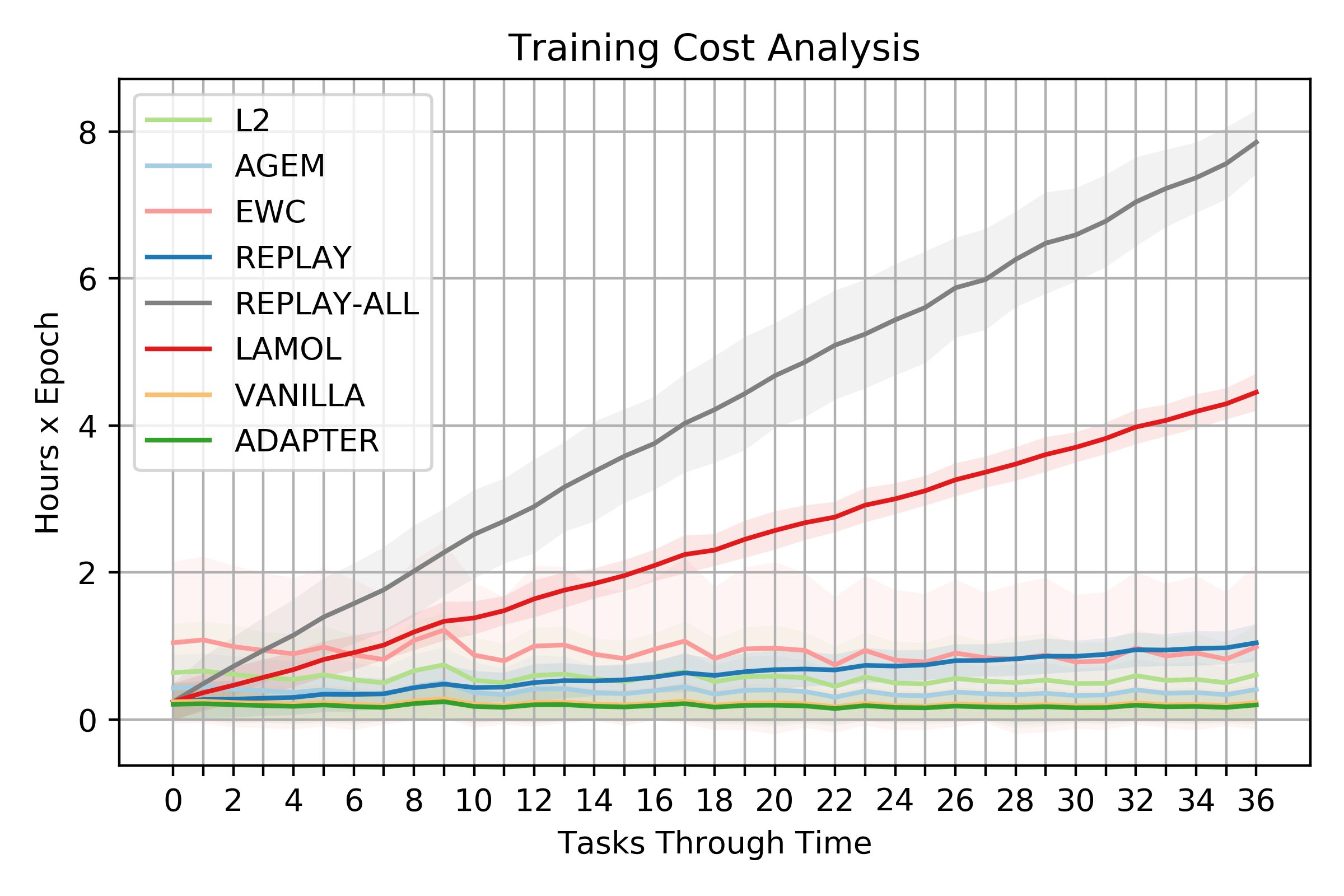}
    \caption{Training time for adding a new domain for different CL techniques.}
    \label{fig:time_analysis}
\end{figure}

\subsection{Training Time Analysis}
From Figure~\ref{fig:time_analysis} we plot the training time (Hours $\times$ Epochs) required to add a new domain to an existing model. A clear trend is shown where rehearsal based methods (REPLAY, LAMOL) requires a linearly increasing amount of time to add new domains, while ADAPTER and VANILLA the time remain constant across time. This is even more evident when the entire training-set of all the previous tasks is used for training (REPLAY-ALL), which lead to a very expensive process to add new domains. The average time across domain for all the baseline is shown in Table~\ref{tab:e2e}. ADAPTER based CL requires also an additional cost in selecting which parameters to use during testing. By using a single 2080ti, the average time to select the adapter is $0.069\pm 0,003$ seconds, which is as expensive as decoding 4 tokens.

\begin{figure}[t]
    \centering
    \includegraphics[width=0.9\linewidth]{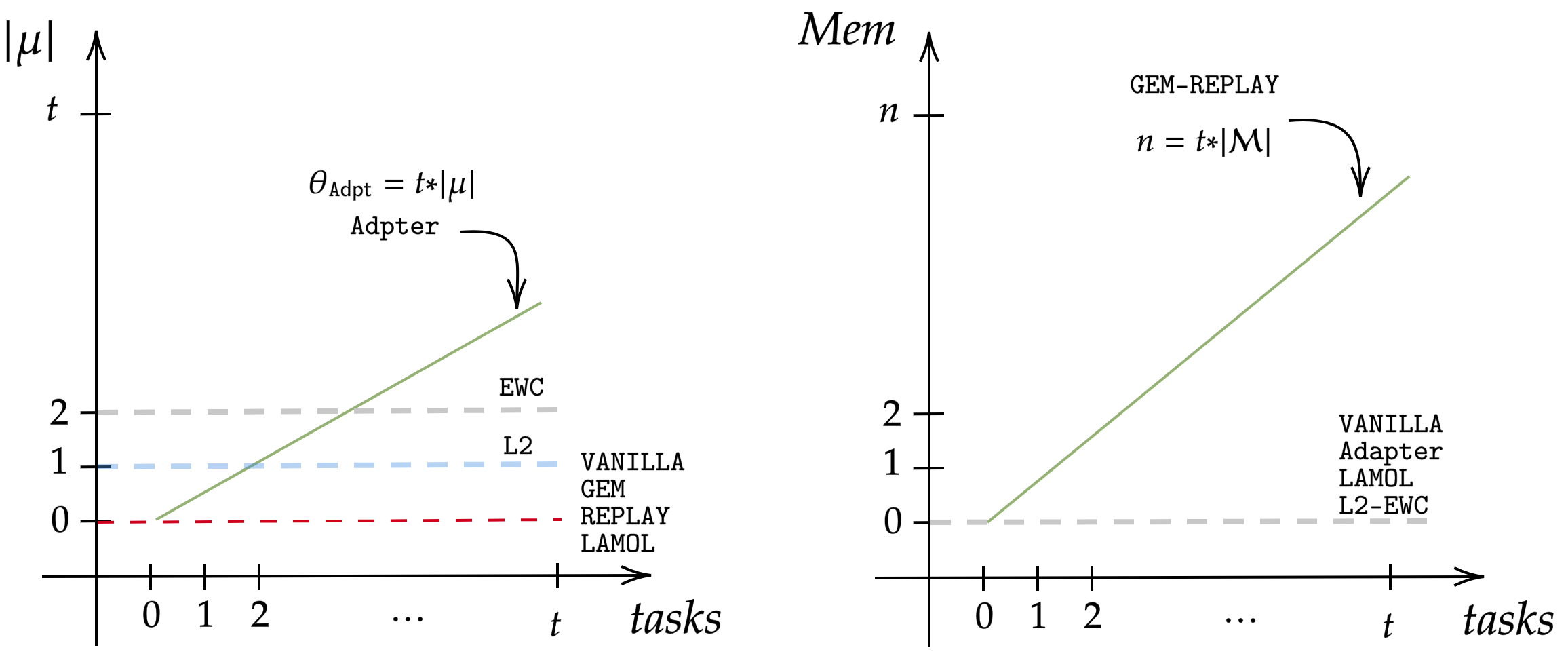}
    \caption{"No free lunch" in CL. Plot of the trade-off between number of parameters added per task and the size of the episodic memory $M$.}
    \label{fig:freelunch}
\end{figure}
\subsection{No Free Lunch} Based on the results shown in Table~\ref{tab:e2e}, and especially based on the resources used by each method, we conclude that there is a no free lunch in terms of resources needed to avoid the catastrophic forgetting problem. To elaborate, in both REPLAY and AdapterCL, the resources used grow linearly with the number of tasks; i.e., in REPLAY the number of samples stored in the episodic memory grows linearly (50 times the number of tasks), and in AdapterCL the number of parameters grows linearly (number of adapter parameters times the number of tasks). Figure~\ref{fig:freelunch} describes the high-level intuition behind this concept by plotting the number of tasks and parameters and the episodic memory sizes needed.

Therefore, given a resource budget, different baselines are preferable in terms of memory or parameters. The main advantage of using memory-based methods (e.g., REPLAY) is that no parameters are added, and thus the resulting model is closer to the multitask baseline. However, this comes with the disadvantage of losing the learned weights of the original pre-trained model. This is particularly critical for large pre-trained language model which provide a good starting point for fine-tuning new tasks. On the other hand, the main advantage of parameter-isolation methods (e.g., AdapterCL) is the ability to retain the original weights and to control which tasks to trigger, given a certain input. The latter is important in scenarios where just a subset of a domain is shown to the user (e.g., only one particular restaurant API). The main disadvantage, however, is the lack of knowledge transfer among tasks, since each dataset is trained in isolation. 

\begin{figure}[t]
    \centering
    
    \begin{subfigure}{.5\textwidth}
    \centering
        \includegraphics[width=\textwidth]{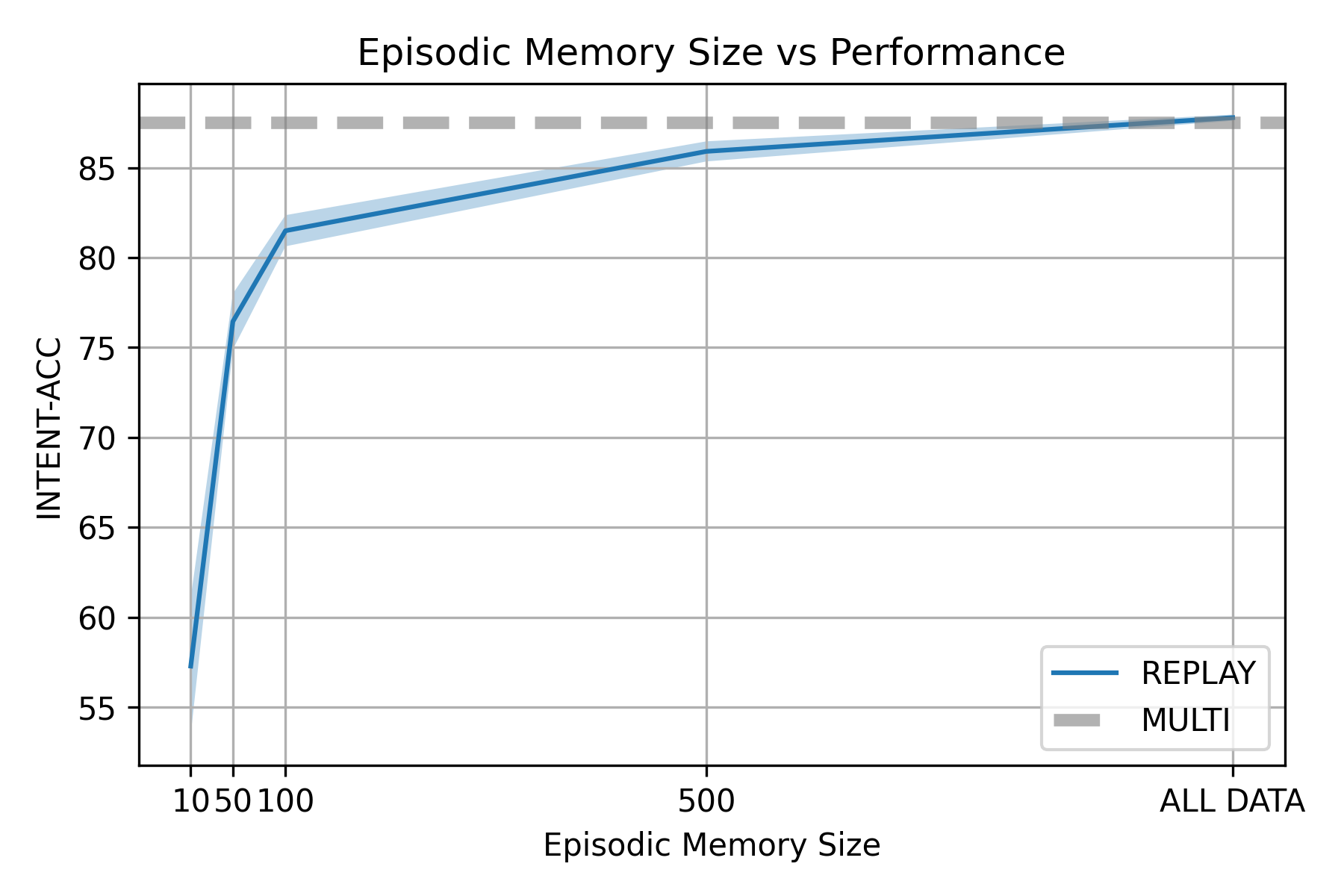}
    \end{subfigure}%
    \begin{subfigure}{.5\textwidth}
        \centering
        \includegraphics[width=\textwidth]{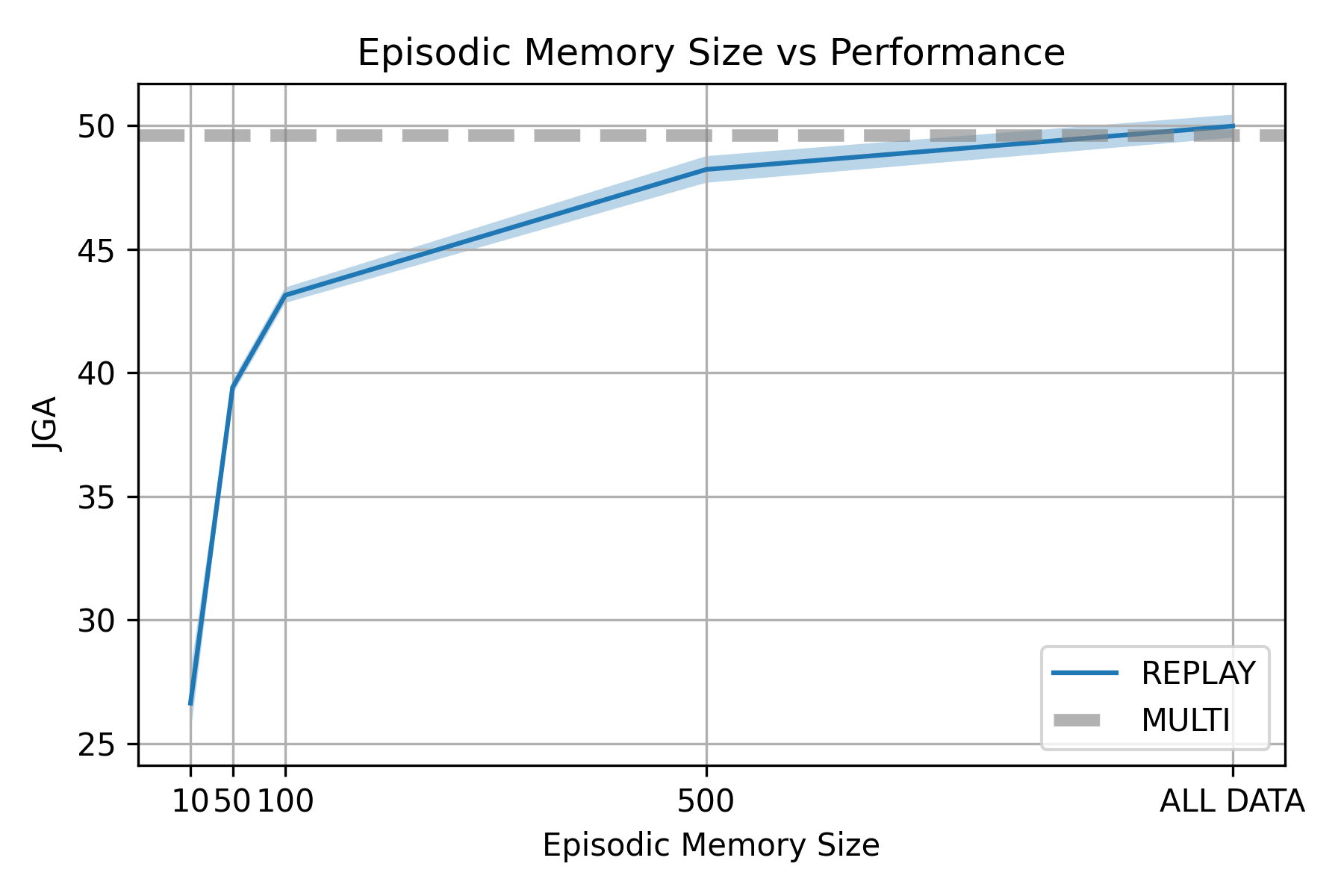}
    \end{subfigure}
    \begin{subfigure}{.5\textwidth}
        \centering
        \includegraphics[width=\textwidth]{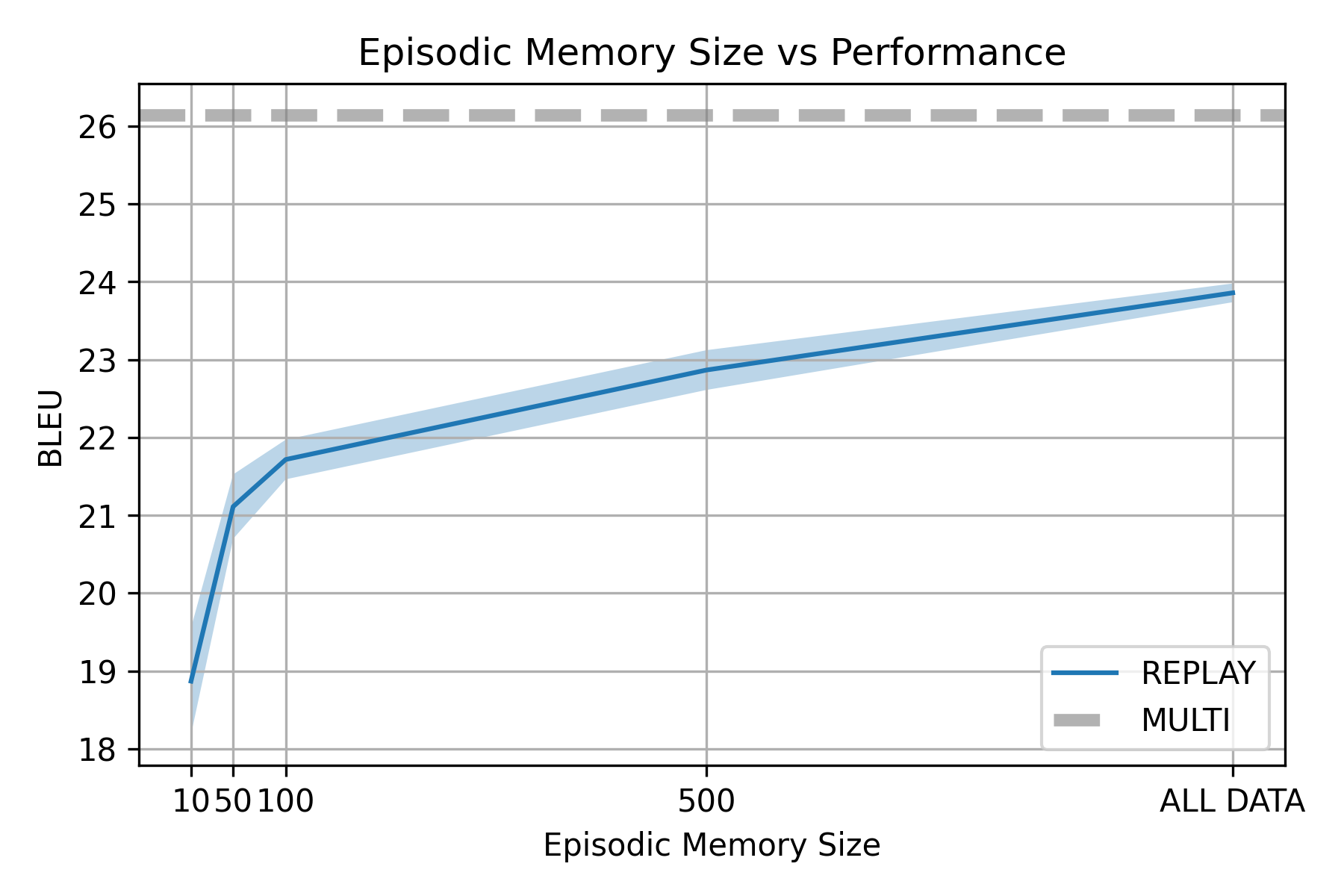}
    \end{subfigure}%
    \begin{subfigure}{.5\textwidth}
        \centering
        \includegraphics[width=\textwidth]{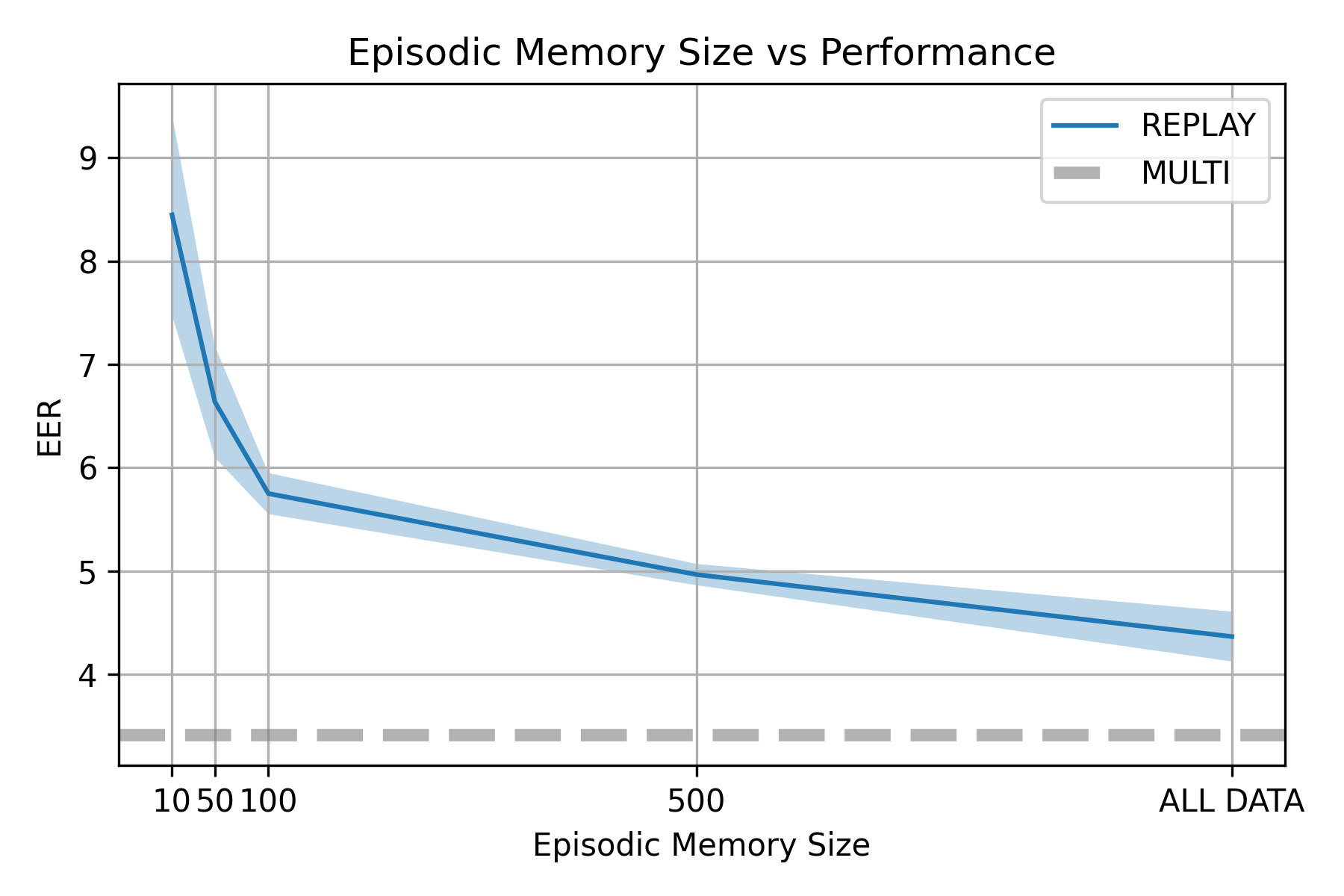}
    \end{subfigure}
    \caption{Ablation study on the size of the episodic-memory vs Intent Accuracy/JGA/BLEU and EER. }
    \label{fig:JGAMEM}
\end{figure}

\subsection{Analysis: Episodic Memory Size}
In this section, we analyze the effect of increasing the episodic memory size for the REPLAY method. Trivially, by including all the training samples in the memory, the model, in the last task, converges to the multitask baseline. Then, the question of how many samples to keep per task to avoid catastrophic forgetting becomes important. In light of this, Figure~\ref{fig:JGAMEM} shows the performance of the model at different episodic memory sizes on the different tasks. Here, we observe that by storing only a few samples per task (10--50) the model still greatly suffers from catastrophic forgetting, where with around 500 samples, which is equivalent to a total of 18,500 samples in our setting, the performance is closer to that of the multitask baseline (i.e., a possible upper bound).

\section{Short Summary}
We propose a benchmark for CL in TODs, with 37 tasks to be learned continuously on four settings: intent recognition, DST, NLG, and end-to-end. Then, we implement three CL methodologies: regularization, rehearsal and architectural. For the latter, we propose a simple yet effective method based on residual adapters and use a perplexity-based classifier to select an adapter to use at inference time. Finally, we analyze the trade-off between the performance, the number of parameters, and the episodic memory sizes of the evaluated baselines, unveiling an insightful trade-off (``no-free lunch") among the methods.

\chapter{Controlling Multi-Skill Dialogue Systems}

Unlike humans, who can do both, task oriented dialogue~\citep{williams2007partially,young2013pomdp} and chit-chat ~\citep{serban2016generative,vinyals2015neural} systems must often remain separate. A more desirable approach for users, however, would be to have a single chat interface that can handle both casual talk and tasks such as reservation and scheduling. This can be formulated as a problem of learning different conversational skills across multiple domains. A skill can be either querying a database, generating daily conversational utterances, or interacting with users in a particular task-domain (e.g. booking a restaurant). One challenge of having multiple skills is that existing datasets either focus only on chit-chat or on goal-oriented dialogues. This is due to the fact that traditional task-oriented systems are modularized~\citep{williams2007partially,hori2009statistical,lee2009example,levin2000stochastic,young2013pomdp}; thus, they cannot be jointly trained with end-to-end architecture as in chit-chat. 
However, recently proposed end-to-end trainable models~\citep{eric-manning:2017:EACLshort,wu2019global,reddy2018multi,yavuzdeepcopy} and datasets~\citep{bordes2016learning,ericKVR2017} allow us to combine task oriented dialogue systems~\citep{budzianowski2018multiwoz,ericKVR2017} and chit-chat~\citep{personachat} into a single benchmark dataset with multiple conversational skills, as shown in Table~\ref{Example-AoP}.

A straighforward solution is to have a single model for all conversational skills, which has shown to be effective to a certain extent by~\citep{zhao2017generative} and \citep{mccann2018natural}. However, putting aside the performance in the tasks, such a fixed shared-parameter framework, without any task-specific designs, would lose controllability and interpretability in response generation. Instead, we propose to model multiple conversational skills using the \textit{Mixture of Experts} (MoE)~\citep{jacobs1991adaptive} paradigm, which is, a model that learns and combines independent specialized experts using a gating function.  

\begin{table}[t]
\centering

\includegraphics[width=0.9\linewidth]{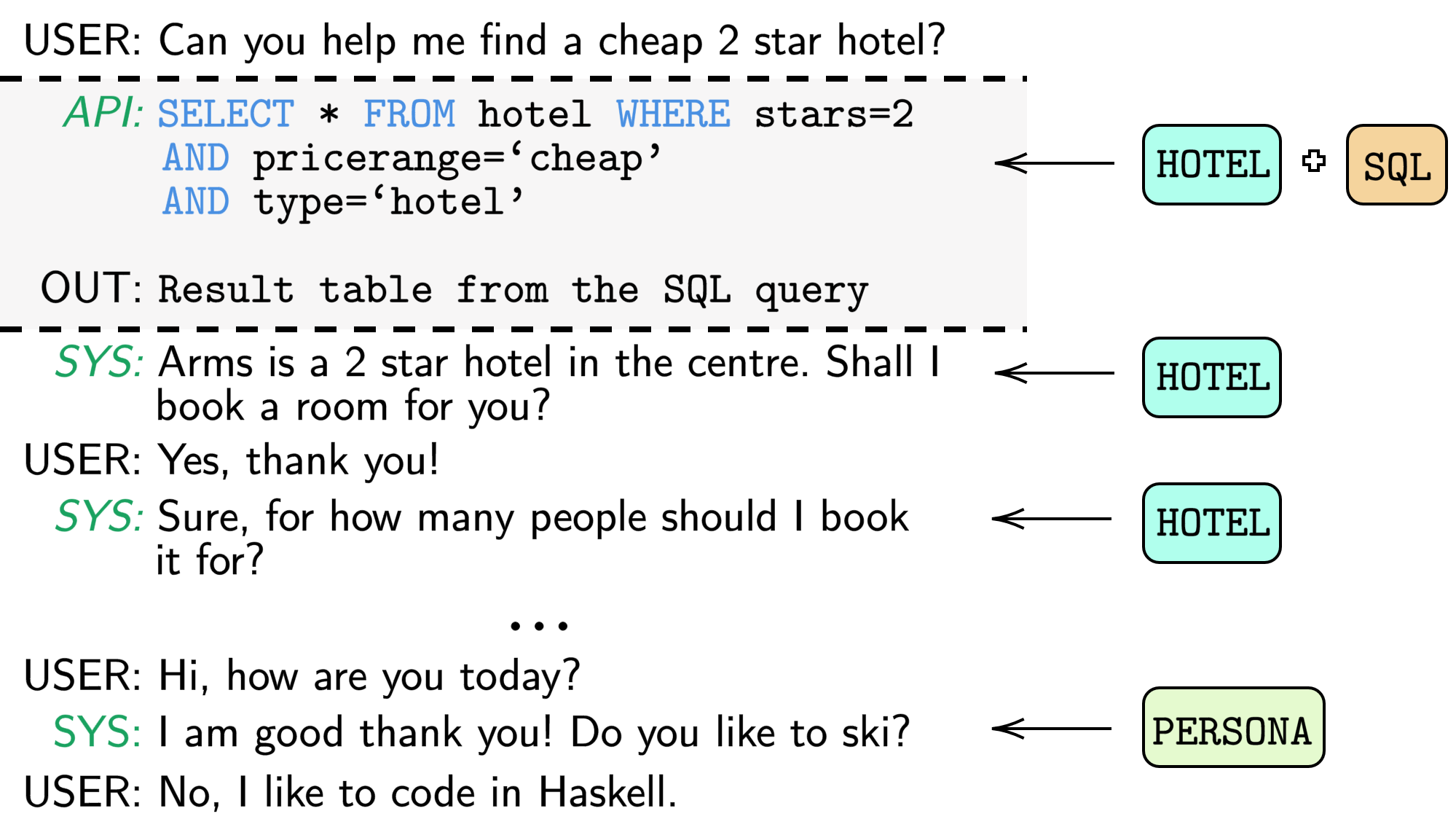}
\caption{An example from the dataset which includes both chit-chat and task-oriented conversations. The model predicts all the \textit{Sys} turns, which include SQL query and generating a response from the API output, which is dynamically updated with the query results. The skills are the prior knowledge needed for the response, where "Persona" refers to chit-chat.}
\label{Example-AoP}
\end{table}

\begin{figure}[t]
    \centering
    \includegraphics[width=0.92\linewidth]{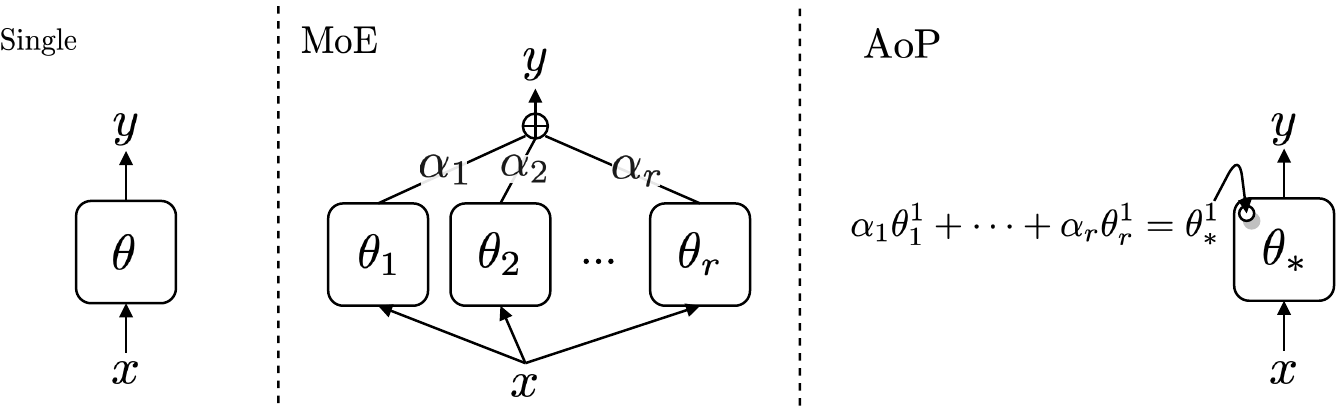}
        \caption{Comparisons between single model, Mixture of Experts (MoE)~\citep{jacobs1991adaptive}, and Attention over Parameters (\textit{AoP}).}
    \label{fig:aopdiagram}
\end{figure}

For instance, each expert may be specialized in different dialogues domains (e.g., Hotel, Train, Chit-Chat etc.) and skills (e.g., generating a SQL query). A popular implementation of MoE \citep{shazeer2017outrageously,kaiser2017one} uses a set of linear transformations (i.e., experts) in between two LSTM \citep{schmidhuber:1987:srl} layers. However, several problems arise with this implementation: 1) the model is computationally expensive as it has to decode each expert multiple times and make the combination at the representation-level; 2) no prior knowledge is injected in the expert selection (e.g., domains); 3) the Seq2Seq model has limited ability to extract information from a Knowledge Base (KB) (generated by the SQL query)~\citep{ericKVR2017}, as required in end-to-end task-oriented dialogues systems~\citep{bordes2016learning}. The latter can be solved by using more advanced multi-hop models, like Transformer~\citep{vaswani2017attention}, but the remaining two problems still need to be addressed. Hence, we: 
\begin{itemize}
    \item propose a novel Transformer-based architecture called \textit{Attention over Parameters} (AoP). This model parameterizes the conversational skills of end-to-end dialogue systems with independent decoder parameters (experts), and learns how to dynamically select and combine the appropriate decoder parameter sets by leveraging prior knowledge from the data, such as domains and skill types; 
    \item prove that AoP is algorithmically more efficient compared to forwarding all the Transformer decoders and then mixing their output representations, like is normally done in MoE. Figure~\ref{fig:aopdiagram} illustrates the high-level intuition of the difference;
    \item empirically show the effectiveness of using specialized parameters in a combined dataset of MultiWOZ~\citep{budzianowski2018multiwoz}, In-Car Assistant~\citep{ericKVR2017}, and Persona-Chat~\citep{personachat}, which to the best of our knowledge, is the first evaluation of this genre, namely end-to-end large-scale multi-domains and skills. Moreover, we show that our model is highly interpretable and is able to combine different learned skills to produce compositional responses.
\end{itemize}

\section{Methodology}
We use the standard encoder-decoder architecture and avoid any task-specific designs~\citep{wu2019global,reddy2018multi}, as we aim to build a generic conversation model for both chit-chat and task-oriented dialogues. Following the notation in Chapter 4, we define two input-output sequences types as 
\[
\begin{array}{cc}
U \rightarrow S_{API} & \mathrm{(API)} \\
U + S_{OUT}\rightarrow S & \mathrm{(Response)},
\end{array}
\]
where $U$ is the dialogue history, $S_{API}$ is an API call, $S_{OUT}$ is the output of the API and $S$ is the system response. $S_{OUT}$ it is the result of a API execution (e.g., table) or plain texts (e.g., persona description), depending on the task. Therefore, we define a dialogue dataset as $D_V=\{(X_i,Y_i)\}_{i}^{N}$, where $(X_i,Y_i)$ is a general input-output pair from the two possible types (API and Response). Finally, we define a binary skill vector $V_i=\{ v_1,\dots,v_{r} \}$ that specifies the type of skill required to generate $Y_i$. This can be considered as a prior vector for learning to select the correct expert during training.~\footnote{The vector $V$ will be absent during the testing.} For example, in Table~\ref{Example-AoP}, the first response is of type API in the Hotel domain. Thus the skill vector $V$ will have $v_{API}=1$ and $v_{\text{Hotel}}=1$, while all the other skills/domains are set to zero.~\footnote{With the assumption that each index in $V$ is assigned a semantic skill (e.g., API position $i$).} More importantly, we may set the vector $V$ to have multiple skills to force the model to compose skills to achieve a semantic compositionality of different experts.

\subsection{Attention over Parameters}
\label{sec:aop}
By following the Transformer notation in Chapter 2, the main idea is to produce a single set of parameters for decoder TRS$_{dec}$ by the weighted sum of $r$ independently parameterized decoders. This process is similar to attention~\citep{luong-pham-manning:2015:EMNLP}, where the memories are the parameters and the query is the encoded representation. Let us define $\Theta=[ \theta_1,\dots,\theta_r ]$ as the list of parameters for $r$ decoders, since a TRS$_{dec}$ is represented by its parameters $\theta$. Since each $\theta$ can be sized in the order of millions, we assign the corresponding key vectors to each $\theta$, similar to key-value memory networks~\citep{miller2016key}. Thus, we use a key matrix $K\in \mathbb{R}^{d_{model} \times r}$ and a Recurrent Neural Network (RNN), in this instance a GRU~\citep{cho2014learning}, to produce the query vector by processing the encoder output $H$.  The attention weights for each decoder's parameters is computed as follows:
\begin{align}
    q      &= \textrm{RNN}(H)\\
    \alpha &= \text{Softmax}(qK), 
    \label{eq_att}
\end{align}
where $q \in \mathbb{R}^{d_{model}}$ and $\alpha \in \mathbb{R}^r$ is the attention vectors where each $\alpha_i$ is the score corresponding to $\theta_i$. Hence, the new set of parameters is computed as follows:
\begin{equation}
    \theta^* = \sum_i^r \alpha_i \theta_i,\label{eq:aop}
\end{equation}
The combined set of parameters $\theta^*$ are then used to initialize a new TRS$_{dec}$, and Equation 2.29 (Chapter 2) is applied to the input based on this. Equation~\ref{eq_att} is similar to the gating function proposed in~\citep{shazeer2017outrageously} and \cite{jacobs1991adaptive}, but the resulting scoring vector $\alpha$ is applied directly to the parameter instead of the output representation of each decoder, holding an algorithmically faster computation. 

\paragraph{Theorem}\quad \\
\textit{The computation cost of Attention over Parameters (AoP) is always lower than of Mixture Of Experts (MoE) for a sequence longer than 1.}
\paragraph{Proof}\quad \\
Let $f_\theta: \mathbb{R}^d \rightarrow \mathbb{R}^n$ be a generic function parametrized by $\theta$. Without loss of generality, we define $\theta$ as an affine transformation $W \in  \mathbb{R}^{d \times n}$. Let $X \in  \mathbb{R}^{t \times d}$ be a generic input sequence of length $t$ and $d$ dimensional size. Let the set $F=[f_{\theta_1},\cdots,f_{\theta_r}]$ be the set of $r$ experts. Hence, the operation done by MoE is:
\begin{equation}
    \textrm{MoE}(X)= f_{\theta_1}(X) + \cdots + f_{\theta_r}(X)= XW_1 + \cdots + XW_r,
\end{equation}
Thus the computational cost in terms of operation is $\mathcal{O}(rtdn + rtn)$ since the cost of $ f_{\theta_i}(X)$ is $\mathcal{O}(tdn)$ and it is repeated $r$ times, and the cost of summing the representation is $\mathcal{O}(rtn)$. On the other hand, the operation done by AoP is:
\begin{gather}
    \theta^* = \theta_1 + \cdots + \theta_r = W_1 + \cdots + W_r\\
    \textrm{AoP}(X) = f_{\theta^*}(X) =  XW^*.
\end{gather}
In this case the computational cost in terms of operation is $\mathcal{O}((r+t)dn)$ since the cost of summing the parameters is $\mathcal{O}(rdn)$ and the cost of $f_{\theta^*}$ is $\mathcal{O}(tdn)$. Hence, it is easy to verify that if $t>1$, then
\begin{equation}
    rtdn + rtn \geq (rt)dn \geq (r+t)dn.
\end{equation}
Furthermore, the assumption of using a simple affine transformation $W$ is actually an optimal case. Indeed, assuming that the cost of parameters sum is equal to the number of operations is optimistic. For instance, already, by using attention, the number of operations increases, but the number of parameters remains constant. $\blacksquare$
\\

Importantly, if we apply $\alpha$ to each of the output representations $T_i$ generated by the TRS$_{dec}^i$, we end up having a Transformer-based implementation of MoE. We call this model \textit{ Attention over Representation} (AoR). Finally, an additional loss term is used to supervise the attention vector $\alpha$ by using the prior knowledge vector $V$. Since multiple decoder parameters can be selected at the same time, we use binary cross-entropy to train each $\alpha_i$. Thus a second loss is defined as: 
\begin{equation}
L_{\theta^*}(V) =- \sum_{j=1}^{r} v_{j} \times \text{log} \sigma(qK)_{j} + (1-v_{j}) \times \text{log} (1- \sigma(q K)_{j}),
\end{equation}
The final loss is the summation of $L_{\theta^*}(D^V)$ and $L_{\theta^*}(V)$. 

Finally, in AoP, in general in the MoE framework, stacking multiple layers (e.g., Transformer) leads to models with a large number of parameters, since multiple experts are repeated across layers. An elegant workaround is the Universal Transformer~\citep{dehghani2018universal}, which loops over an \textit{unique} layer and, as shown by~\citep{dehghani2018universal}, holds similar or better performance than a multi-layer Transformer. In our experiment, we report a version of AoP that uses this architecture, which for instance, does not add any further parameters to the model.

\section{Experiments and Results}
\subsection{Dataset} 
To evaluate the performance of our model for different conversational skills, we propose to combine three publicly available datasets: MultiWOZ~\citep{budzianowski2018multiwoz}, Stanford Multi-domain Dialogue~\citep{ericKVR2017} and Persona-Chat~\citep{personachat}. 

\begin{wrapfigure}{r}{0.5\textwidth}

\begin{tabular}{r|ccc}
\hline
\multicolumn{1}{l|}{} & \textbf{SMD} & \textbf{MWOZ} & \textbf{Persona} \\ \hline
\textit{\#Dialogues} & 2425 & 8,438 & 12,875 \\ \hline
\textit{\#turns} & 12,732 & 115,424 & 192,690 \\ \hline
\textit{Avg. turns} & 5.25 & 13.68 & 14.97 \\ \hline
\textit{Avg. tokens} & 8.02 & 13.18 & 11.96 \\ \hline
\textit{Vocab} & 2,842 & 24,071 & 20,343 \\ \hline
\end{tabular}
\makeatletter\def\@captype{table}\makeatother
\caption{Datasets statistics}
\label{data_stat}
\end{wrapfigure}
\paragraph{MultiWOZ (MWOZ)} is a human-to-human multi-domain goal-oriented dataset annotated with dialogue acts and states. In this dataset, there are seven domains: Taxi, Police, Restaurant, Hospital, Hotel, Attraction, Train. There are also two API interfaces, SQL and BOOK, the former of which is used to retrieve information about a certain domain and the latter of which is used to book restaurants, hotels, trains, and taxis. 
We refine this dataset to include SQL/BOOK queries and their outputs using the same annotation schema as~\citep{bordes2016learning}.

Each response can either be plain text conversation with the user or SQL/BOOK queries, and the memory is dynamically populated with the results from the queries as the generated response is based on such information. 
This transformation allows us to train end-to-end models that learns how and when to produce SQL queries, to retrieve knowledge from a dynamic memory, and to produce plain text responses.
A detailed explanation is reported in Appendix B, together with some samples. 

\paragraph{Stanford Multi-domain Dialogue (SMD)}is another human-to-human multi-domain goal-oriented dataset that is designed for end-to-end training. There are three domains in this dataset (Point-of-Interest, Weather, Calendar). The difference between this dataset and MWOZ is that each dialogue is associated with a set of records relevant to the dialogues. The $S_{OUT}$ is fixed in this case. Thus the model does not need to issue any API-calls. However, retrieving the correct entities from the memory is more challenging as the model has to compare alternatives among records.

\paragraph{Persona-Chat}is a multi-turn conversational dataset, in which two speakers are paired and different persona descriptions (4--5 sentences) are randomly assigned to each of them. For example, ``\textit{I am an old man}'' and ``\textit{I like to play football}'' are one of the possible persona descriptions provided to the system. Training models using this dataset results in a more persona consistent and fluent conversation compared to other existing datasets~\citep{personachat}. Currently, this dataset is the standard benchmark for chit-chat systems. Thus, we include it in our evaluation.

For all three datasets, we use the training/validation/test split provided by the author and we keep all the \textit{real entities} in input instead of using their delexicalized versions as in~\citep{budzianowski2018multiwoz} and \cite{ericKVR2017}. This makes the task more challenging, but at the same time more interesting since we force the model to produce real entities instead of generic and frequent placeholders. Table~\ref{data_stat} summarizes the dataset statistics in terms of number of dialogues, turns, and unique tokens. Finally, we merge the three datasets, obtaining 154,768/19,713/19,528 for training, validation and, test respectively and a vocabulary size of 37,069 unique tokens.

\subsection{Evaluation Metrics}
\paragraph{Goal-Oriented} For both MWOZ and SMD, we follow the evaluation done by existing works~\citep{eric-manning:2017:EACLshort,zhao2017generative,madotto2018mem2seq} and \cite{wu2017dstc6}. We use the BLEU\footnote{Using the \texttt{multi-bleu.perl} script} score~\citep{pAPIneniBLEU2002} to measure the response fluency and Entity F1-Score~\citep{wen2016network,zhao2017generative} to evaluate the ability of the model to generate relevant entities from the dynamic memory. Since MWOZ also includes SQL and BOOK queries, we compute the exact match accuracy ($ACC_{SQL}$ and $ACC_{BOOK}$) and BLEU score ($BLEU_{SQL}$ and $BLEU_{BOOK}$). Furthermore, we also report the F1-score for each domain in both MWOZ and SMD. 

\paragraph{Chit-Chat} We compare the perplexity, BLEU score, F1-score~\citep{dinan2020second}, and consistency score of the generated sentences with the human-generated prediction. The consistency score~\cite{madotto2019personalizing} is computed using a Natural Language Inference (NLI) model trained on dialogue NLI~\citep{dnli}, a recently proposed corpus based on the Persona dataset. We fine-tune a pre-trained BERT model~\citep{devlin2018bert} using the dialogue DNLI corpus and achieve a test set accuracy of 88.43\%, which is similar to the best-reported model in~\citep{dnli}. The consistency score is defined as follows:
\begin{align}
    &\text{\textbf{NLI}}(u, p_j) = \Bigg\{
  \begin{array}{rcr}
    1 & \text{if $u$ entails $p_j$} \\
    0 & \text{if $u$ is independent to $p_j$} \\
    -1 & \text{if $u$ contradicts $p_j$} \\
  \end{array} \nonumber
  \\
  &\text{\textit{\textbf{C}}}(u) = \sum_j^{m} \text{\textbf{NLI}}(u, p_j),
\end{align}
where $u$ is a generated utterance and $p_j$ is one sentence in the persona description. In~\cite{dnli} and \cite{madotto2019personalizing}, the authors showed that by re-ranking the beam search hypothesis using the DNLI score (i.e., \textit{\textbf{C}} score), they achieved a substantial improvement in dialogue consistency. Intuitively, having a higher consistency \textit{\textbf{C}} score means having a more persona-consistent dialogue response.

\subsection{Baselines}
In our experiments, we compare Sequence-to-Sequence (\textit{Seq2Seq})~\citep{see-liu-manning:2017:Long}, Transformer (\textit{TRS})~\citep{vaswani2017attention}, Mixture of Experts (\textit{MoE})~\citep{shazeer2017outrageously} and Attention over Representation (\textit{AoR}) with our proposed Attention over Parameters (\textit{AoP}). In all the models, we used the same copy-mechanism as in~\citep{see-liu-manning:2017:Long}. In \textit{AoR}, instead of mixing the parameters as in Equation \ref{eq:aop}, we mix the output representation of each Transformer decoder (i.e. Equation 2.29). For all \textit{AoP}, \textit{AoR}, and \textit{MoE}, $r=13$ is the number of decoders (experts): two skills for SQL and BOOK and 10 different domains for MWOZ+SMD and one for Persona-Chat. Furthermore, we also include the following experimental models: \textit{AoP} using the gold attention vector $V$, which we refer to as \textit{AoP} w/ Oracle (or \textit{AoP} + O); \textit{AoP} trained by removing the $L_{\theta^*}(V)$ from the optimization, which we refer to as \textit{AoP w/o $L_{\theta^*}(V)$}; and as previously mentioned, the Universal Transformer for both AoP, which we call \textit{AoP + U}, and the standard Transformer, which we call \textit{TRS + U} (six hops). Detailed descriptions of all models and the full set of hyper-parameters used in the experiments are reported in Appendix C. 

\subsection{Results} 
Table~\ref{mwoz-smd} and Table~\ref{persona} show the respective evaluation results on the MWOZ+SMD and Persona-Chat datasets. From Table~\ref{mwoz-smd}, we can identify three patterns 1) \textit{AoP} and \textit{AoR} perform consistently better then the other baselines, which shows the effectiveness of combining parameters using the correct prior $V$; 2) \textit{AoP} performs consistently, but marginally, better than \textit{AoR}, with the advantage of an algorithmically faster inference and 3) using Oracle (\textit{AoP}+O) gives the highest performance in all measures, which shows the performance upper-bound for \textit{AoP}. 
\begin{wrapfigure}{r}{0.5\textwidth}
\begin{tabular}{r|cccc}
\hline
\multicolumn{1}{c|}{\textbf{Model}} & \textbf{Ppl.} & \textbf{F1} & \textbf{C} & \textbf{BLEU} \\ \hline
\textit{Seq2Seq}         & 39.42  &	6.33  &	0.11  &	2.79  \\ \hline
\textit{TRS}             & 43.12  &	7.00  &	0.07  &	2.56  \\ \hline
\textit{MoE}             & \textbf{38.63}  & \textbf{7.33}  &	0.19  &	2.92  \\ \hline
\textit{AoR}            & 40.18   & 6.66  & 0.12  & 2.69  \\ \hline
\textit{AoP}       & 39.14  &	7.00  &	\textbf{0.21}  &	\textbf{3.06}  \\ \hline\hline   
\textit{TRS + U}         & 43.04  &	\textbf{7.33}  & 0.15  &	2.66  \\ \hline
\textit{AoP + U}         & \textbf{37.40}  &	7.00  &	\textbf{0.29}  &	\textbf{3.22}  \\ \hline \hline 
\textit{AoP w/o $L_{\theta^*}(V)$}       & 42.81  &	6.66  &	0.12  &	2.85  \\ \hline
\textit{AoP + O}         & \textit{40.16}  &	\textit{7.33}  &	\textit{0.21}  &	\textit{2.91}  \\ \hline 
\end{tabular}
\caption{Results for the Persona-Chat dataset.} \label{persona}
\end{wrapfigure}Hence, the performance gap when not using Oracle attention is most likely due to the error in attention $\alpha$ (2\% error rate). Moreover, Table~\ref{mwoz-smd} shows that by removing $L_{\theta^*}(V)$ (\textit{AoP w/o $L_{\theta^*}(V)$}) the model performance decreases, which confirms that good inductive bias is important for learning how to select and combine parameters (experts). Additionally, in Appendix D and E, we report the per-domain F1-score for SQL, BOOK and sentences, and Table~\ref{persona} and Table~\ref{mwoz-smd} with the standard deviation among the three runs. 

Furthermore, from Table~\ref{persona}, we can see that \textit{MoE} has the lowest perplexity and F1-score, but \textit{AoP} has the highest consistency and BLEU score. Note that the perplexity reported in~\citep{personachat} is lower since the vocabulary used in their experiments was smaller. In general, the difference in performance among the models is marginal, except for the consistency score; thus, we can conclude that all the models can learn this skill reasonably well. Consistent with the previous results, when $L_{\theta^*}(V)$ is removed from the optimization, the models' performance decreases. 
\begin{table}[t]
\centering
\resizebox{0.9\textwidth}{!}{%
\begin{tabular}{r|cc|cc|cc}
\hline
\multicolumn{1}{c|}{\textbf{Model}} & \textbf{F1} & \textbf{BLEU} & \textbf{SQL$_{Acc}$} & \textbf{SQL$_{BLEU}$} & \textbf{BOOK$_{Acc}$} & \textbf{BOOK$_{BLEU}$} \\ \hline
\textit{Seq2Seq}        & 38.37  &	9.42     & 49.97  &	81.75  & 39.05  & 79.00  \\ \hline
\textit{TRS}            & 36.91  &	9.92   & 61.96  &	89.08  & 46.51  & 78.41  \\ \hline
\textit{MoE}            & 38.64  &	9.47    & 53.60  &	85.38  & 37.23  & 78.55  \\ \hline
\textit{AoR}        & 40.36  &	10.66  &	69.39  &	90.64  &	52.15  &	81.15              \\ \hline
\textit{AoP}  & \textbf{42.26}  &	\textbf{11.14}   &\textbf{71.1}  &	\textbf{90.90}  & \textbf{56.31}  & \textbf{84.08}  \\ \hline \hline
\textit{TRS + U}        & 39.39  &9.29	  & 61.80 &89.70	& 50.16 & 79.05  \\ \hline
\textit{AoP + U}        & \textbf{44.04}  &	\textbf{11.26}  &	\textbf{74.83}  &	\textbf{91.90}  &	\textbf{56.37}  &	\textbf{84.15}              \\ \hline \hline
\textit{AoP w/o $L_{\theta^*}(V)$}        & 38.50  &	10.50     & 61.47  &	88.28  & 52.61  & 80.34  \\ \hline
\textit{AoP+O}   & \textit{46.36}  &	\textit{11.99}   & \textit{73.41}  &	\textit{93.81}  & \textit{56.18} & \textit{86.42}  \\ \hline
\end{tabular}
}
\caption{Results for the goal-oriented responses in both MWOZ and SMD. The last raw uses the Oracle, and bold-faced are best in each setting (w and w/o Universal). Results are averaged among three run (full table in Appendix F).} \label{mwoz-smd}
\end{table}

Finally, in both Table~\ref{persona} and Table~\ref{mwoz-smd}, we report the results obtained by using the Universal Transformer, for both AoP and the Transformer. By adding the layer recursion, both models are able to consistently improve on all the evaluated measures, in both Persona-Chat and the Task-Oriented tasks. AoP, in particular, achieves better performance than Oracle (single layer) on SQL accuracy, and has a consistently better performance in the Persona-Chat evaluation. 

\begin{table}[t]
\caption{Selecting different skills thought the attention vector $\alpha$ results in a skill-consistent response. \textit{AoP} response activates SQL and Train.}\label{Composit}
\centering
\includegraphics[width=\linewidth]{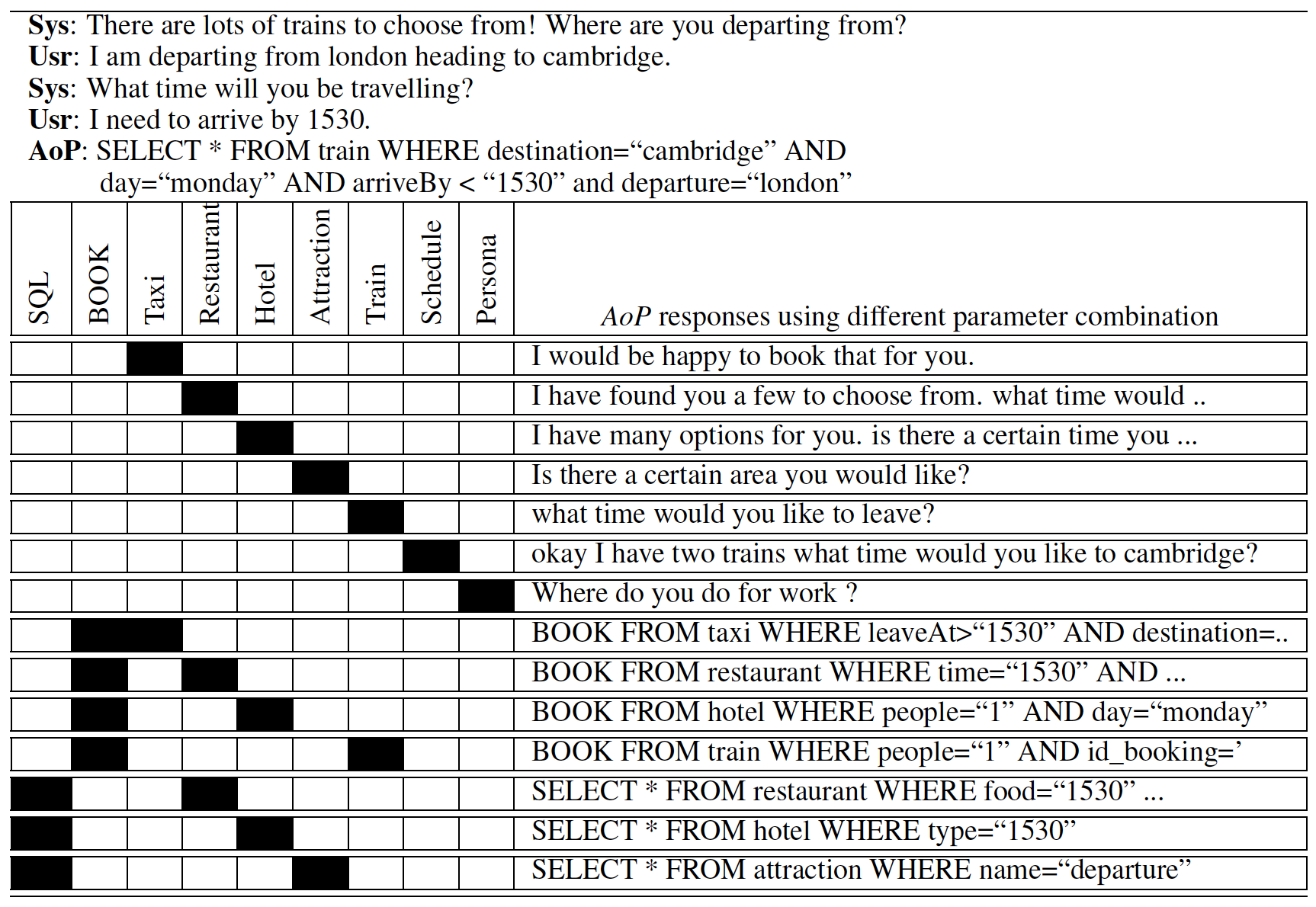}
\end{table}
\section{Skill Composition}
To demonstrate the effectiveness of our model in learning independent skills and composing them together, we \textit{manually trigger} skills by modifying $\alpha$ and generate 14 different responses for the same input dialogue context. This experiment allows us to verify whether the model accurately captures the meaning of each skill and whether it can properly learn to compose the selected parameters (skills). Table~\ref{Composit} first shows the dialogue history along with the response of \textit{AoP} on the top, and then different responses generated by modifying $\alpha$ (black cells correspond to 1 in the vector, while the white ones are 0). By analyzing Table~\ref{Composit}, we can notice that:

\begin{itemize}
    \item The model learns the correct semantics of each skill. For instance, the \textit{AoP} response is of type SQL and Train, and by deactivating the SQL skill and activating other domain-skills, including Train, we can see that the responses are grammatical and they are coherent with the selected skill semantics. For instance, by just selecting \textit{Train}, the generated answer becomes \textit{``what time would you like to leave?''}, which is coherent with the dialogue context since such information has not been yet provided. Interestingly, when the \textit{Persona} skill is selected, the generated response is conversational and also coherent with the dialogue, even though it is less fluent.
    
    \item The model effectively learns how to compose multiple skills. For instance, when SQL or BOOK is triggered the response produces the correct SQL syntax (e.g. ``SELECT * FROM ...'' etc.). By also adding the corresponding domain-skill, the model generates the correct query format and attributes relative to the domain type (e.g., in \textit{SQL, Restaurant}, the model queries with the relevant attribute \textit{food} for restaurants). 
\end{itemize}

\begin{figure}[t]
    \centering
    \includegraphics[width=0.98\linewidth]{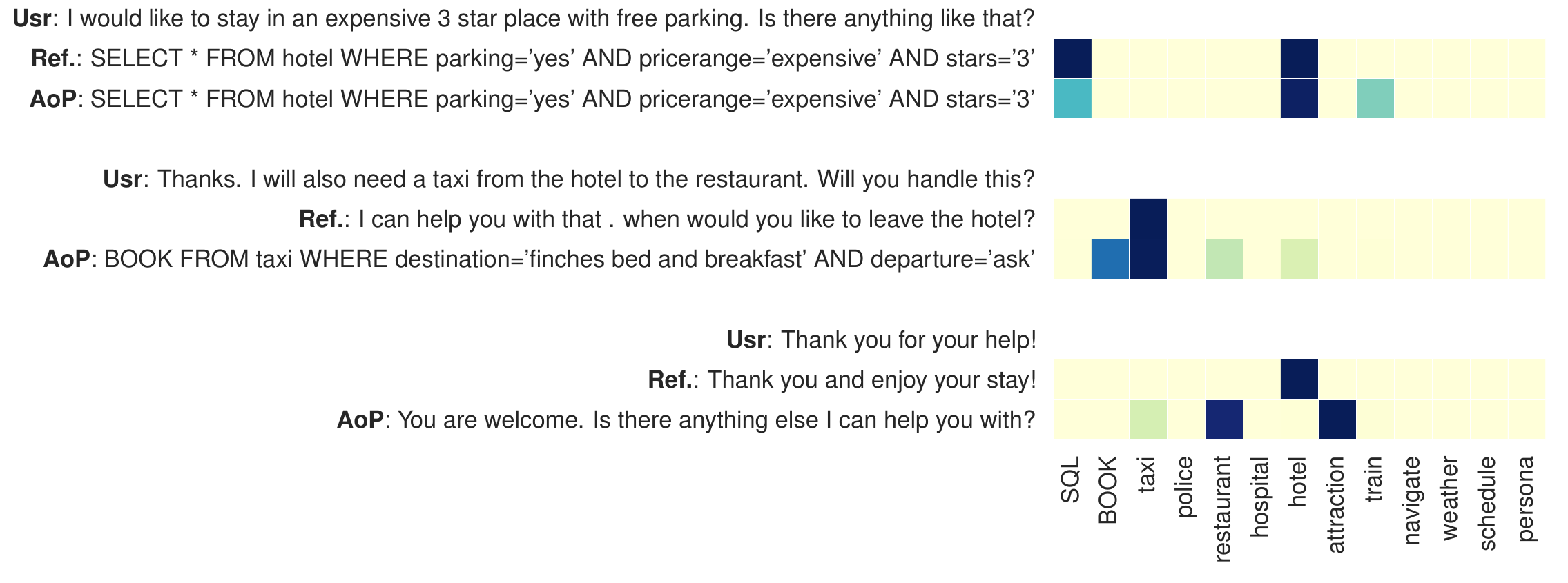}
    \caption{AoP visualization, vector $\alpha$ for different reference, (\textbf{Ref.}) and \textit{AoP}, generated answers. Top rows (\textbf{Usr}) are the last utterances from each dialogue context.}
    \label{fig:Viz}
\end{figure}

\section{Attention Visualization} 
Figure~\ref{fig:Viz} shows the attention vector $\alpha$ over parameters for different generated sentences. Note that the sentences shown in the figure are the expected responses, not the given input, which is omitted since it can be very large. In this figure, and by analyzing more examples, we can identify two patterns: 
\begin{itemize}
    \item \textit{AoP} learns to focus on the correct skills (i.e., SQL, BOOK) when API-calls are needed. From the first example in Figure~\ref{fig:Viz}, we can see that the activations in $\alpha$ are consistent with those in the correct attention vector $P$. There are also false positives, in which \textit{AoP} puts too high a weight on BOOK when the correct response is plain text that should request more information from the user (\textit{i can help you with that. when would you like to leave the hotel?}). However, we can see that this example is, in fact, "almost correct" as triggering a booking API-call may also be considered a valid response. Meanwhile, the third example also fails to attend to the correct skill, but, in fact, generates a very fluent and relevant response. This is most likely because the answer is simple and generic.
    \item The attention often focuses on multiple skills not directly relevant to the task. We observe this pattern especially when there are other skill-related entities mentioned in the context or the response. For example, in the second dialogue example in Figure~\ref{fig:Viz}, we notice that \textit{AoP} not only accurately focuses on the \textit{taxi} domain, but also has non-negligible activations for \textit{restaurant} and \textit{hotel}. This is because the words ``hotel" and ``restaurant" are both mentioned in the dialogue context and the model has to produce two entities of the same type (\textit{finches bed and breakfast} and \textit{ask}). 
\end{itemize}


\section{Short Summary}
We propose a novel way to train a single end-to-end dialogue model with multiple composable and interpretable skills. Unlike previous work, which has mostly focused on the representation-level mixing~\citep{shazeer2017outrageously}, our proposed approach, \textit{Attention over Parameters}, learns how to softly combine independent sets of specialized parameters (making an SQL-Query, conversing with a consistent persona, etc.) into a single set of parameters. By doing so, we not only achieve compositionality and interpretability but also gain an algorithmically faster inference speed. To train and evaluate our model, we organize a multi-domain task-oriented dataset into end-to-end trainable formats and combine it with a conversational dataset (Persona-Chat). Our model learns to consider each task and domain as separate skills that can be composed with each other, or used independently, and we verify the effectiveness of the interpretability and compositionality with competitive experimental results and thorough analysis.  
\chapter{Conclusion}

In this thesis, we focused on the controllability of deep learning-based, end-to-end, generative dialogue systems in both task-oriented and chit-chat scenarios. In particular, we explored the different aspects of controlling generative dialogue systems: we described how to control the style and topics of chit-chat models, how to continuously control and extend task-oriented dialogue systems, and how to compose and control multi-skill dialogue models.
 
In controlling the style and topics of generative chit-chat dialogue systems, we first proposed and evaluated plug-and-play methods for controllable response generation, which do not require dialogue specific datasets nor rely on fine-tuning a large model. While effective, the decoding procedure induce considerable computational overhead, rendering the conversational model unsuitable for interactive usage. To overcome this, we introduced an approach that does not require further computation at decoding time, nor does it require any fine-tuning of a large language model. We demonstrated, through extensive automatic and human evaluation, a high degree of control over the generated conversational
responses with regard to multiple desired attributes, while also being fluent.

In continuously controlling and extending task-oriented dialogue systems, we proposed a benchmark for task-oriented dialogue systems with 37 domains to be learned continuously in four settings: intent recognition, state tracking, natural language generation, and end-to-end. Moreover, we implemented and compared multiple existing continual learning baselines, and  we proposed a highly controllable architectural method based on residual adapters. Our experiments demonstrated that the proposed architectural method and a simple replay-based strategy perform comparably well, but they both achieve inferior performance to the multi-task learning baseline, in which all the data are shown at once, showing that this setting is a challenge. Furthermore, we reveal several trade-offs between learning methods in terms of parameter usage and memory size, which are important in the design of task-oriented dialogue systems. 

In learning to compose and control multi-skill dialogue models, we proposed to learn a dialogue system that independently parameterizes dialogue skills, and learns to select and combine each of them through Attention over Parameters (AoP). The experimental results showed that this approach achieves competitive performance on a combined dataset of MultiWOZ, In-Car Assistant, and Persona-Chat, and we demonstrated that each dialogue skill is effectively learned and can be combined with other skills to produce selective responses.

Overall, the common technique proposed in the thesis for controlling style\&topics, continuous domain learning and multi-skills system, is the independent parameterization of different dialogue skills. This is done by using residual adapters in style\&topics and continuously learning domains, and by using Transformer decoders in multi-skill systems. Independently by the network choice, the parametrization of each dialogue skill leads to highly controllable, extendable and composable dialogue systems. The choice of residual adapter is especially meaningful for lightweight plug-and-play skills like style\&topics and adding dialogue skills through time. While mixing different Transformer decoders is important in multi-skills systems since allows to mix and match different skills and achieve compositional behaviours.    

In future work, we expect to develop dialogue systems that are easy to control and are also more data-efficient and can learn from user interaction rather than supervised data. In fact, in the three chapters of this thesis, we open new exciting research directions for dialogue systems: 1) controlling style and topic is still in its infancy. In Chapter 3, we control seven attributes, while in future works, we would like to extend our methodology to control many more attributes (e.g., emotions) and to compose multiple adapters to achieve compositionality (e.g., positive + sport etc.). 2) controlling how to extend dialogue systems with a clear distinction between tasks is challenging and exciting, as seen in Chapter 4. However, removing the distinction between tasks leads to a more challenging setting where the model is forced to learn from a data sample stream. This streaming learning reassembles human learning and would make a model that can learn from user interaction rather than supervised data. 3) controlling multi-skills dialogue systems opens future research directions similar to point 1 and 2, plus the possibility of studying more advanced methods to compose different skills to obtain exponential dialogue skills by using only a linear number of experts. Although in Chapter 5, we show some compositional behaviours, it is still an open and exciting challenge to achieve compositional behaviours with more dialogue skills (e.g., style in task-oriented dialogue). 


\begin{figure}[h]
    \centering
    \includegraphics[width=\linewidth]{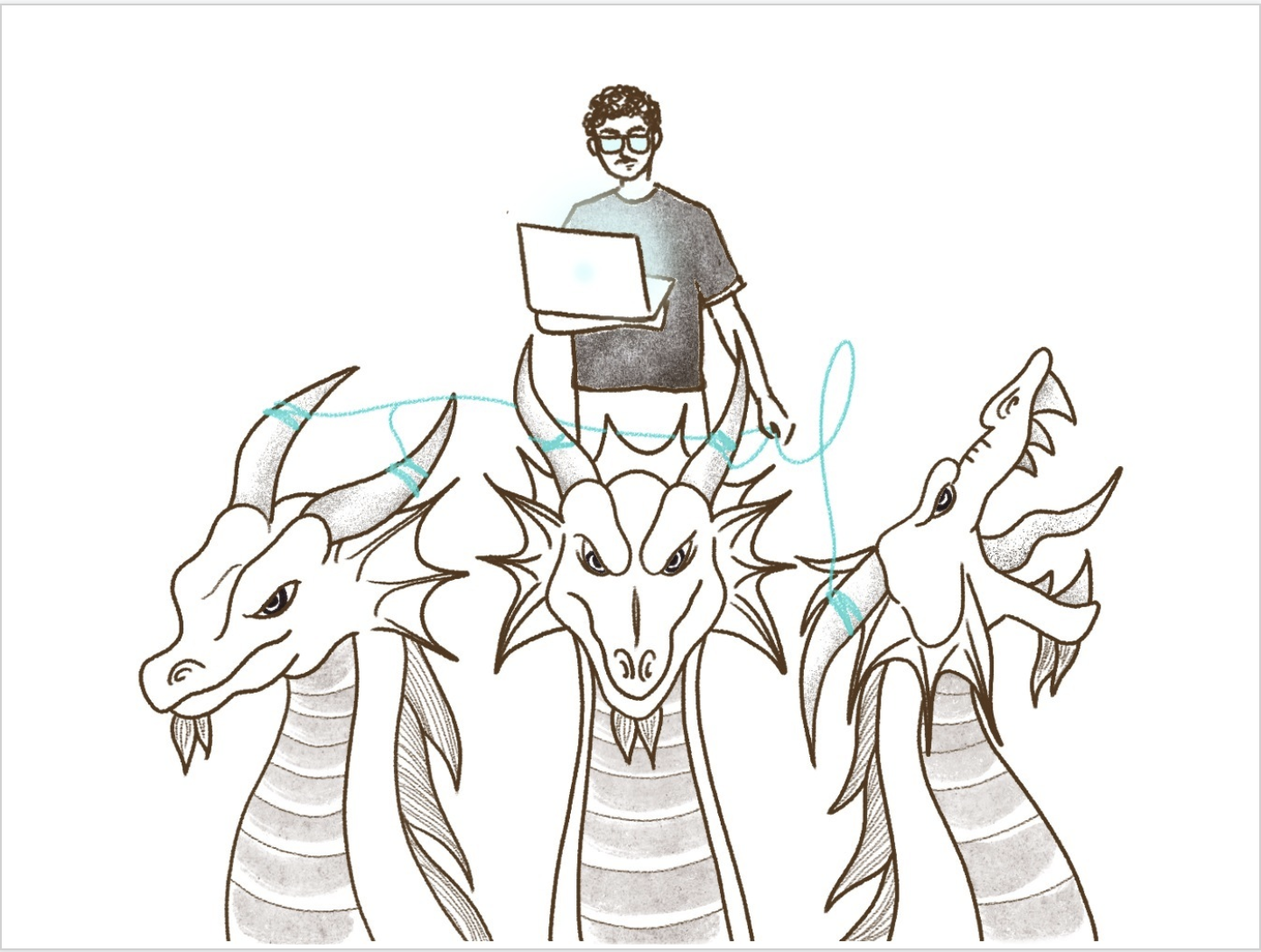}
    \caption{The beasts are tamed.}
\end{figure}

\newpage
\addcontentsline{toc}{chapter}{Reference}
\bibliographystyle{IEEEtranN}
\bibliography{reference} 



\end{document}